\pdfoutput=1

\documentclass[11pt]{article}

\usepackage[preprint]{acl}

\usepackage{times}
\usepackage{latexsym}

\usepackage[T1]{fontenc}

\usepackage[utf8]{inputenc}

\usepackage{microtype}

\usepackage{nicefrac}       
\usepackage{hyperref}
\usepackage{url}
\usepackage{footmisc}

\usepackage{inconsolata}

\usepackage{graphicx}

\usepackage{booktabs}
\usepackage{algorithm}
\usepackage{algpseudocode}
\usepackage{adjustbox}
\usepackage{user}
\usepackage{svg}
\usepackage{xcolor}
\usepackage{microtype}
\usepackage{tcolorbox}
\usepackage{enumitem}

\title{Attention Instruction: Amplifying Attention in the Middle via Prompting}

\author{Meiru Zhang \\
  University of Cambridge \\
  \texttt{mz468@cam.ac.uk} \\\And 
  Zaiqiao Meng\thanks{Corresponding Author.} \\
  University of Glasgow \\
  \texttt{zaiqiao.meng@glasgow.ac.uk} \\\And 
  Nigel Collier \\
  University of Cambridge \\
  \texttt{nhc30@cam.ac.uk} \\}

\begin{document}
\maketitle
\begin{abstract}

The context window of large language models has been extended to 128k tokens or more. However, language models still suffer from position bias and have difficulty in accessing and using the middle part of the context due to the lack of attention. We examine the relative position awareness of LLMs and the feasibility of mitigating disproportional attention through prompting. We augment the original task instruction with \texttt{attention instructions}\footnote{Code: \href{https://github.com/meiru-cam/AttentionInstruction}{github.com/meiru-cam/AttentionInstruction}} that direct language models to allocate more attention towards a selected segment of the context. We conduct a comprehensive investigation on multi-document question answering task with both position-based and index-based instructions. We find that language models do not have relative position awareness of the context. Nevertheless, they demonstrate the capacity to adapt attention to a specific segment using matching indexes. Our analysis contributes to a deeper understanding of position bias in LLMs and provides a pathway to mitigate this bias by instruction, thus benefiting LLMs in locating and utilizing relevant information from retrieved documents in RAG applications.

\end{abstract}

\section{Introduction}

Retrieval-Augmented Generation (RAG) is an established method for enabling continuous knowledge updates \citep{wu2023ragtruth, gao2023retrieval, chu2024improve, lewis2020retrieval} and reducing hallucination \citep{ji2023survey,zhang2023siren} by leveraging in-context learning ability of pre-trained large language models (LLMs). Traditional RAG approaches involve retrieving documents, sorting them by relevance, chunking and adding the documents as part of the prompt of LLMs \citep{glass2022re2g, xu2023recomp}. However, recent research has discovered that increasing the number of documents in the context can actually introduce challenges for RAG, as it may distract the model and degrade performance \citep{weller2022defending, oh2023detrimental}, even when they contain accurate and relevant information \citep{sauchuk2022role}. 

\begin{figure}[t]
    \centering
    \includegraphics[width=\columnwidth]{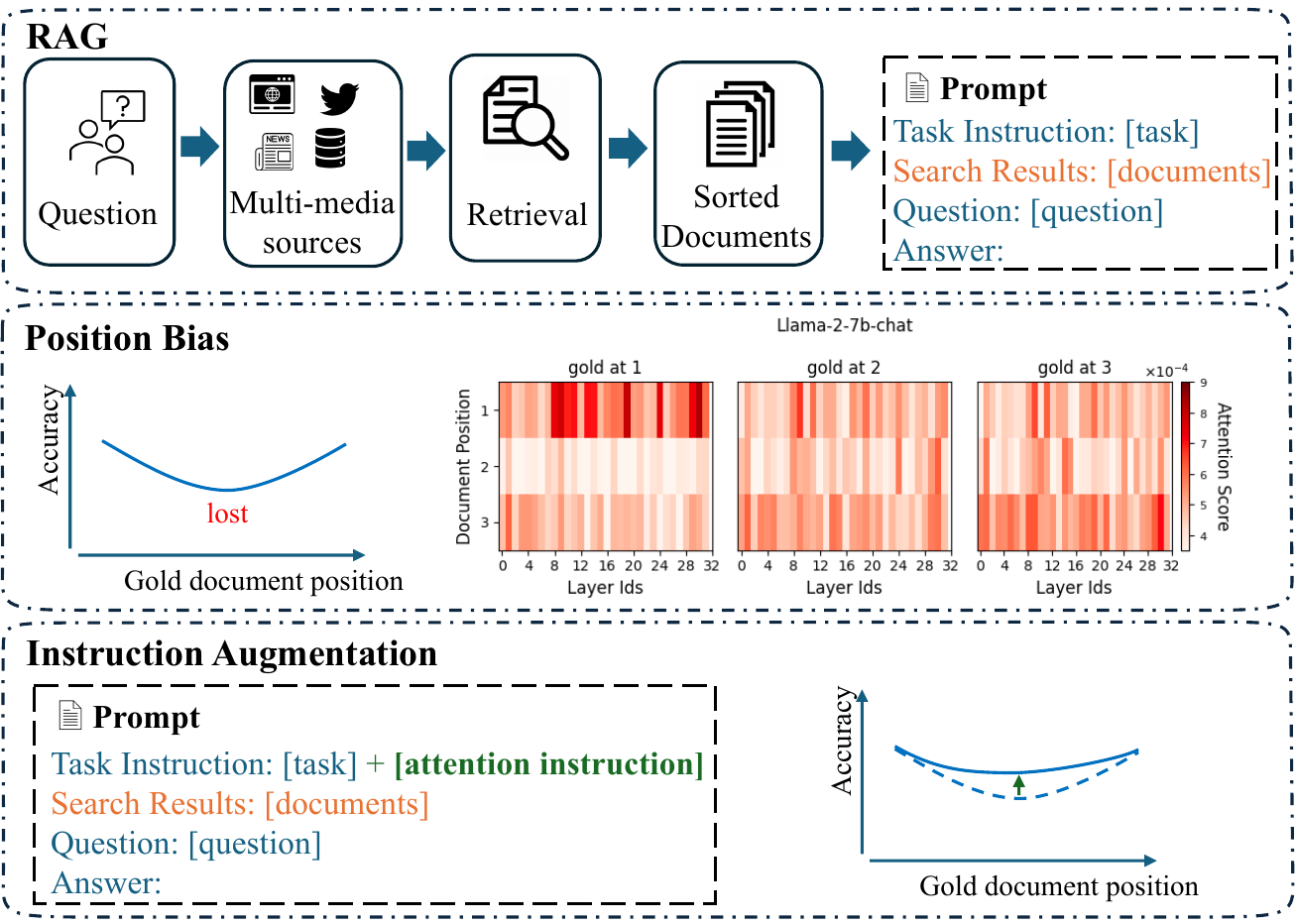}
    \caption{\textbf{Top:} An example of RAG for open question answering, where the prompt contains the sorted documents. \textbf{Middle:} The position bias (i.e. lost in the middle) can be visualized by attention score, which shows a significant drop in the middle wherever the gold answer is placed. \textbf{Bottom:} We solve this by augmenting the prompt with an attention instruction.}
    \label{fig:RAG_app}
    \vspace{-1.2em}
\end{figure}

Indeed, increasing evidence indicates that LLMs struggle to use context effectively due to the ordering and position of the informative content, a phenomenon known as position bias~\citep{xiao2023efficient, liu2023instruction, zheng2023large,qin2023nlp}. This bias causes models to favor the beginning or end text within the context~\cite{liu2024lost}, leading to the ``lost-in-the-middle'' problem, where LLMs have difficulty accessing relevant information in the middle of long contexts. \Cref{fig:RAG_app} illustrates this challenge in the RAG pipeline for the open question answering task, where multiple retrieved documents are added to the prompt, with the gold document containing the answer appearing at any position. Our attention score visualization for a 3-document question answering example aligns with previous works \cite{chen2023fortify, zhang2024found, he2024position}, showing that the second document always receives less attention, regardless of the position of the gold document. This bias can lead to incorrect answers when the gold document is in the middle.

To address position bias, many researchers have explored either finetuning or modifying position embeddings. For example, recent studies have explored finetuning with rich training data containing crucial information in the middle part ~\citep{fu2024data, an2024make}, or restructuring the generation task into multiple steps \citep{zhang2024position} to mitigate this bias. Other approaches have investigated the utilization of relative position embeddings \citep{su2024roformer}, which adjust the attention values by modifying the position embedding \cite{chen2023fortify, he2024position}. However, finetuning-based approaches lack adaptability and require additional computation, whereas embedding-based approaches require multiple rounds of inferencing or hyperparameter search, which is inefficient. 
 
Instead of finetuning the models or adjusting the relative position embeddings, we focus on addressing the position bias using the instruction following ability of LLMs \citep{machlab2024llm}. We investigate how to instruct LLMs to pay more attention to the specific positions within the context, thereby compensating for position bias. In particular, 
we design two types of \texttt{attention instructions} that \textit{instruct} LLMs to adjust their attention using either relative position words or absolute document indexes. We conduct comprehensive experiments with these two types of attention instructions on five open-sourced LLMs based on the multi-document question-answering task. Our investigation focuses on the relative position awareness of LLMs, their sensitivity to attention-following instructions in different prompt settings, and the feasibility of mitigating position bias through attention instructions. In summary, our findings are as follows:
\begin{itemize}
    \item Our experimental results indicate that language models lack an understanding of positional concepts and therefore fail to follow the relative attention instruction. However, we found that there exists latent correlations between position words and document ID.
    \item Our investigation on absolute attention instruction shows evidence that the attention of LLMs to a segment within the context can be enhanced semantically. 
    \item We illustrate that relative regional attention control can be achieved by attaching the same index to multiple documents.
\end{itemize}

The rest of the paper is organized as follows: \Cref{sec:experimetal_setup} describes the experimental setup in detail. \Cref{sec:RQ1} presents the results of applying relative attention instruction, while \Cref{sec:RQ2} discusses the use of absolute attention instruction with document ID indexes. \Cref{sec:RQ3} explores replacing document ID with position words to achieve regional control. We provide a detailed discussion of related works in \Cref{sec:related}. Finally, we conclude the paper and discuss potential future research directions in \Cref{sec:conclusion}.

\section{Experimental Setup}
\label{sec:experimetal_setup}
\subsection{Overview}
\label{sec:overview}
To test the effectiveness of instructing the models to increase attention on different segments of the search results, we designed a series of experiments on the multi-document question answering (MDQA) task \cite{singh2021end} under the setting that only one document contains the gold answer, namely the gold document. Given that the gold document can appear at any position and be overlooked because of the position bias, we design attention instructions, which is a two-sentence instruction, to guide LLMs to focus on a selected segment therefore prevent overlooking of information. 

\begin{figure*}[t]
    \centering
    \includegraphics[width=0.95\linewidth]{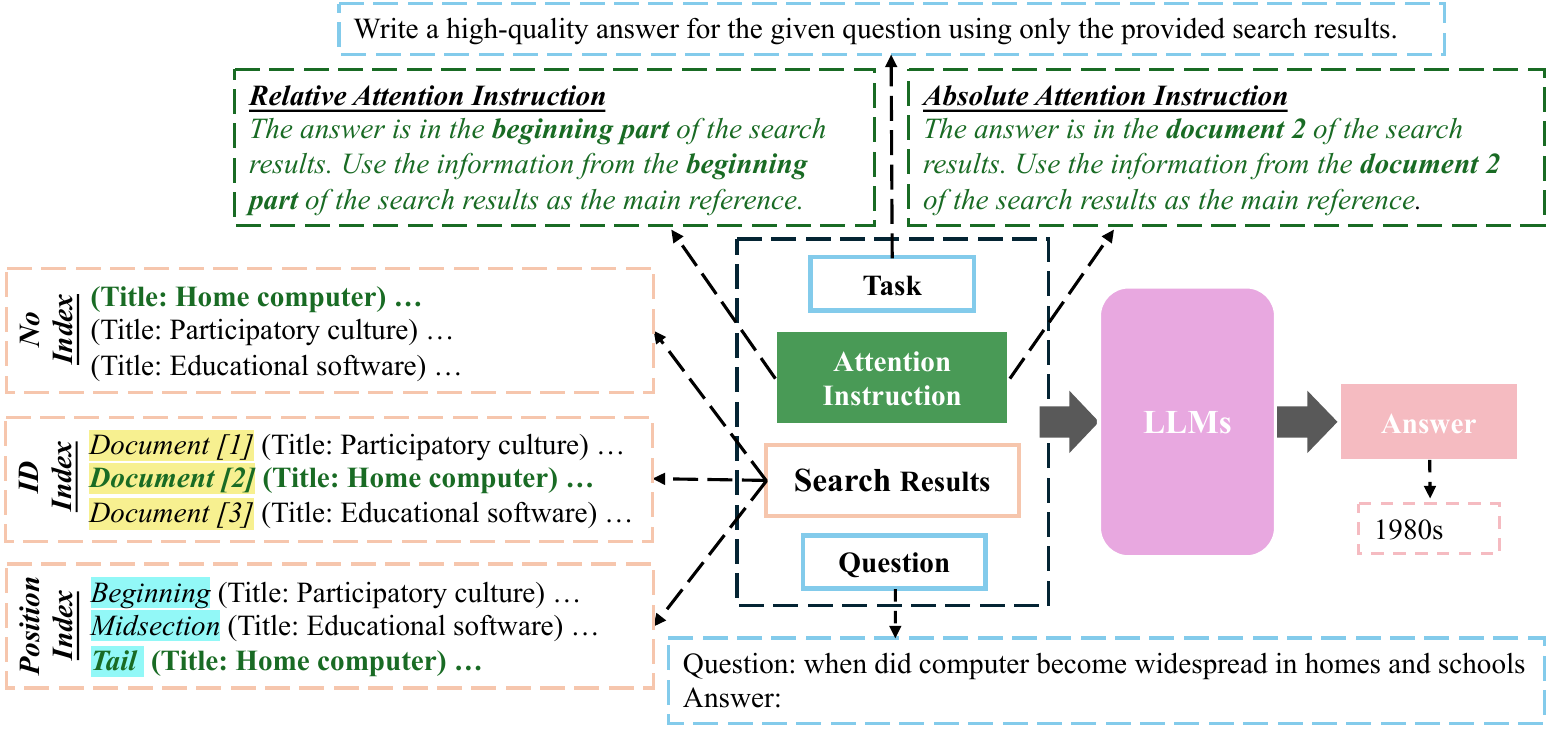}
    \caption{Prompt structure. The prompt is structured in four parts: the MDQA task instruction, the attention instruction, the search results containing the provided documents, and the question to answer. 
    The top two \textcolor{my-green}{$\Box$} boxes show the two types of \textcolor{my-green}{attention instructions}, where the attention segment phrase is marked in \textcolor{my-green}{\textbf{\textit{bold}}}. Three index types for documents (highlighted in \colorbox{my-yellow}{\textit{ }} for ID-index and \colorbox{my-blue}{\textit{ }} for position-index) are shown in the left \textcolor{my-amber}{$\Box$} boxes, with the \textcolor{my-green}{\textbf{gold document}} shown in different positions.}
    \label{fig:main_figure}
\end{figure*}

An overview of the input prompt and some toy examples can be seen in \Cref{fig:main_figure}. The prompt contains four parts: the task instruction describing the MDQA task, the attention instruction added after the instruction to guide the models' attention, the search results containing the provided documents with optional indexes and the question of the task. The position of the gold document is named the ``gold document position''. By controlling the gold document position and attention segment specified in the instructions, we aim to evaluate the LLMs' ability to follow attention instructions accurately. We test different experiment settings resulting from the combination of attention instruction type and index type to investigate the effectiveness of attention instructions. A detailed explanation of the attention instructions and document indexes will be provided in \Cref{sec:attention_instruction} and \Cref{sec:format}, respectively.

\paragraph{Dataset} 
We use the dataset from \citet{liu2024lost}, which contains 2,655 question-answer-gold document triplets from NaturalQuestion-Open \citep{kwiatkowski2019natural} and $n-1$  distractor documents that are relevant but do not contain the answer as. The distractor documents are retrieved using a retrieval system (Contriever, finetuned on MS-MARCO; \citep{izacard2021unsupervised}). To ensure consistency and control input length, all documents are chunked to a maximum of 100 tokens. In this study, we test scenarios with a total of 3 and 9 documents. The details are provided in \Cref{tab:gold_position}. Following \citet{liu2024lost} and \citet{mallen2023not}, we use accuracy as the primary evaluation metric, considering an answer correct if the gold answer exists in the generated output. 

\begin{table}[h]
    \centering
    \begin{tabular}{p{0.3\columnwidth}p{0.15\columnwidth}p{0.15\columnwidth}p{0.15\columnwidth}}
        \# Documents & \multicolumn{3}{c}{Gold Document Position}\\
        \toprule
        3 & \ \ \ \ 1st  & \ \ 2nd  & \ \ 3rd  \\
        9 & \ \ \ \ 2nd  & \ \ 5th  & \ \ 8th  \\
    \end{tabular}
    \caption{Gold document positions we tested for guided multi-document question answering. For 9-document scenarios, we keep 3 documents as a subgroup and place the gold document at the middle of each subgroup.}
    \label{tab:gold_position}
\end{table}

\paragraph{Models} We experiment with five state-of-the-art open-sourced models that are instruction-tuned and have sizes between 7-8B to investigate their sensitivity to attention instructions. These models include Llama-2-chat \citep{touvron2023llama}, Llama-3 \citep{llama3}, Tulu-2 \citep{ivison2023camels}, Mistral-instruct-v0.1 and Mistral-instruct-v0.2 \citep{jiang2023mistral}. Due to limited space, we primarily include the results of Llama-2-chat, Llama-3, and Mistral-instruct-v0.2 in the main paper (All results can be found in \Cref{sec:appendix}).

\subsection{Attention Instruction}
\label{sec:attention_instruction}
The attention instruction is a two-sentence instruction that aims to guide the model to focus on a positional segment of the search results. Hereafter we refer to the phrase representing the position of segment in instructions as ``attention segment phrase''. The first sentence explicitly informs the model where the answer is located, while the second sentence directs the model to use that segment as the main reference when answering the question. To investigate the effectiveness of attention instructions in mitigating position bias, we explore ``relative attention instruction'' and ``absolute attention instruction''. The details are as follows:
\begin{itemize}
    \item \textbf{Relative Attention Instruction}: We use the phrase \texttt{``{position word} part''} to guide the model's focus on a positional segment of the search results. The position words \textit{beginning}, \textit{midsection}, and \textit{tail} are used to virtually split the search results into three parts. An example is shown in the top left of \Cref{fig:main_figure}.
    \item \textbf{Absolute Attention Instruction}: We use the document indexes as the segment phrase in attention instruction. For ID-index, we use \texttt{``document [ID]''} (see the top right example in \Cref{fig:main_figure}). For the position-index, we directly use the position words as the attention segment phrase. A 9-document example of the position-index setting is shown in \Cref{fig:pos_token} in \Cref{sec:appendix}, in which the gold document is placed at the middle of each 3-document subgroup, and the position word is repeated for each subgroup.
\end{itemize}

\subsection{Indexing Documents in Search Results}
\label{sec:format}
This work focuses on the MDQA task under the RAG setting, where search results comprising multiple documents are given. As described in \Cref{sec:overview}, only one document contains the gold answer and locating this gold document from the context documents is crucial to generating accurate answers for the MDQA task. To aid this process, we can add an optional index to each of the documents that serves as a reference for index-based attention instructions. There are three types of indexes:

\begin{itemize}
    \item \textit{No-index}: No index is added to the front of each document.
    \item \textit{ID-index}: The relative ID (e.g., 1, 2, 3) is added as before the document passages, written as \textit{Document [ID]}.
    \item \textit{Position-index}: The relative position of the document within the search results is used as an index (e.g., \textit{beginning}, \textit{midsection}, or \textit{tail}).
\end{itemize}
\Cref{fig:main_figure} shows a toy example of these index types.

\section{Can LLMs follow relative attention instructions?}
\label{sec:RQ1}

\begin{figure*}[!htbp]
    \centering
    \begin{minipage}[b]{0.3\linewidth}
        \centering
        \includegraphics[width=\columnwidth]{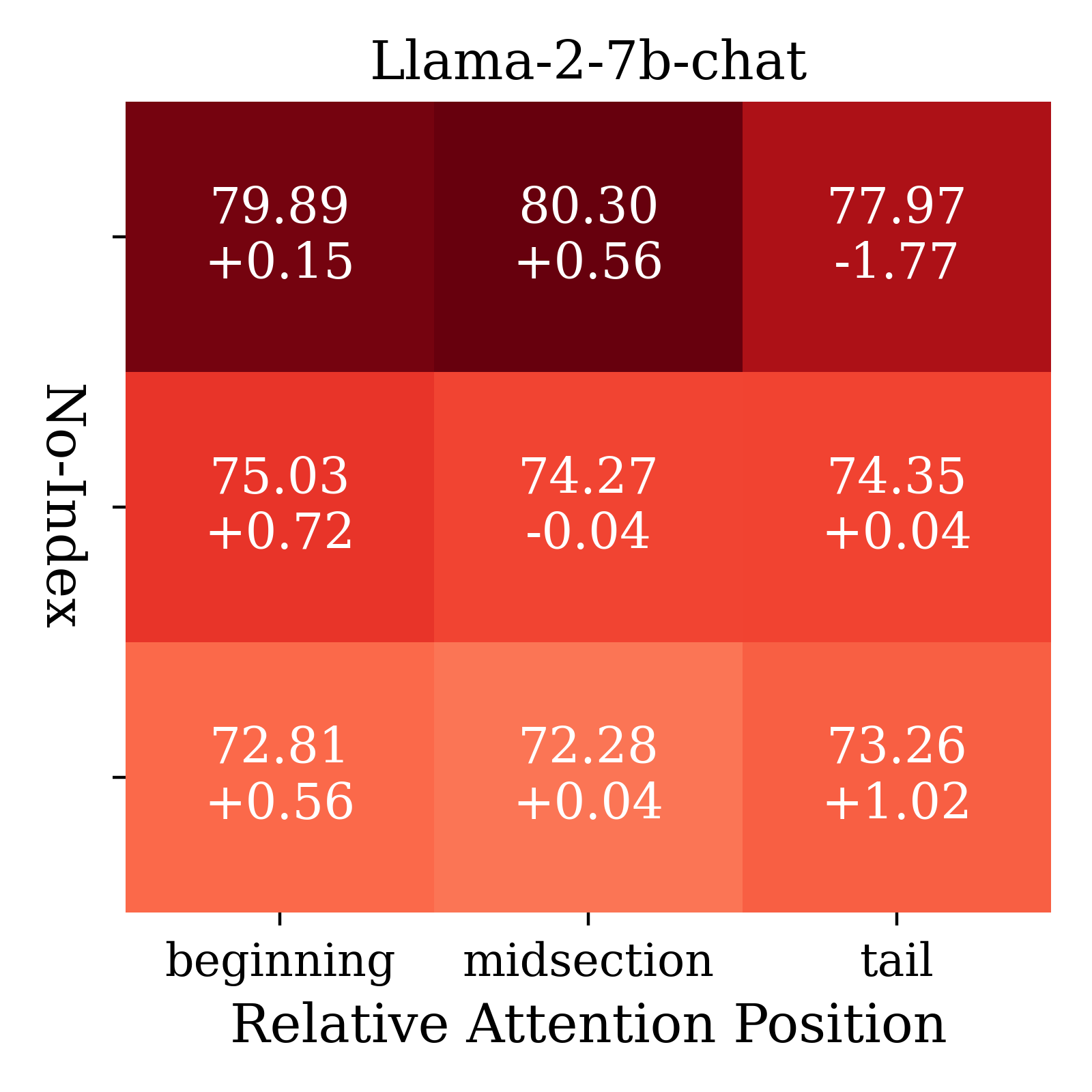}
    \end{minipage}
    \begin{minipage}[b]{0.3\linewidth}
        \centering
        \includegraphics[width=\columnwidth]{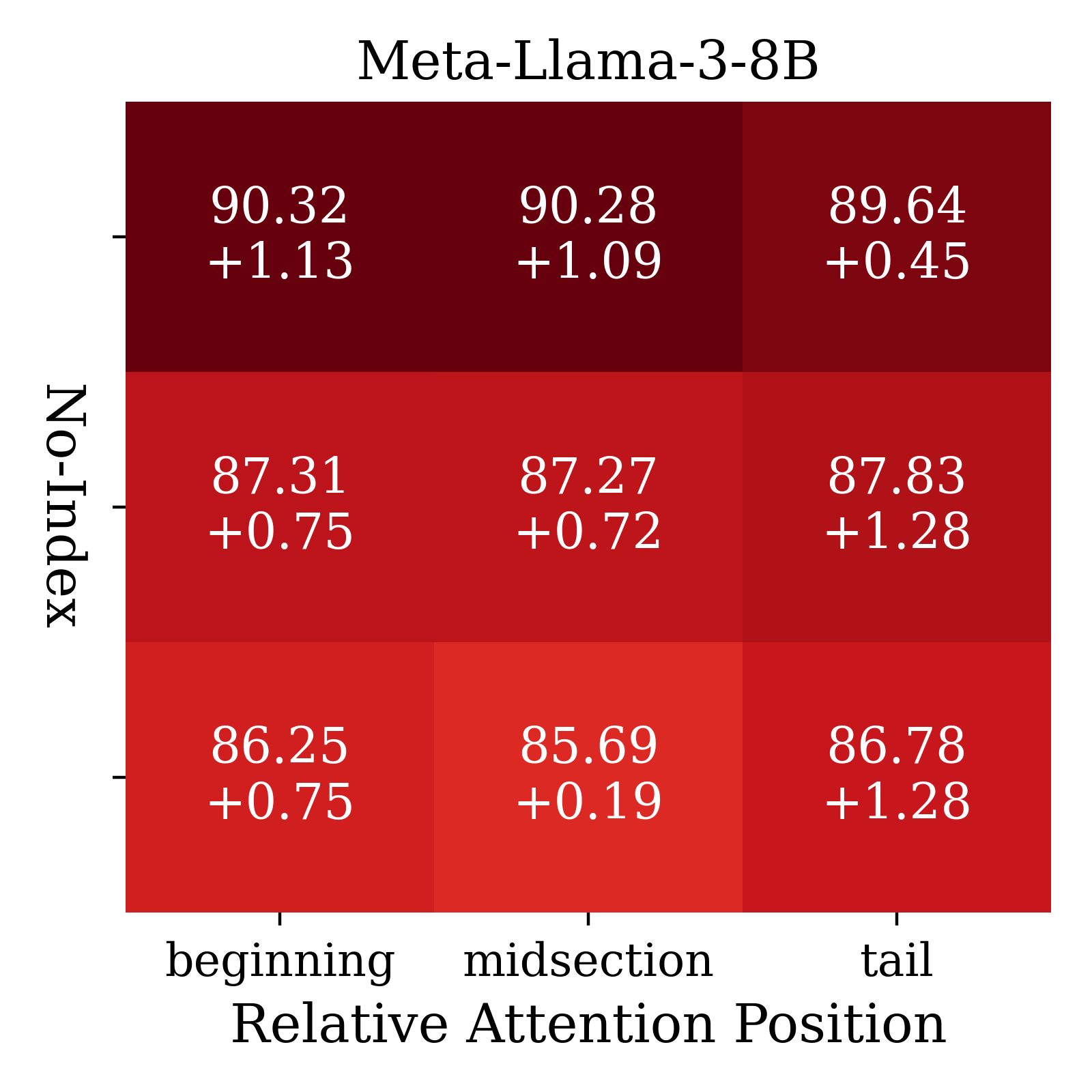}
    \end{minipage}
    \begin{minipage}[b]{0.3\linewidth}
        \centering
        \includegraphics[width=\columnwidth]{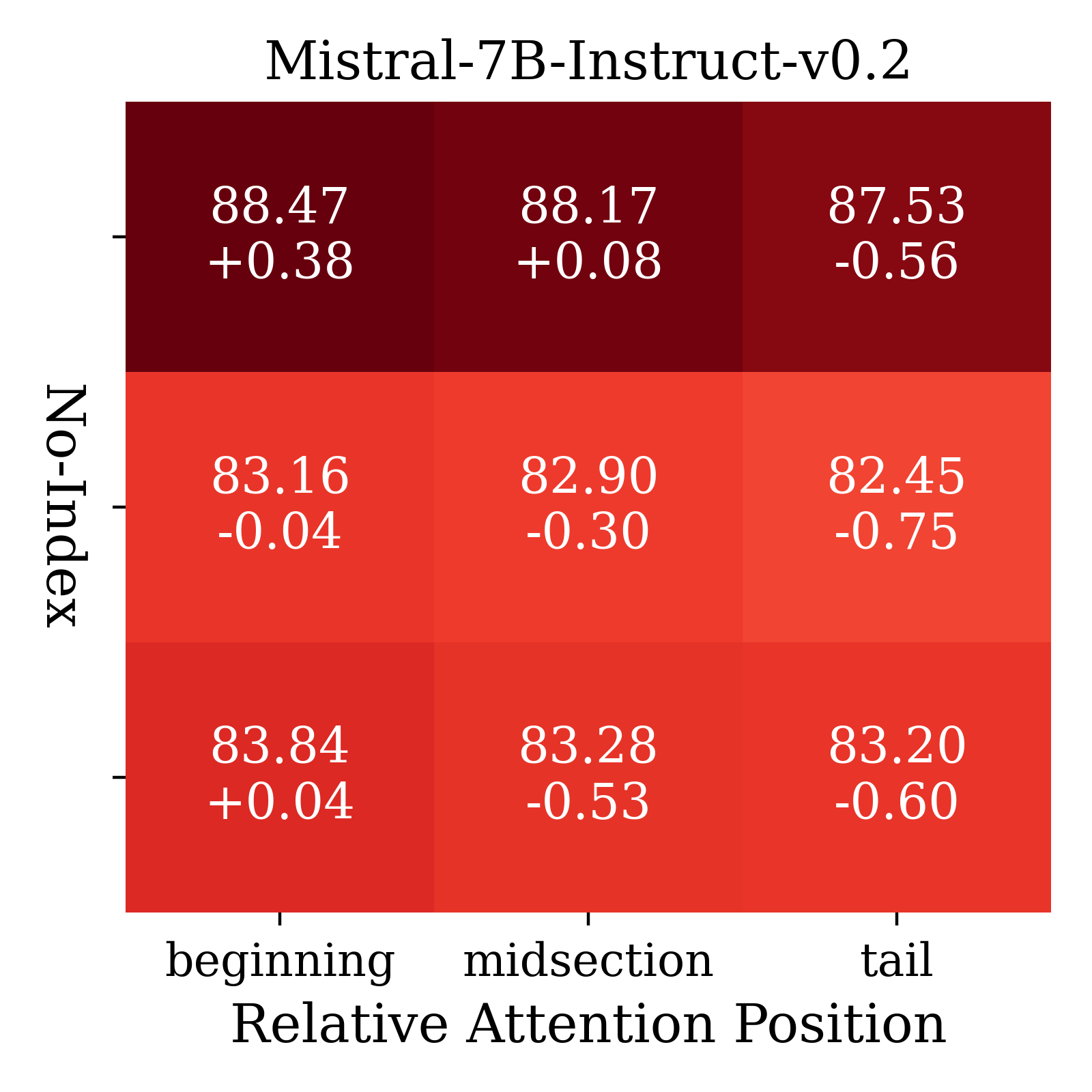}
    \end{minipage}
    \begin{minipage}[b]{0.3\linewidth}
        \centering
        \includegraphics[width=\columnwidth]{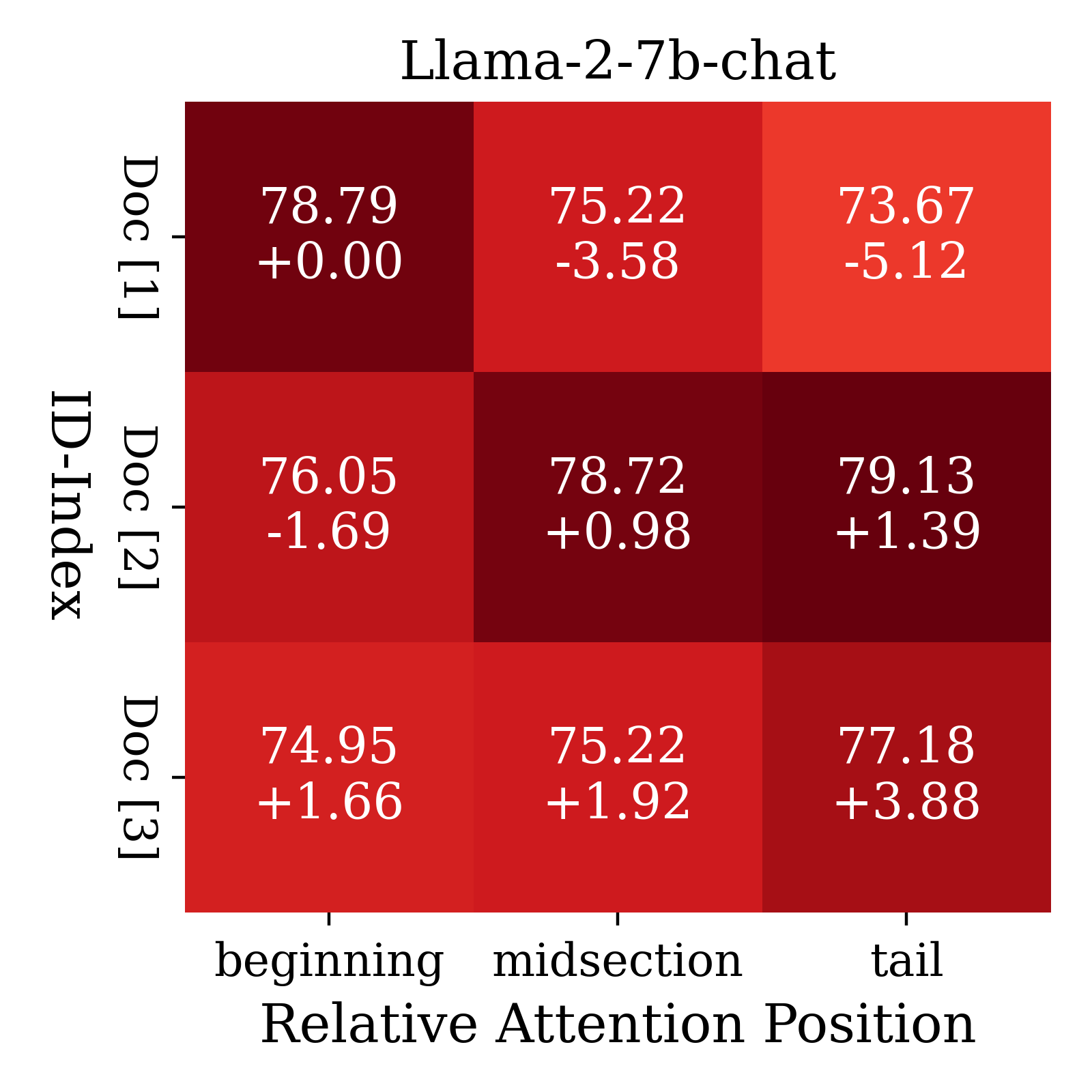}
    \end{minipage}
    \begin{minipage}[b]{0.3\linewidth}
        \centering
        \includegraphics[width=\columnwidth]{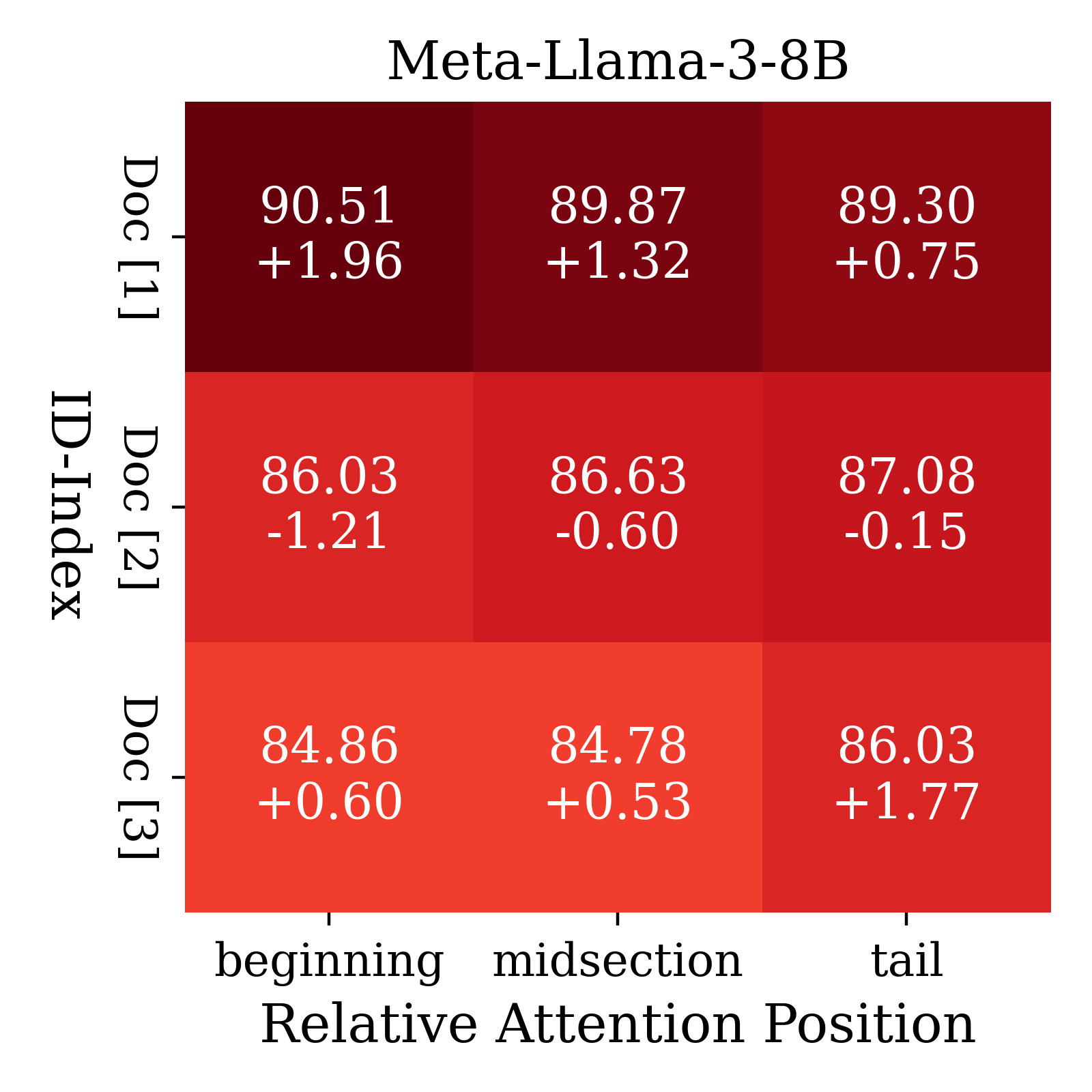}
    \end{minipage}
    \begin{minipage}[b]{0.3\linewidth}
        \centering
        \includegraphics[width=\columnwidth]{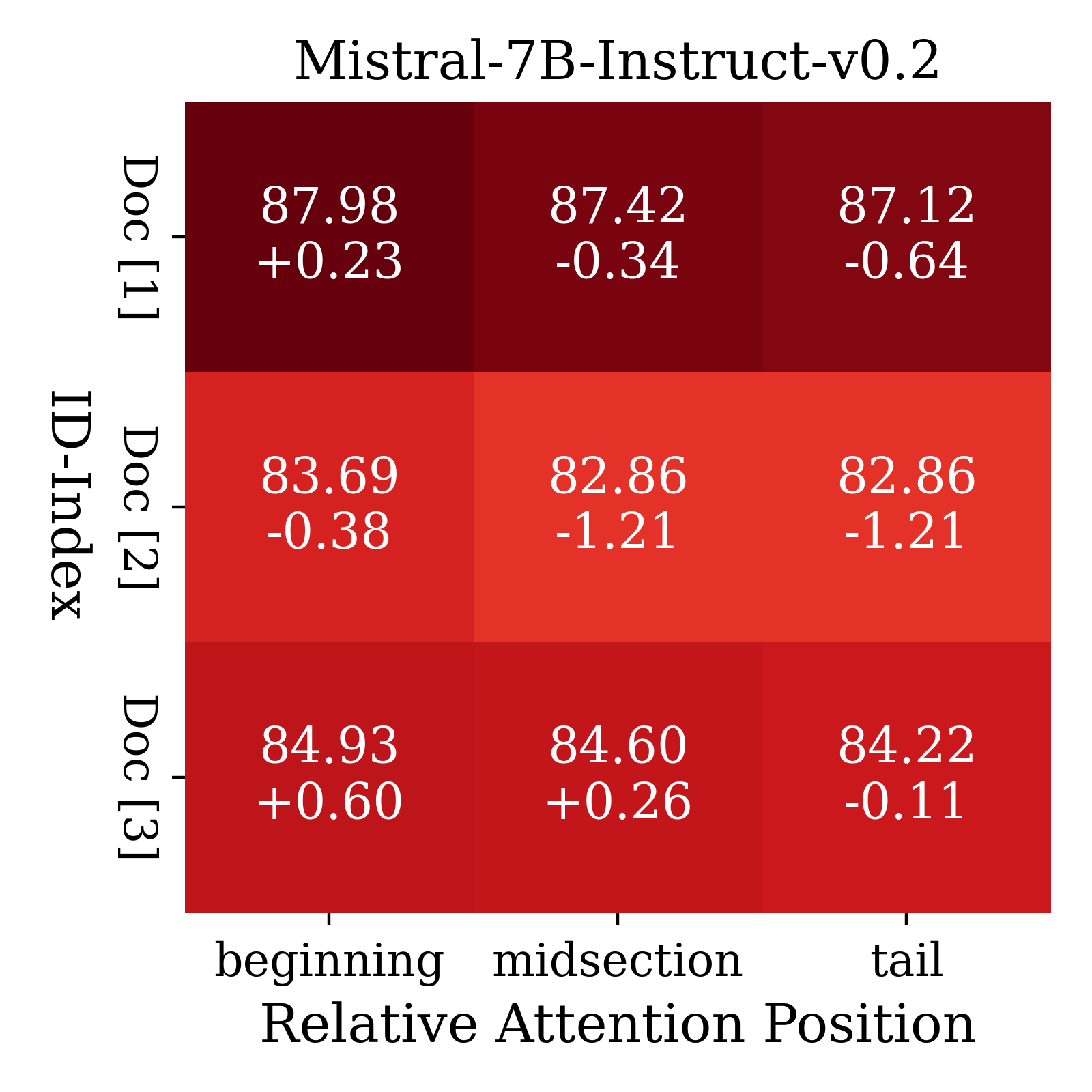}
    \end{minipage}
    \caption{Accuracy heatmaps of Llama-2-chat, Llama-3 and Mistral-Instruct-v0.2 when using relative attention instruction. \textbf{Top Row}: results in no-index setting. \textbf{Bottom Row}: results when using ascending ids as document index. In each cell of the heatmaps, the accuracy value is shown in \% and the $+-$ indicates the performance difference compared to without using attention instruction. The darker the color of the cell, the higher the accuracy. }
    \label{fig:relative_position}
\end{figure*}

As described in \Cref{sec:attention_instruction}, we virtually split the search results into three parts and represent the relative position with the words \textit{beginning}, \textit{midsection}, and \textit{tail}. For LLMs to understand and follow the relative attention instruction, they will need to have the relative position awareness and map the instruction to higher attention scores on the segment mentioned in the instruction. 

\paragraph{Setting} We test the performance of the models using relative attention instruction in no-index and ID-index settings. We place the gold document in the pre-defined positions under 3-document and 9-document settings as described in \Cref{tab:gold_position}. To assess the LLMs' ability to follow fine-grained attention instructions, we create a 3x3 accuracy heatmap for each model in one combination of index types and attention instruction types. The heatmaps' y-axis represents the gold document position, while the x-axis represents the selected attention segment. It is worth noting that diagonal cells in the heatmap reflect instances where attention segments align with the positions of the gold documents. The improvements in these diagonal cells indicate the ability of LLMs to follow the attention instructions effectively. Conversely, we anticipate a decrease in performance in the non-diagonal cells as the model is directed to utilize information from the distractor documents. Both these will result in a deeper color along the diagonal, a phenomenon we refer to as the ``diagonal effect''.

\paragraph{Results} The accuracy heatmaps of Llama-2-chat, Llama-3, and Mistral-Instruct-v0.2 are shown in \Cref{fig:relative_position}. In general, most of the top rows of each heatmap perform the best compared to the other rows, which is consistent with the findings of \citet{liu2024lost}. The top three heatmaps demonstrate the performance when no index is used, from which we can observe little effect of the relative attention instruction on all three models. In particular, Llama-3 outperforms Llama-2-chat by ~10+\% but still experiences position bias, with the performance at the beginning position being ~3\% higher than the midsection and tail. Mistral-Instruct-v0.2 is better than Llama-2 but not as competitive as Llama-3, and the position bias results in a 5\% drop in accuracy when the gold document is placed at midsection and tail part. The bottom row shows the results with an ID index, revealing no significant improvement observed in diagonal cells across different gold positions on any of the three compared models.

\paragraph{Discussion} The absence of significant differences after using relative attention instructions suggests a lack of relative position awareness among LLMs. However, when using the ID index, there is a slight improvement in diagonal cells observed in Llama-2-chat, indicating a potential correlation between the numerical ID indexes and relative positions. This result suggests that Llama-2-chat's sensitivity to the attention instruction may be attributed to its lower performance compared to Llama-3 and Mistral-7b in reasoning over the entire context, leaving more room for improvement. The same trend can be observed on Tulu-2 and Mistral-Instruct-v0.1 and in 9-document settings as well, these results support the finding that LLMs do not have relative position awareness therefore cannot follow relative attention instruction (in \Cref{appx:relative_no_index} and \Cref{appx:relative_docid}).

\section{Can we instruct LLMs to attend to a document using absolute attention instruction?}
\label{sec:RQ2}
We have shown that LLMs do not understand the relative position of documents in search results. However, there is a correlation between document IDs and position words. In this section, we test LLMs' attention-following capability using absolute attention instructions, in which the ID of each document is used as an index to locate the segments of the search results.

\paragraph{Setting} The documents are arranged in the same order as in \Cref{sec:RQ1}, with relative ID as the index for each document in the search results. We also conduct experiments with document IDs in reversed order settings to consolidate the finding that the performance change is caused by the absolute attention instructions (see \Cref{appx:absolute_reverse_docid}). To investigate if attention instructions influence the attention score distribution, we visualize the average attention scores of the last token across all heads. We split the input sequence into instruction, documents, and question prompt segments, calculating the mean weight for each segment. \Cref{sec:flatten_weight} describes the detailed algorithm for calculating the attention score of each segment.

\begin{figure}[h]
    \centering
    \begin{minipage}[b]{0.47\columnwidth}
        \centering
        \includegraphics[width=\columnwidth]{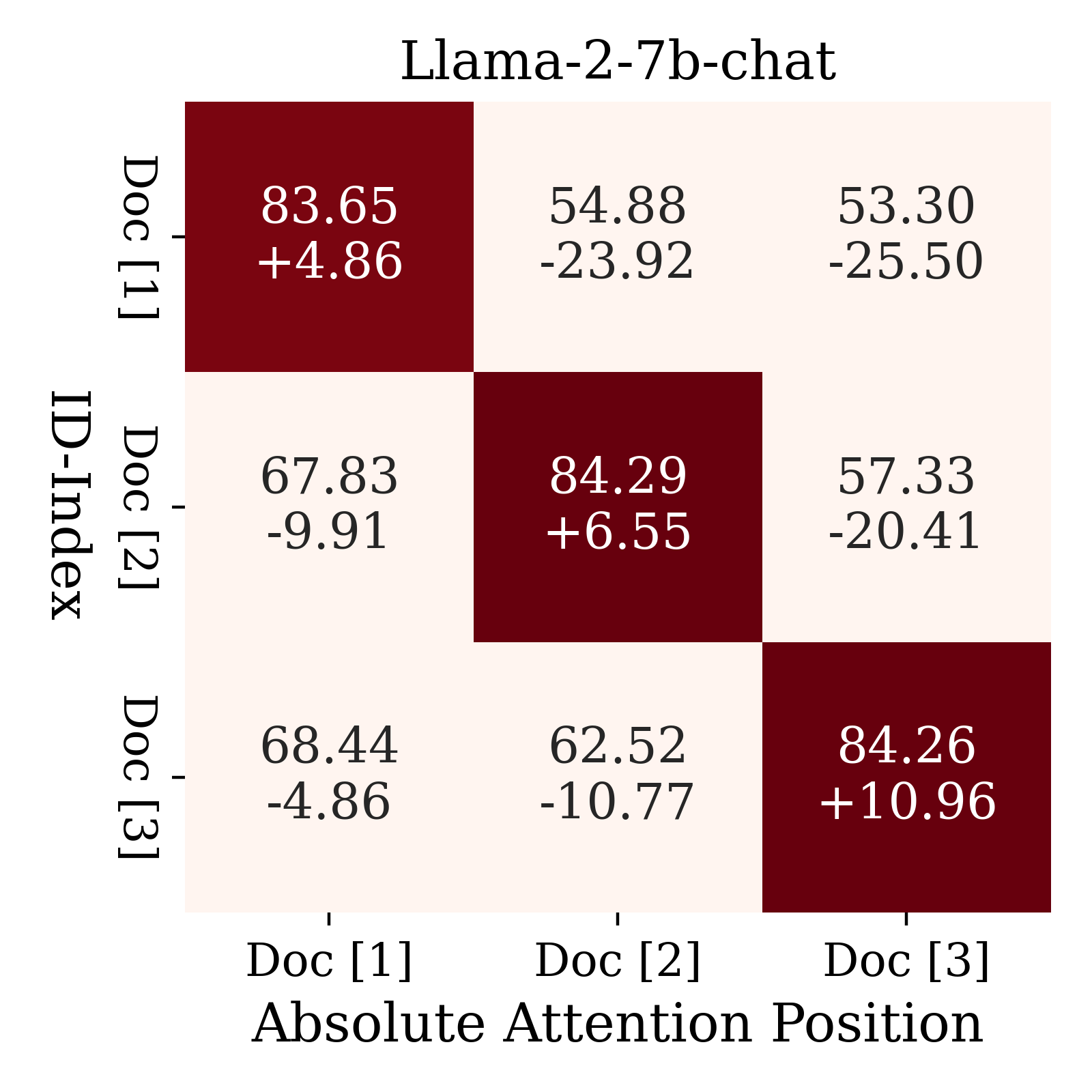}
    \end{minipage}
    \begin{minipage}[b]{0.47\columnwidth}
        \centering
        \includegraphics[width=\columnwidth]{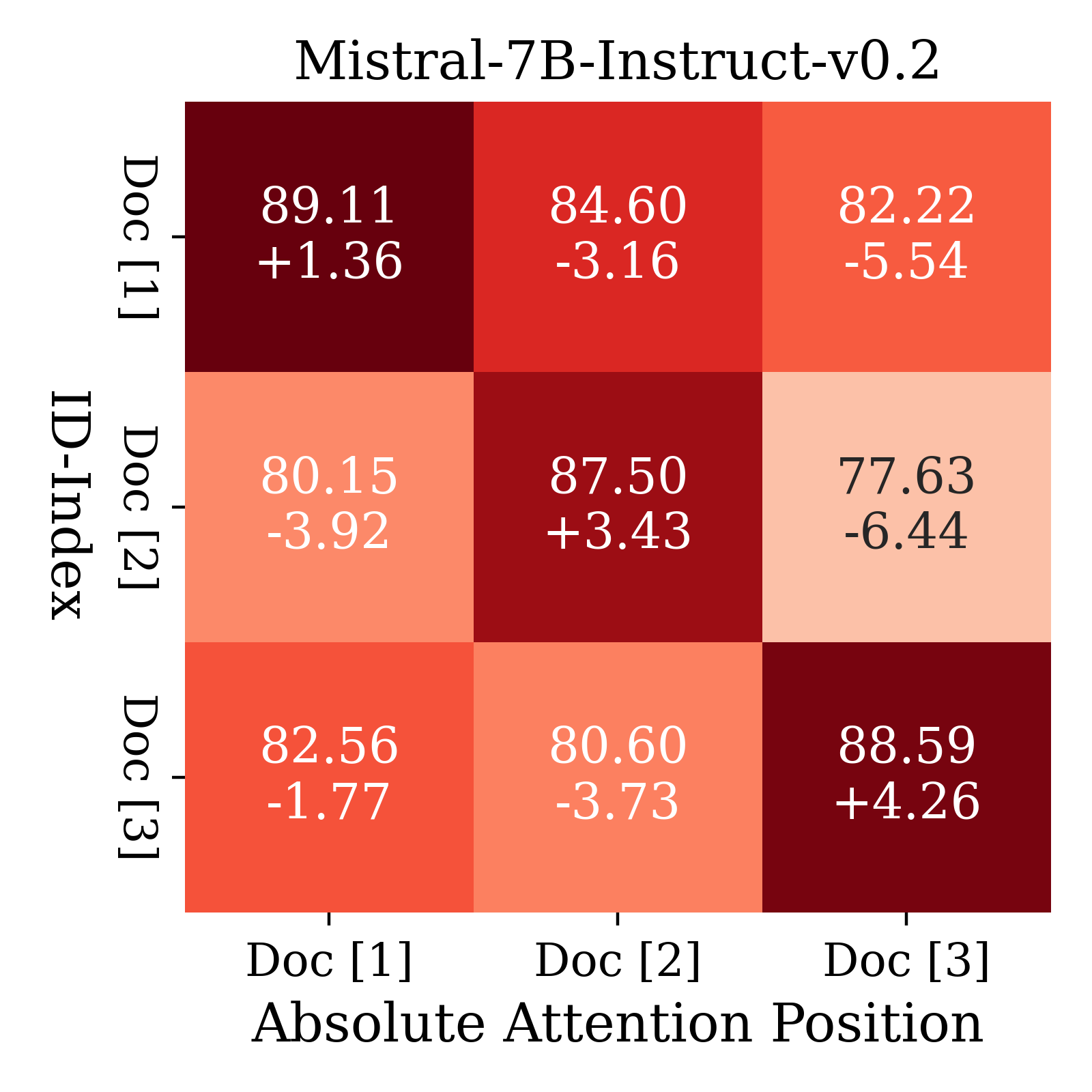}
    \end{minipage}
    \begin{minipage}[b]{0.47\columnwidth}
        \centering
        \includegraphics[width=\columnwidth]{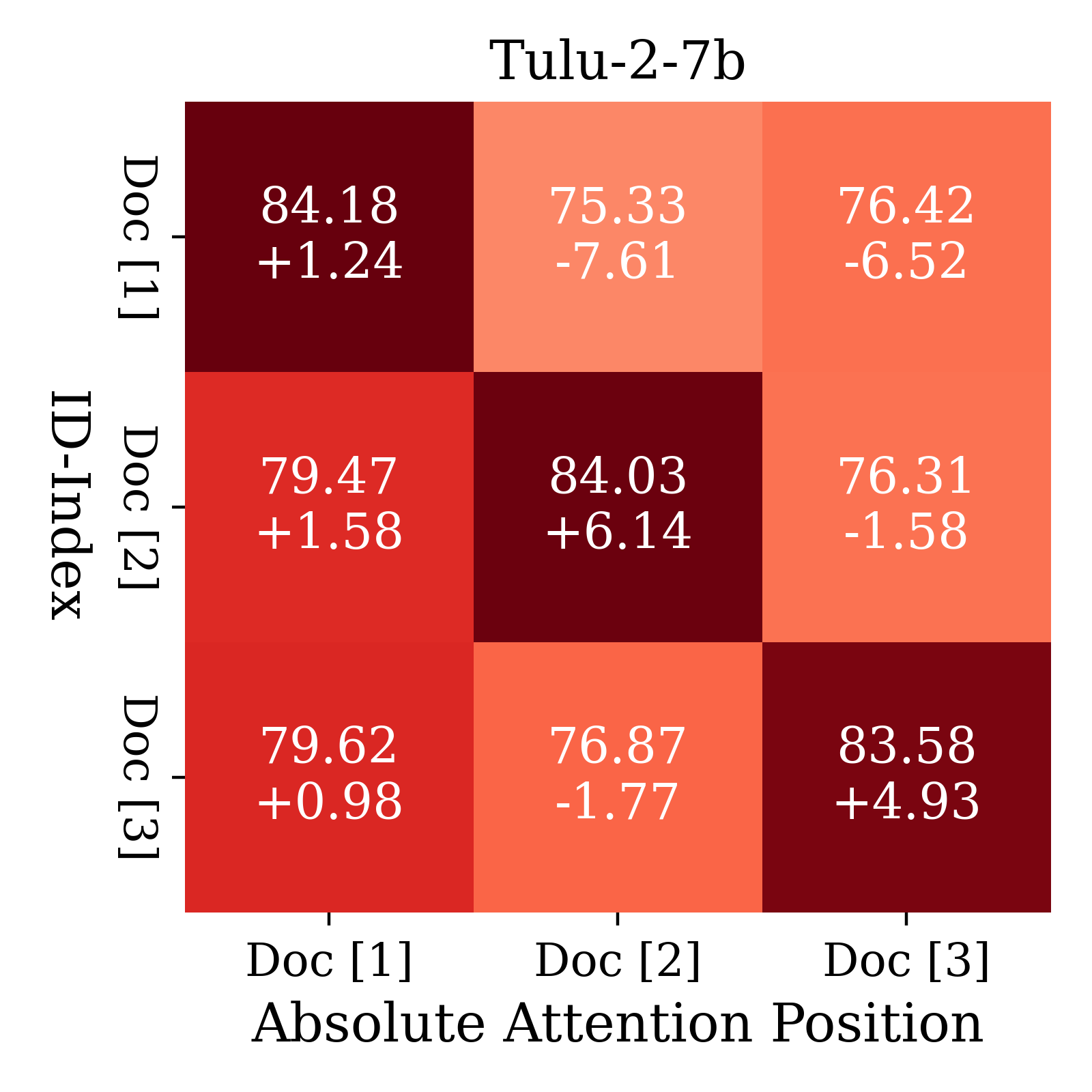}
    \end{minipage}
    \begin{minipage}[b]{0.47\columnwidth}
        \centering
        \includegraphics[width=\columnwidth]{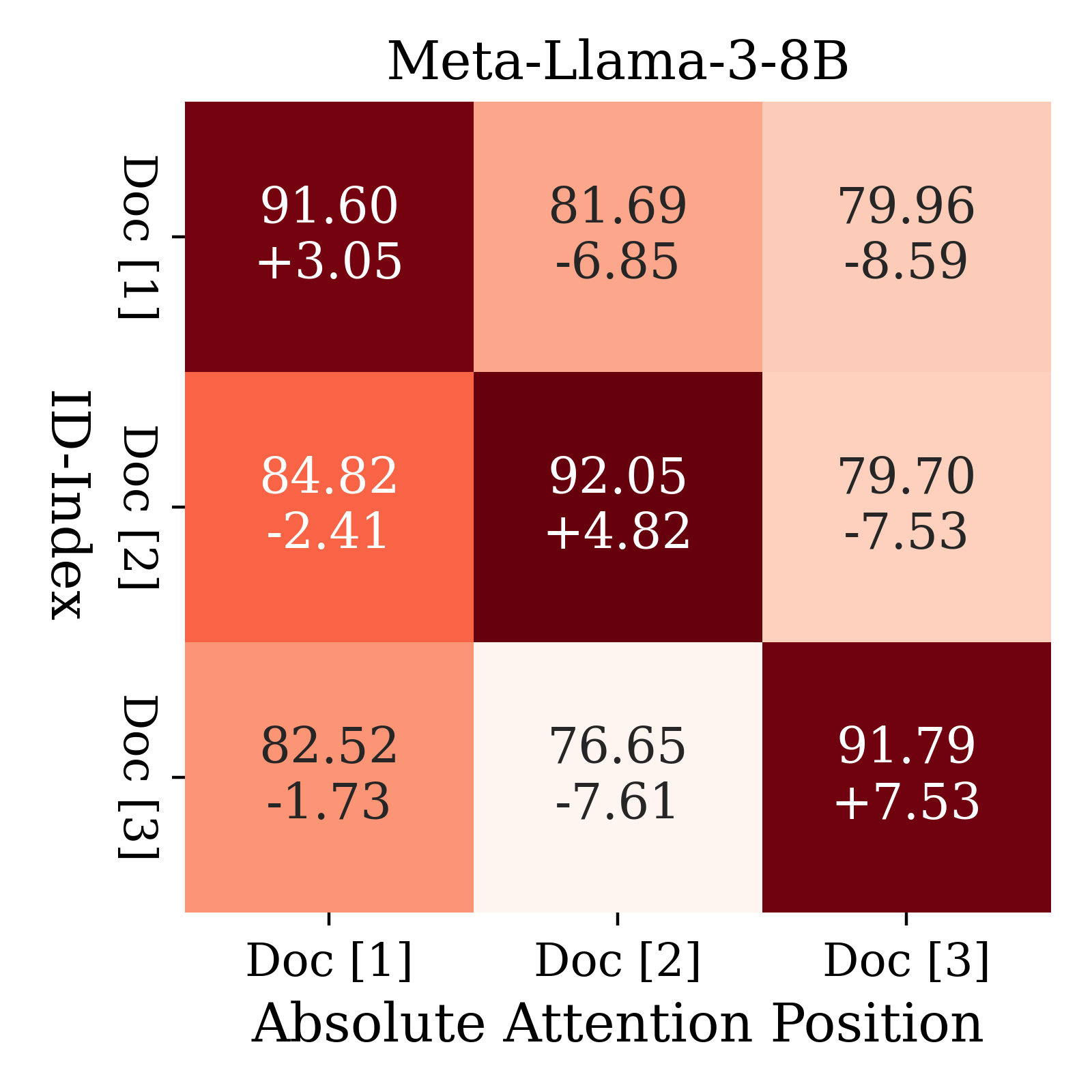}
    \end{minipage}
    \caption{Results of Llama-2-chat, Mistral-Instruct-v0.2, Tulu-2, Llama-3 using absolute attention instruction with relative numerical IDs as document indexes.}
    \label{fig:token_ascending_docid}
\end{figure}

\paragraph{Results} \Cref{fig:token_ascending_docid} presents the results of Llama-2-chat, Tulu-2, Mistral-v0.2, and Llama-3 using absolute attention instructions with ascending ID indexes. When the document ID is used as a reference, the models' reasoning is significantly affected, with boosted performance on the diagonals across all models, especially Llama-2-chat (4\% to 10\% $\uparrow$). Conversely, when the gold document is placed at a mismatched position, the performance drops significantly (e.g., 25\% $\downarrow$ for Llama-2-chat when the gold document is at the beginning). The heatmaps show that models perform similarly when the attention instruction guides them to focus on the gold document. Tulu-2 exhibits the largest relative improvement when the gold document is in the middle, indicating a more pronounced lost-in-the-middle problem compared to the other models, which are more affected by the primary bias. Interestingly, Tulu-2 also shows a unique behavior: when prompted to focus on document 1, its performance improves slightly regardless of the gold document position. For Mistral-Instruct-v0.2, the improvement in diagonal cells is less significant but still noticeable. While Mistral-Instruct-v0.2 outperforms Llama-2 in absolute accuracy, Llama-3 remains the strongest model overall. We also test the absolution of attention instruction using a 9-document setting with gold document positions as described in \Cref{tab:gold_position}. The results (\Cref{fig:all_token_have_docid_9docs} in \Cref{appx:absolute_docid}) further validate the generalized applicability of absolute attention instruction. Despite increasing number of distractor documents, referencing the exact document ID of the gold document boosts the model performance. This concludes that, by using absolute attention instructions, we can instruct LLMs to pay more attention to specific documents. 

\paragraph{Discussion} The diagonal effects observed in both 3-document and 9-document scenarios across all models demonstrate that absolute attention instruction can be used to mitigate the position bias in LLMs. The results of using absolute attention instructions align with the results in \Cref{sec:RQ1}, indicating that Llama-2-chat is more sensitive to attention instructions. Comparing the 3-document and 9-document results of Llama-2-chat and Mistral-Instruct-v0.2 reveals that the significance of attention instructions is also influenced by the document's relative position (e.g. beginning or tail). In contrast, the influence of attention instructions on Tulu-2 and Llama-3 is less sensitive to document position. Llama-3 exhibits better instruction-following ability than Mistral-Instruct-v0.2, despite having similar absolute accuracy, suggesting that its reasoning is more linearly influenced by the context. Tulu-2, a finetuned Llama-2 model, is less sensitive to absolute attention instruction and more robust when guides to attend to distractor documents compared to Llama-2, possibly due to its extended context window (from 4096 to 8192 tokens) and new data mixture used during finetuning. 

\begin{figure*}[t]
    \centering
    \includegraphics[width=\linewidth]{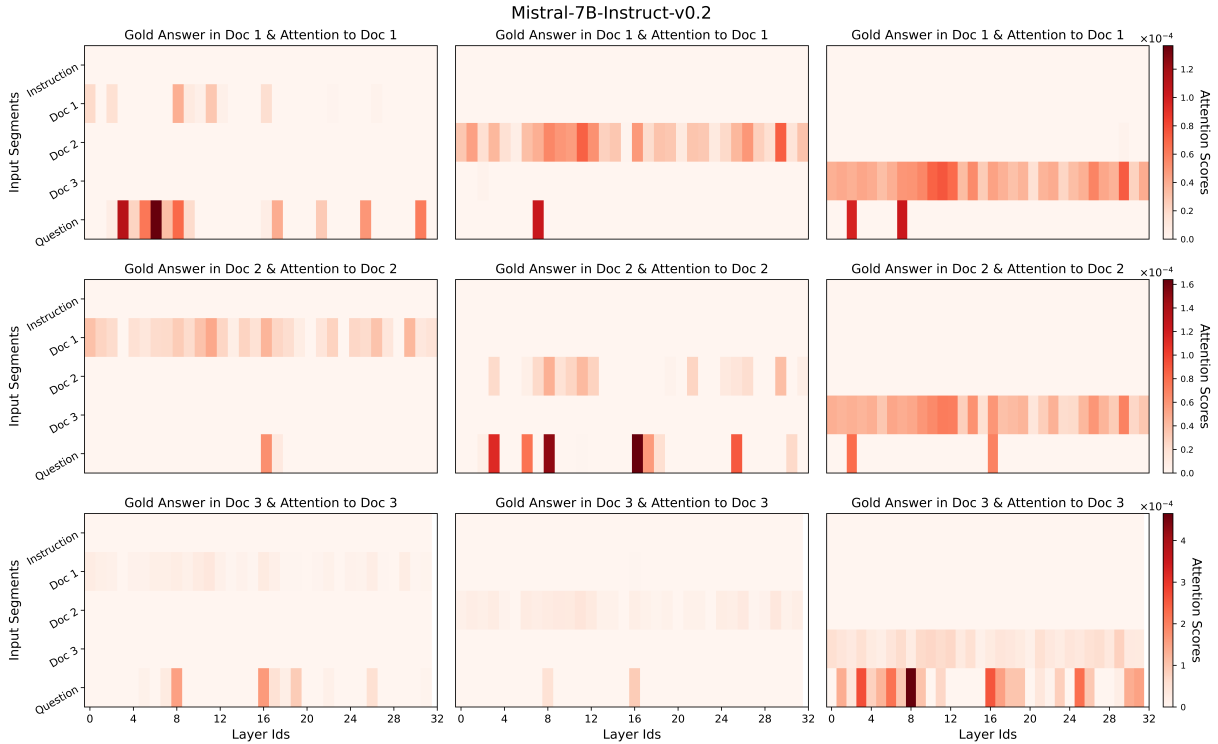}
    \caption{The attention score heatmaps of Mistral-Instruct-v0.2 using absolute attention instruction with document id index. Each subplot is a pair of gold document positions and attention segments in the same arrangement as the accuracy heatmap. The color bar starts with 0, those white areas may have reduced or unchanged attention scores.}
    \label{fig:weight_heatmap}
\end{figure*}

\begin{figure}[!htbp]
    \centering
    \includegraphics[width=0.9\columnwidth]{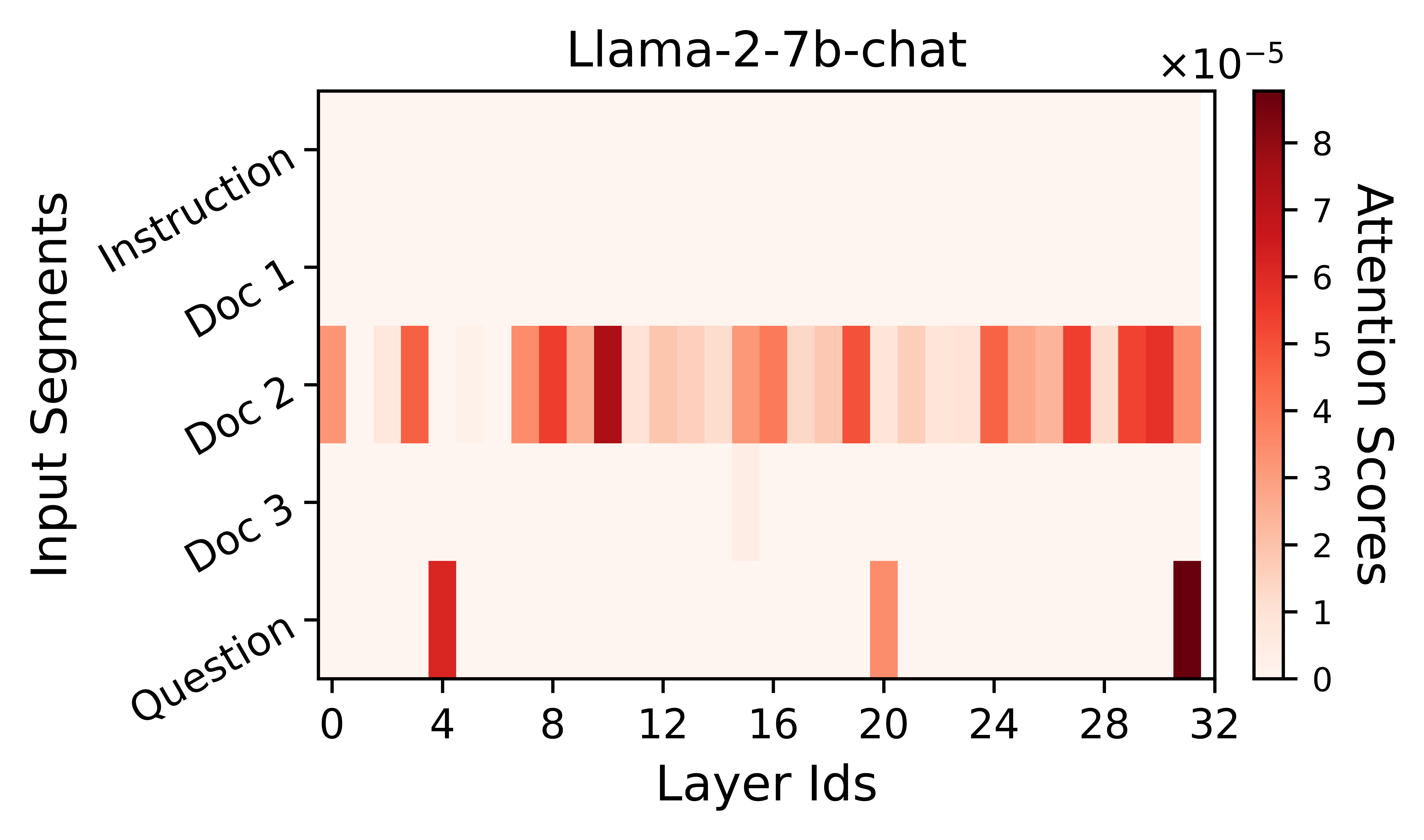}
    \caption{Case study: the attention score of an example that answers correctly after using attention instruction.}
    \label{fig:weight_heatmap_single}
\end{figure}

\paragraph{Attention Score} To verify that the attention scores of the selected segment change according to the instruction, we plot the attention scores for each segment, averaged over 10 random examples, in the same arrangement as the accuracy heatmaps. \Cref{fig:weight_heatmap} displays the attention scores of Mistral-Instruct-v0.2. When the model is asked to pay more attention to a specific document (e.g., document 1, 2, or 3) based on the document ID, there is an increase in the average attention score of the tokens in that document, regardless of the gold document position. Although the increment in attention scores for the selected segment gradually decreases as the gold document moves towards the tail, the attention scores still increase, while the attention scores for the unattended parts decrease or remain unchanged. When the attention segment matches the gold document position, the attention to the question also improves, indicating that the attention instruction encourages the model to think more about the question to find the answer. Comparing across layers, we observe that the front layers are more sensitive to absolute attention instructions.

\Cref{fig:weight_heatmap_single} presents an example where the model initially struggles to answer correctly without additional guidance but provides the correct answer after using an absolute attention instruction. In this example, the gold document is placed in the middle, and we use absolute attention instruction to guide the model to pay more attention to document 2. The increased attention scores on document 2 suggest that self-attention affects answer prediction and that guiding the language model through absolute attention instructions can help address challenging questions.

\section{Can LLMs follow absolute attention instruction with position-index and achieve regional control?}
\label{sec:RQ3}
Having shown that LLMs can follow attention instructions based on document ID indexes, we now explore whether relative position words (e.g. beginning, midsection, and tail) can be used as indexes to achieve regional control of the models' attention.

\paragraph{Setting} In this set of experiments, we replace the document ID with relative position words and refer to the position indexes in the attention instructions. For the 3-document setting, each document is assigned a unique position-index. In the 9-document setting, the three documents in each subgroup share the same position word, allowing the model to focus on three documents simultaneously. 

\begin{figure}[h]
    \centering
    \begin{minipage}[b]{0.47\columnwidth}
        \centering
        \includegraphics[width=\columnwidth]{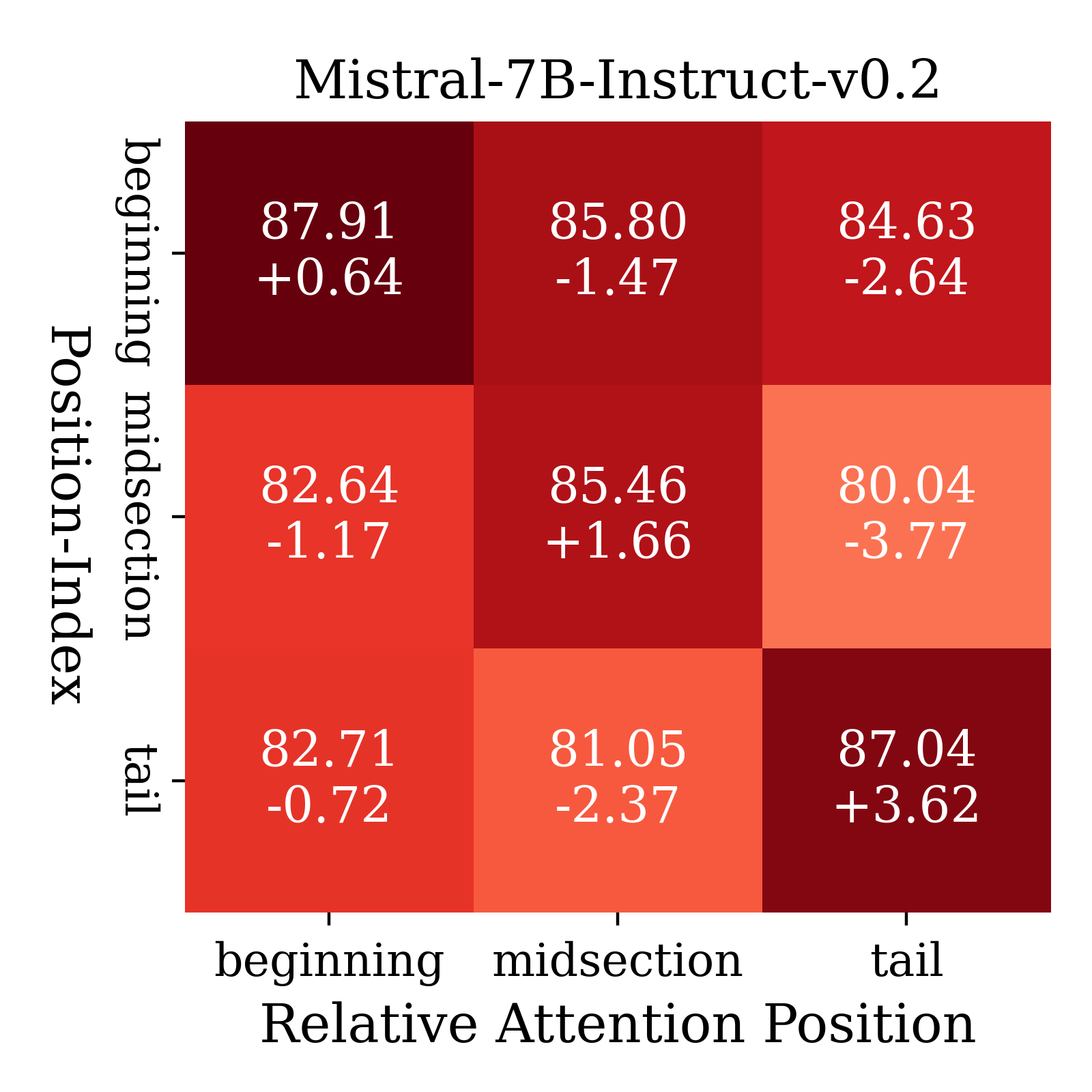}
    \end{minipage}
    \begin{minipage}[b]{0.47\columnwidth}
        \centering
        \includegraphics[width=\columnwidth]{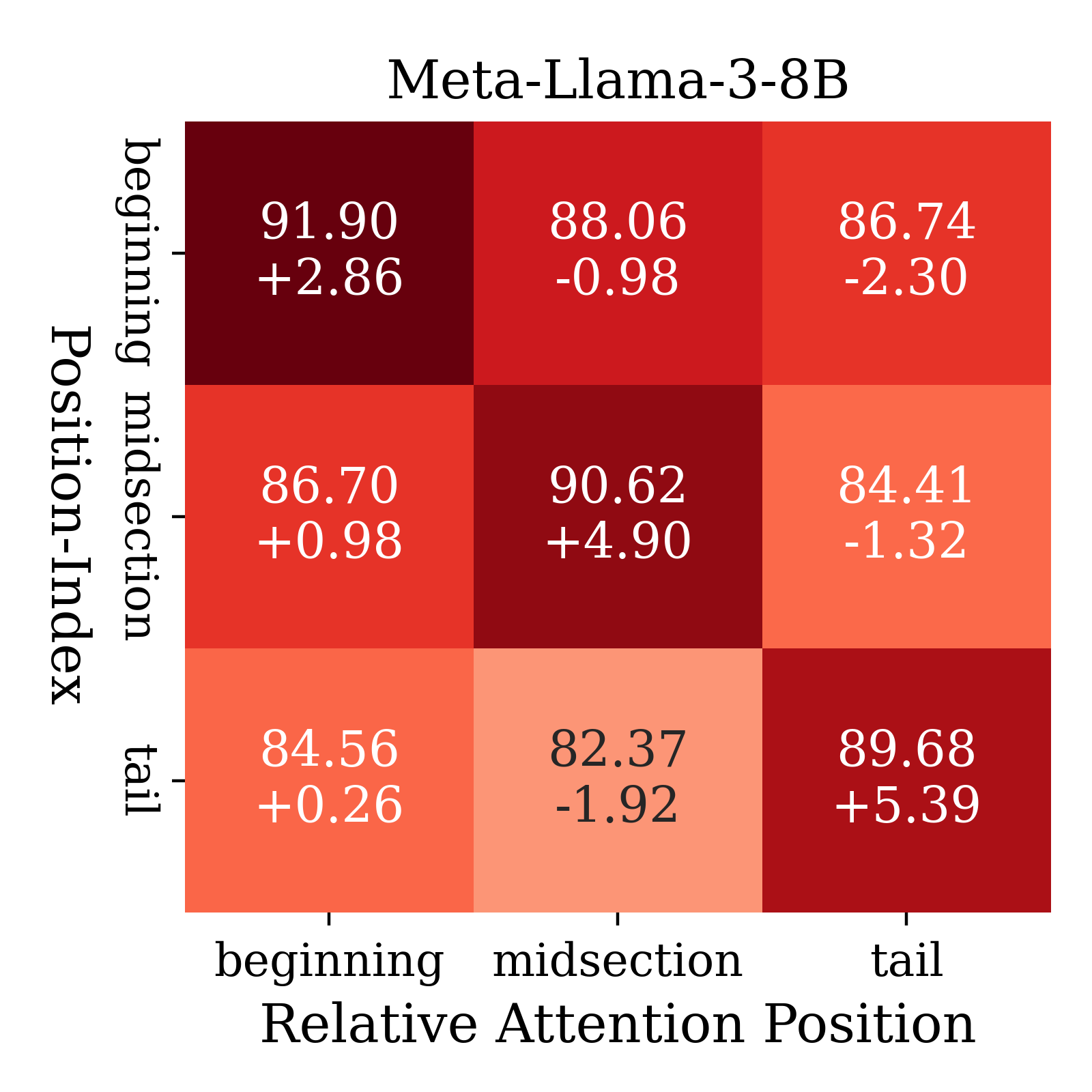}
    \end{minipage}
    \begin{minipage}[b]{0.47\columnwidth}
        \centering
        \includegraphics[width=\columnwidth]{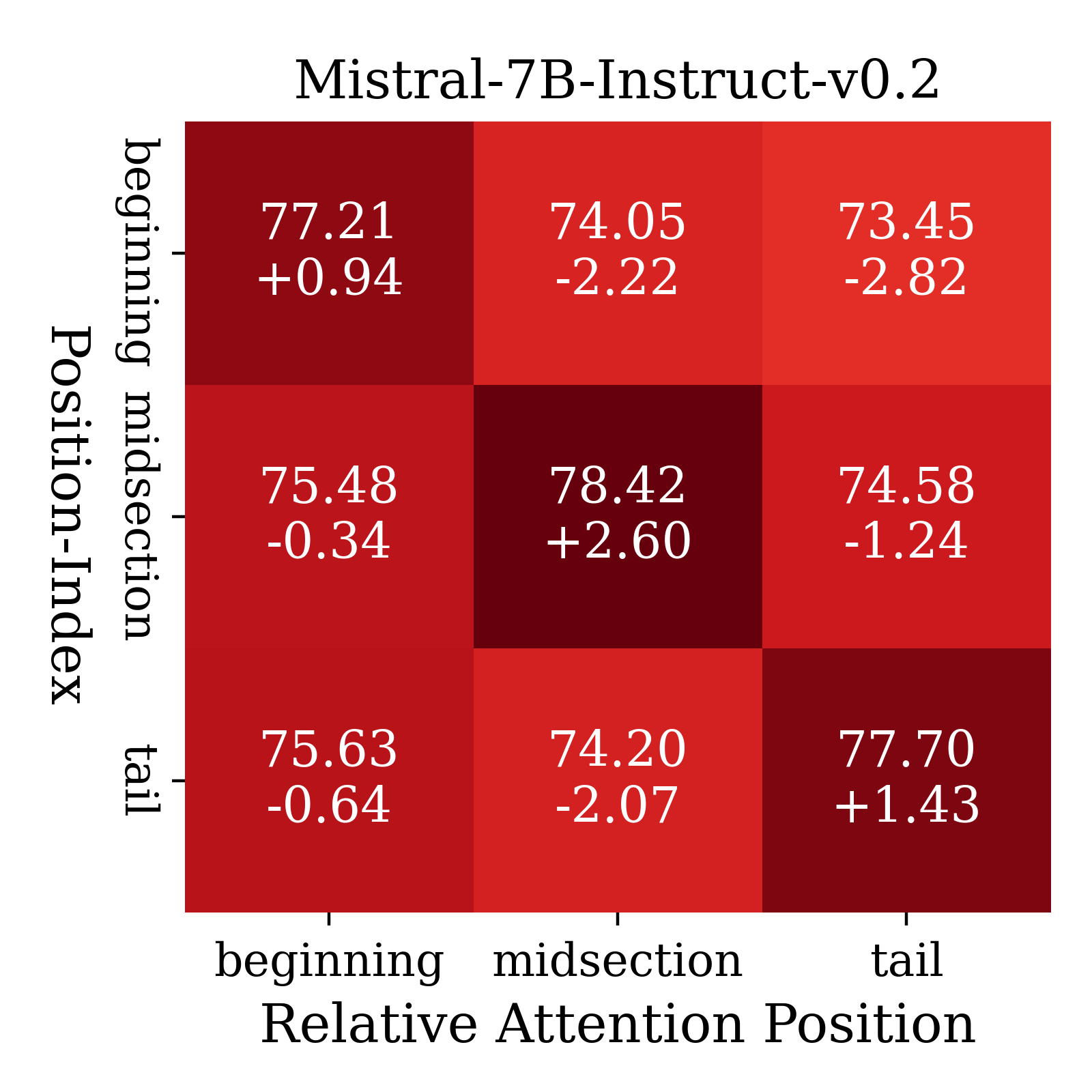}
    \end{minipage}
    \begin{minipage}[b]{0.47\columnwidth}
        \centering
        \includegraphics[width=\columnwidth]{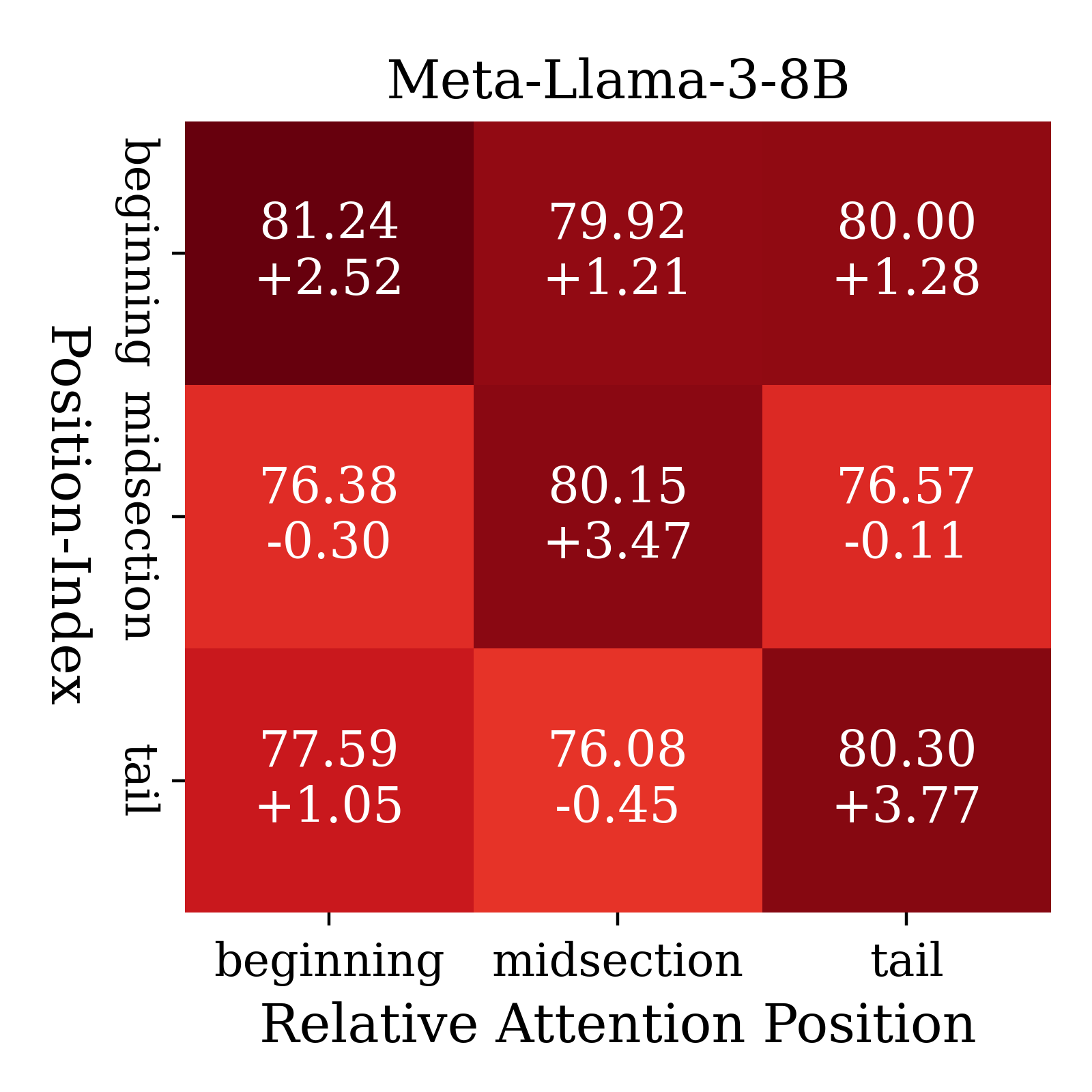}
    \end{minipage}
    \caption{Results of Mistral-Instruct-v0.2 and Llama-3 using absolute attention instruction with position-index. \textbf{Top}: 3-document, \textbf{Bottom}: 9-document}
    \label{fig:ascending_posword}
\end{figure}

\paragraph{Results}\Cref{fig:ascending_posword} presents the results for both the 3-document and 9-document settings. The absolute attention instructions successfully guide the models when referencing position-index, albeit slightly less effectively than ID-index. The clear diagonal pattern emerges, indicating improved performance when the position-index and attention segment match, and degraded performance on mismatched cases.

\paragraph{Discussion} The results demonstrate that numerical IDs indexes are not unique in their ability to guide the models' attention. Position words can serve as effective indexes for documents in each part of the search results, enabling regional control through attention instructions.

\section{Related work}
\label{sec:related}
\paragraph{Retrieval Augmented Generation} \citet{petroni2020context} were the first to apply RAG with pretrained language models on unsupervised question answering. \citet{lewis2020retrieval} originated the extractive open-domain question answering with retrieval augmentation. While the external knowledge and information provide solutions to open-domain question answering \citep{izacard2021leveraging}, LLMs still have difficulty in leveraging the retrieved passages effectively \citep{sauchuk2022role, oh2023detrimental}. Despite the conflicting misinformation and detrimental passages \cite{weller2022defending, oh2023detrimental}, disproportional attention distribution towards passages also introduces challenge \citep{akimoto2023context}. This work considers the RAG setting, assuming the search results are given.

\paragraph{Position bias in LLMs} Recent studies have demonstrated that the position of instruction \citep{liu2023instruction} and the order of answer choices \citep{zheng2023large} within the context can affect the performance and generation of LLMs. LLMs also have primary bias and recency bias in which the attention scores are biased towards initial tokens and the context in the end, regardless of their semantic relevance to the task \citep{xiao2023efficient, qin2023nlp}. \citet{liu2024lost} investigated the long-context reasoning of LLMs and noted the challenge that the information in the middle is likely to be overlooked.

\paragraph{Addressing position bias through context reordering and finetuning} Some researchers propose mitigating position bias by reordering the context based on relevance \citep{wang2024augmenting, peysakhovich2023attention, liu2024aligning}. However, these explicitly designed orders may not always work as expected \citep{liu2024lost}. Others suggest addressing position bias through continual finetuning of LLMs \citep{he2023never, an2024make, fu2024data, liu2024zero}. These methods aim to strengthen attention over all parts of the context or scale up LLMs' context window length without losing information accessing capability, but they require processing training data and additional finetuning, which can be computationally expensive.

\paragraph{Addressing position bias through position embedding modification and logits calibration} \citet{chen2023fortify, he2024position, zhang2024found} suggest that RoPE (Rotary Position Embedding) introduces long-term attention decay and propose modifying the position embeddings to address position bias. \citet{chen2023fortify} merges the attention of multiple parallel runs with different RoPE bases, while \citet{zhang2024found} re-scales the position indices to smaller values. \citet{he2024position} adjusts the attention scores by adding placeholder tokens between different segments to mitigate the effect of instruction on the adjacent document. However, these approaches either require parallel runs or hyperparameter tuning, introducing additional computational overhead. Alternatively, \citet{zhou2024batch} introduced Batch Calibration (BC), a zero-shot and inference-only calibration method that estimates the contextual bias by marginalizing the LLM scores in the batched input, addressing biases in LLMs without modifying position embeddings. In contrast to these approaches, we focus on leveraging the instruction-following capability of LLMs to achieve fine-grained usage of different documents and investigate the implicit correlation between semantic attention and the attention scores of LLMs.

\section{Conclusion and Future Work}
\label{sec:conclusion}
We empirically study how sensitive language models are to attention instructions via a series of controlled experiments. We find that language models can be guided to pay more attention to a document or region through direct indexing. However, models do not have positional awareness of where each document is relative to the context. We compare five open-sourced LLMs and find that Llama-3 has strong instruction following capability and better long-context accessing and reasoning performance. Our results and analysis provide new insights into solving the position bias. While in this study, we focus on semantically controlling the distribution of attention to different positions, this can be extended to more industrial directions, such as distributing attention based on relevance scores or source information confidence to achieve more effective RAG. 


\clearpage
\section{Limitations}
Our study has several limitations that should be acknowledged. First, we limited the search results to include only one document containing the gold answer, while real-world scenarios may involve multiple documents with correct or partially correct answers and conflicting information. Moreover, the gold document position is unknown in real-world scenarios, requiring a pre-identification of the attention position when implementing attention instructions in RAG applications. Future research could explore the effectiveness of attention instructions in these more complex settings.
Second, due to computational resource limitations, we experimented with a maximum of 9 documents and tested models with sizes ranging from 7B to 8B, leaving the exploration of larger contexts and models for future work. Finally, we focused on the correlation between semantic prompts and attention values, and did not investigate closed-source language models. Future research could expand the scope by examining the attention instruction following capabilities of these models.
Addressing these limitations and exploring attention instructions in more diverse settings will further enhance our understanding of their potential and guide the development of more effective RAG models.

\section{Ethics Statement}
In preparing and submitting this research paper, we affirm that our work adheres to the highest ethical standards and is devoid of any ethical issues. The study did not involve any human subjects or sensitive data, and all models and datasets used are publicly available. We acknowledge the potential risks associated with large language models and have focused our research on understanding their attention mechanisms to contribute to the development of more transparent and controllable models. The code and data used in this study have been made publicly available to ensure reproducibility and promote further research in this area.


\bibliography{custom}

\clearpage
\appendix


\section{Appendix}
\label{sec:appendix}
\subsection{Prompt template for 9-document}

\begin{figure}[ht]
    \vspace{-0.2em}
    \centering
    \includegraphics[width=\columnwidth]{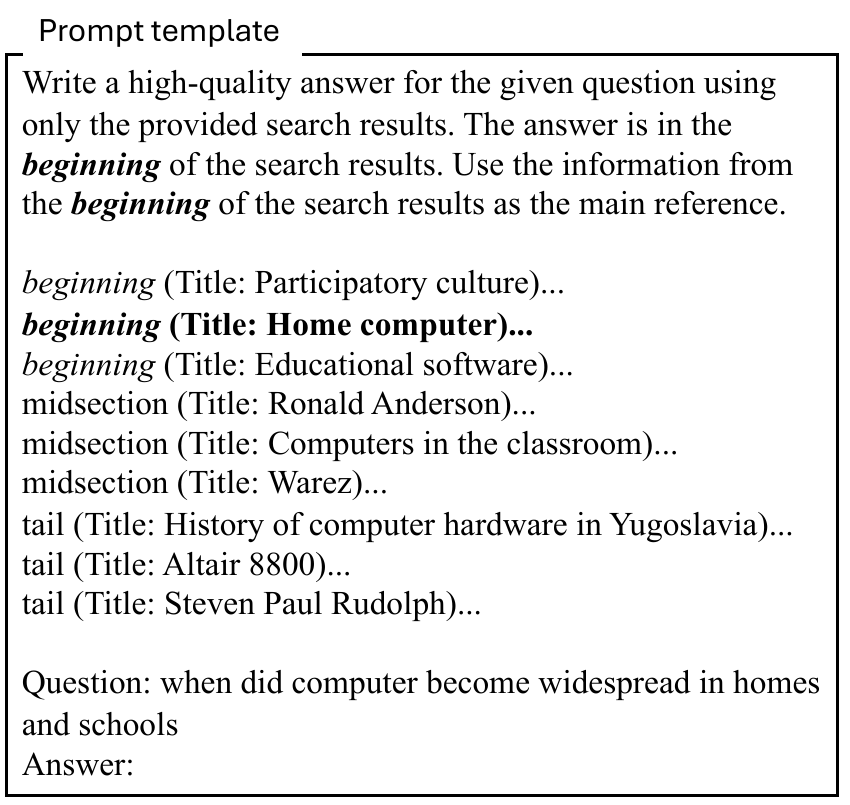}
    \caption{Prompt template for combining absolute attention instruction with position indexes.}
    \label{fig:pos_token}
    \vspace{-0.2em}
\end{figure}

\subsection{Closed-book and Oracle Performance}
In closed-book settings, only the question is provided, and the models generate answers based on their parametric memory. In the oracle setting, the gold document containing the answer is given, and the models are instructed to use that document.
\label{appx:closed-book}


\begin{table}[h]
    \centering
    \resizebox{\linewidth}{!}{
    \begin{tabular}{lcc}
    \toprule
    Model & Closed-Book(\%) & Oracle(\%) \\
    \midrule
    Llama 2 7B Chat & 34.84 & 83.95 \\
    Llama 3 8B Instruct & 39.92 & 91.71 \\
    Tulu 2 7B & 34.12 & 85.05  \\
    Mistral 7B Instruct v0.1 & 25.95 & 87.27 \\
    Mistral 7B Instruct v0.2 & 42.79 & 91.53 \\
    \bottomrule
    \end{tabular}}
    \caption{The models we tested and their performance at multi-document question answering in closed-book and oracle setting}
    \label{tab:model_closedbook}
\end{table}

\subsection{Algorithm to flatten the attention}
\label{sec:flatten_weight}

The algorithm for flattening the attention scores from 4 dimension to 3 dimension. 

\begin{algorithm}
\caption{Flatten 4D Attention Tensor to 2D}
\begin{algorithmic}[1]
\Require $attention\_ori$ \Comment{4D attention tensor (num\_layers, num\_heads, seq\_length, seq\_length)}
\Require $split\_ranges$ \Comment{Ranges for splitting sequences}
\Require $selected\_parts$ \Comment{Indices of selected parts}
\Ensure $attention$ \Comment{2D attention matrix (seq\_length, num\_layers)}
\Ensure $split\_atten$ \Comment{Attention scores for selected parts (num\_parts, num\_layers)}

\State $attention\_ori \gets$ Convert $attention\_ori$ to numpy array
\State $attention\_ori \gets$ Squeeze $attention\_ori$ \Comment{Remove singleton dimensions}
\State $attention \gets \text{np.mean}(attention\_ori, \text{axis=0})$ \Comment{Average over heads}
\State $attention \gets attention[:, -1, :]$ \Comment{Select the last token}
\State $attention \gets \text{np.swapaxes}(attention, 0, 1)$ \Comment{Swap dimensions}

\State $split\_atten \gets \{\}$ \Comment{Initialize empty list}
\For{$i$ in $selected\_parts$}
    \State $sub \gets attention[split\_ranges[i][0]:split\_ranges[i][1]]$
    \State $score\_total \gets \text{np.mean}(sub, \text{axis=0})$
    \State $split\_atten.\text{append}(score\_total)$
\EndFor

\State \Return $attention, split\_atten$
\end{algorithmic}
\end{algorithm}


\clearpage

\subsection{Relative Attention Instruction under No-Index Setting}
\label{appx:relative_no_index}
\begin{figure}[h]
    \centering
    \begin{minipage}[b]{0.47\columnwidth}
        \centering
        \includegraphics[width=\columnwidth]{images/Llama-2-7b-chat-hf_notchat/3_documents_position_level_no_docid.png}
    \end{minipage}
    \begin{minipage}[b]{0.47\columnwidth}
        \centering
        \includegraphics[width=\columnwidth]{images/Meta-Llama-3-8B-Instruct/3_documents_position_level_no_docid.png}
    \end{minipage}
    \begin{minipage}[b]{0.47\columnwidth}
    \centering
    \includegraphics[width=\columnwidth]{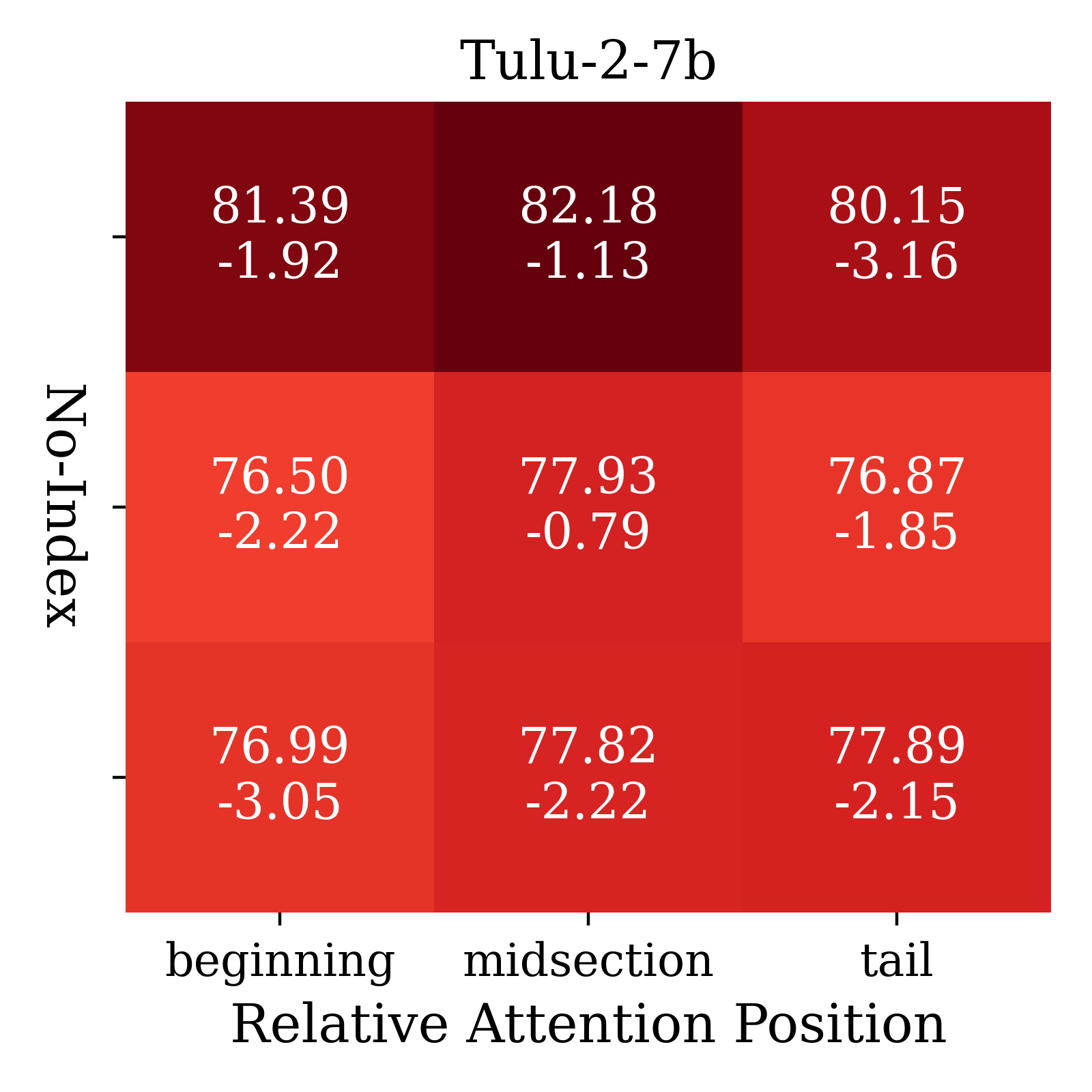}
    \end{minipage}
    \begin{minipage}[b]{0.47\columnwidth}
        \centering
        \includegraphics[width=\columnwidth]{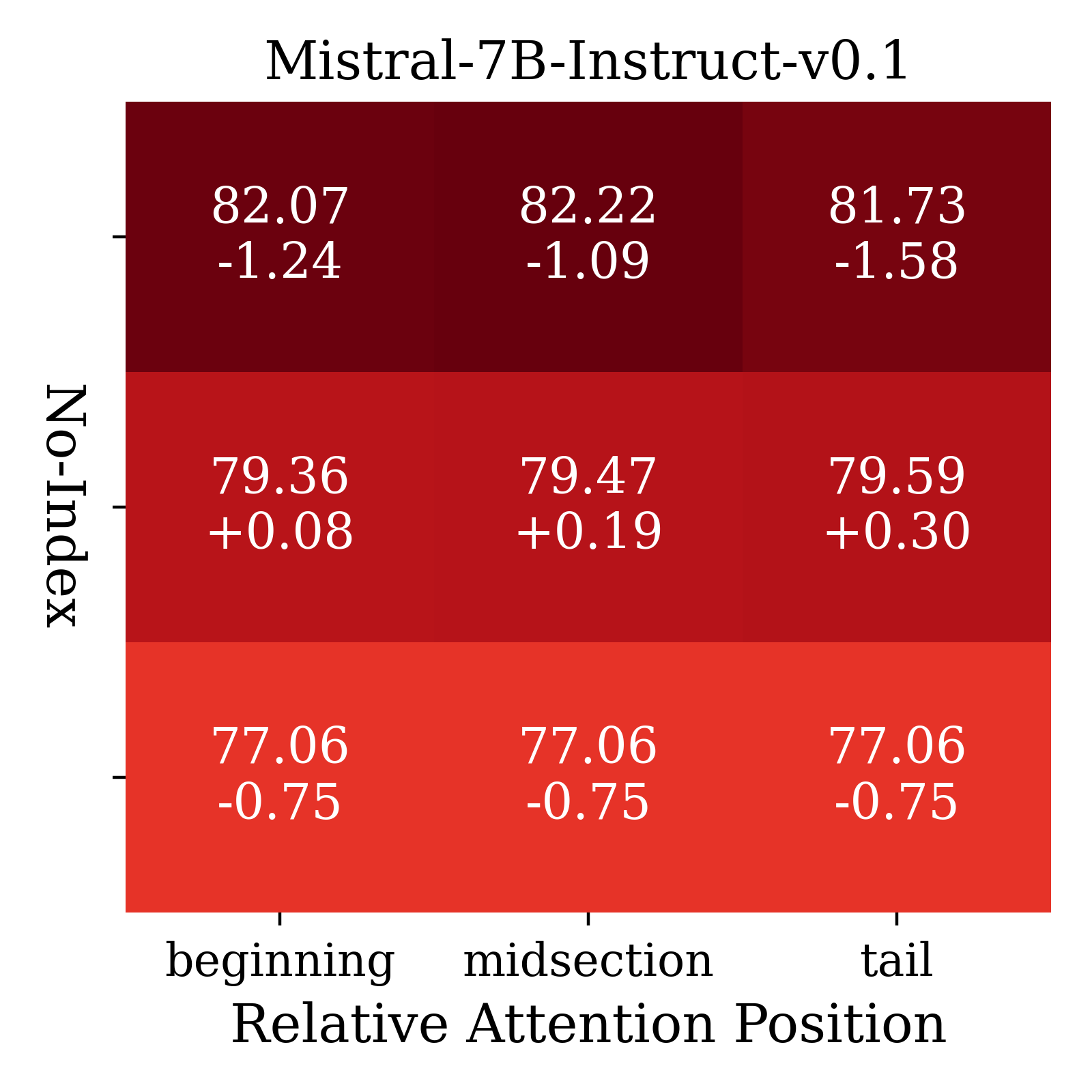}
    \end{minipage}
    \begin{minipage}[b]{0.47\columnwidth}
        \centering
        \includegraphics[width=\columnwidth]{images/Mistral-7B-Instruct-v0.2/3_documents_position_level_no_docid.png}
    \end{minipage}
    \caption{3-document: relative attention instruction under no-index setting}
    \label{fig:all_position_no_docid_3docs}
\end{figure}

\begin{figure}[h]
    \centering
    \begin{minipage}[b]{0.47\columnwidth}
        \centering
        \includegraphics[width=\columnwidth]{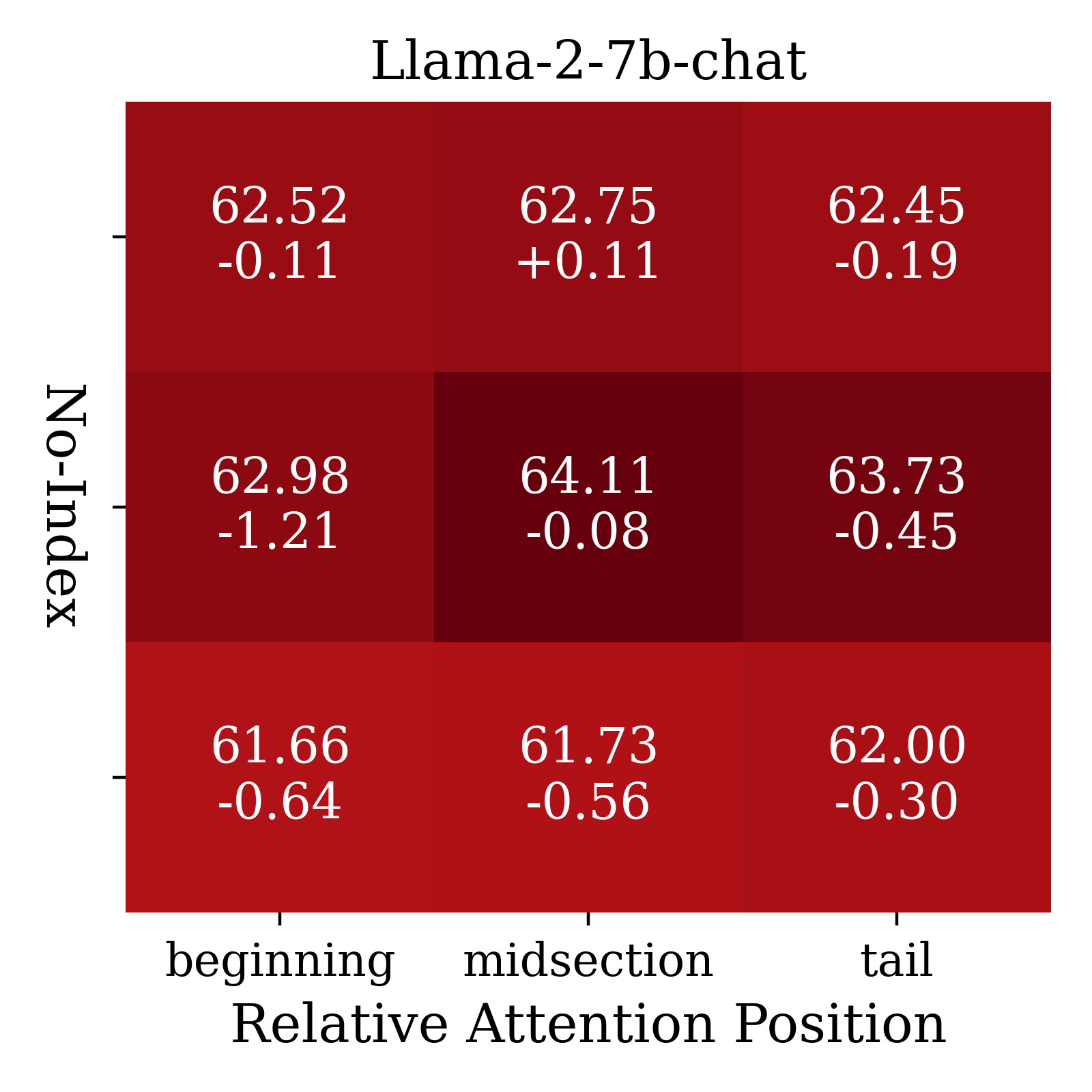}
    \end{minipage}
    \begin{minipage}[b]{0.47\columnwidth}
        \centering
        \includegraphics[width=\columnwidth]{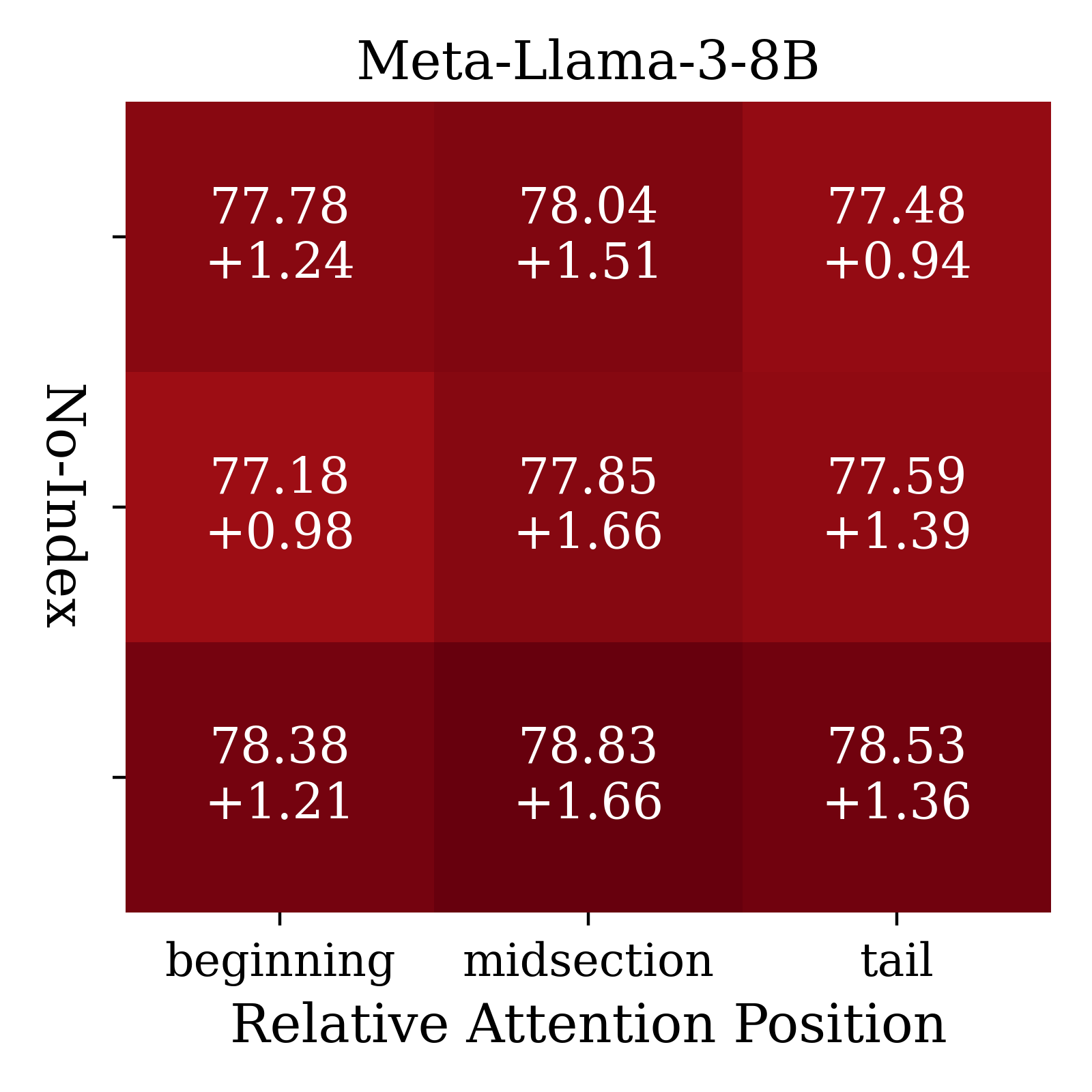}
    \end{minipage}
    \begin{minipage}[b]{0.47\columnwidth}
    \centering
    \includegraphics[width=\columnwidth]{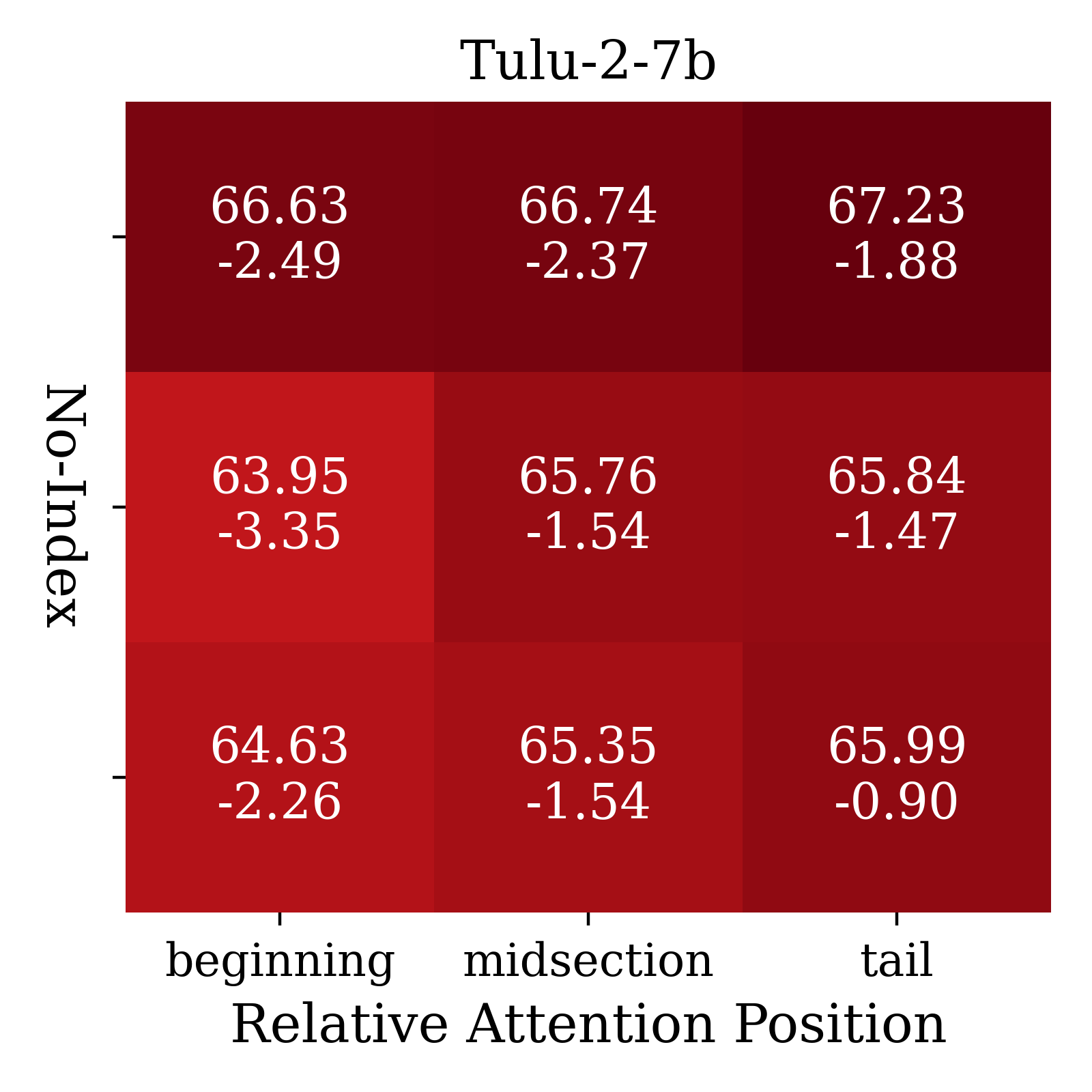}
    \end{minipage}
    \begin{minipage}[b]{0.47\columnwidth}
        \centering
        \includegraphics[width=\columnwidth]{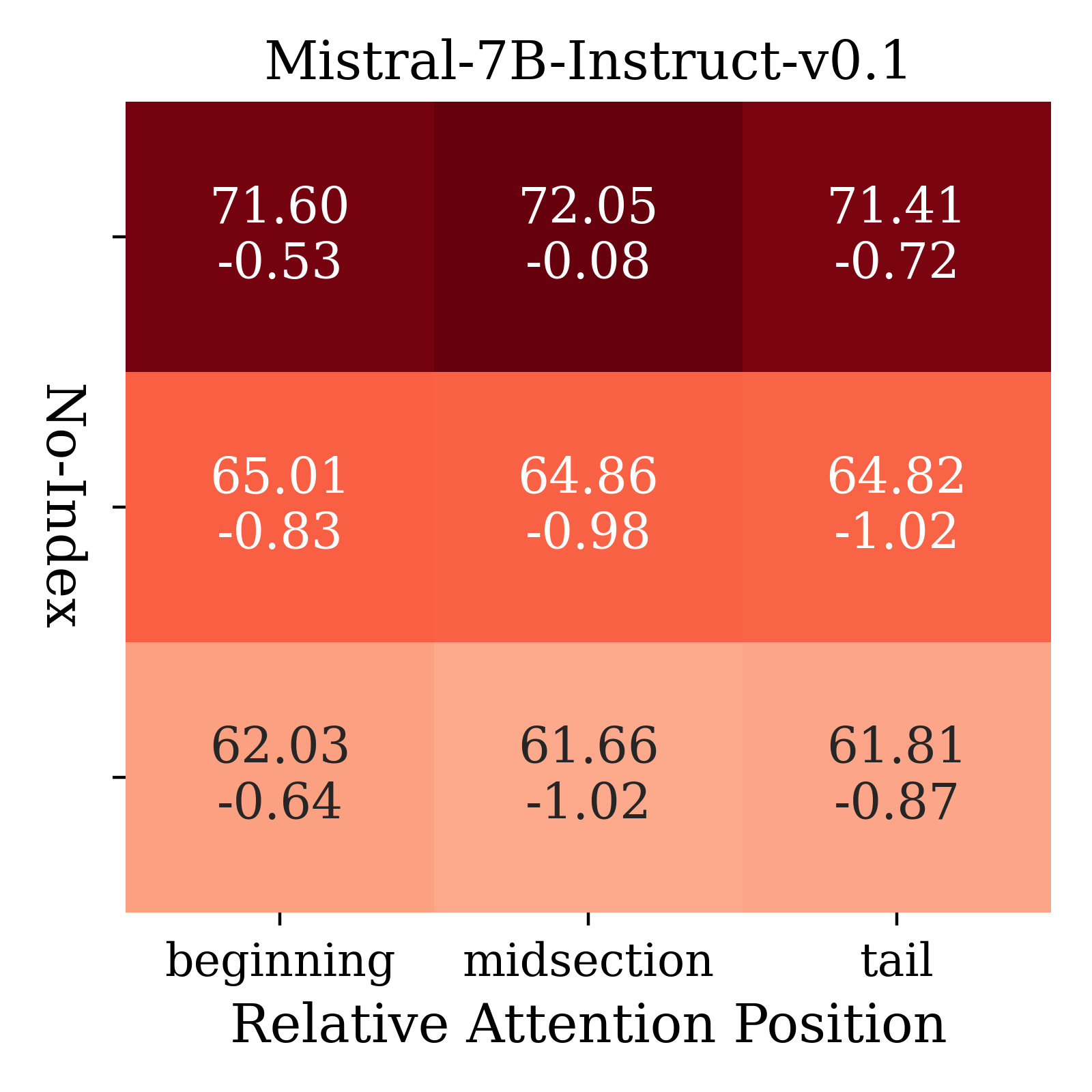}
    \end{minipage}
    \begin{minipage}[b]{0.47\columnwidth}
        \centering
        \includegraphics[width=\columnwidth]{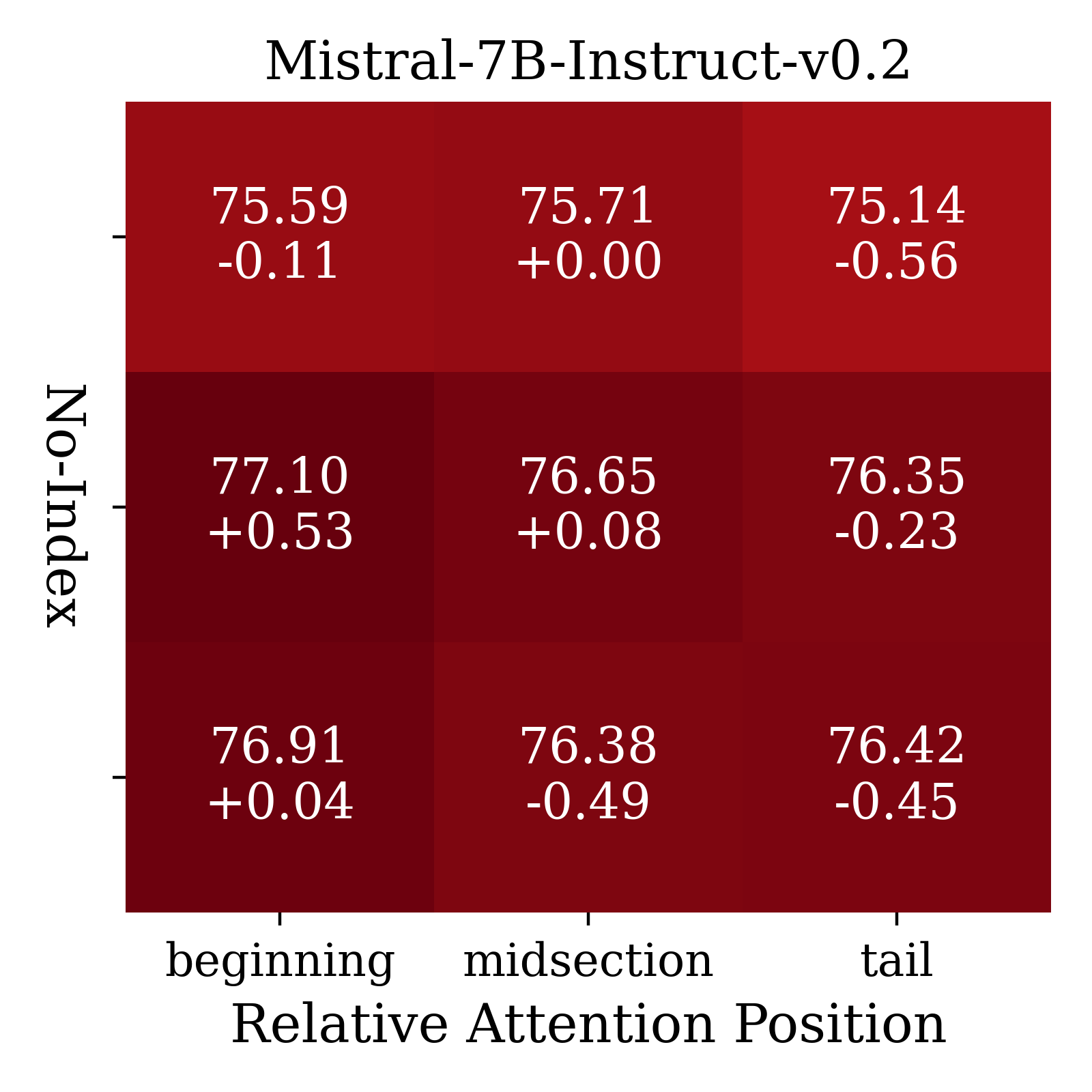}
    \end{minipage}
    \caption{9-document: relative attention instruction under no-index setting}
    \label{fig:all_position_no_docid_9docs}
\end{figure}

3-document results in \Cref{fig:all_position_no_docid_3docs}, 9-document results in \Cref{fig:all_position_no_docid_9docs}

\clearpage

\subsection{Relative Attention Instruction with Ascending Document ID Index}
\label{appx:relative_docid}
\begin{figure}[h]
    \centering
    \begin{minipage}[b]{0.47\columnwidth}
        \centering
        \includegraphics[width=\columnwidth]{images/Llama-2-7b-chat-hf_notchat/3_documents_position_level_have_docid.png}
    \end{minipage}
    \begin{minipage}[b]{0.47\columnwidth}
        \centering
        \includegraphics[width=\columnwidth]{images/Meta-Llama-3-8B-Instruct/3_documents_position_level_have_docid.png}
    \end{minipage}
    \begin{minipage}[b]{0.47\columnwidth}
    \centering
    \includegraphics[width=\columnwidth]{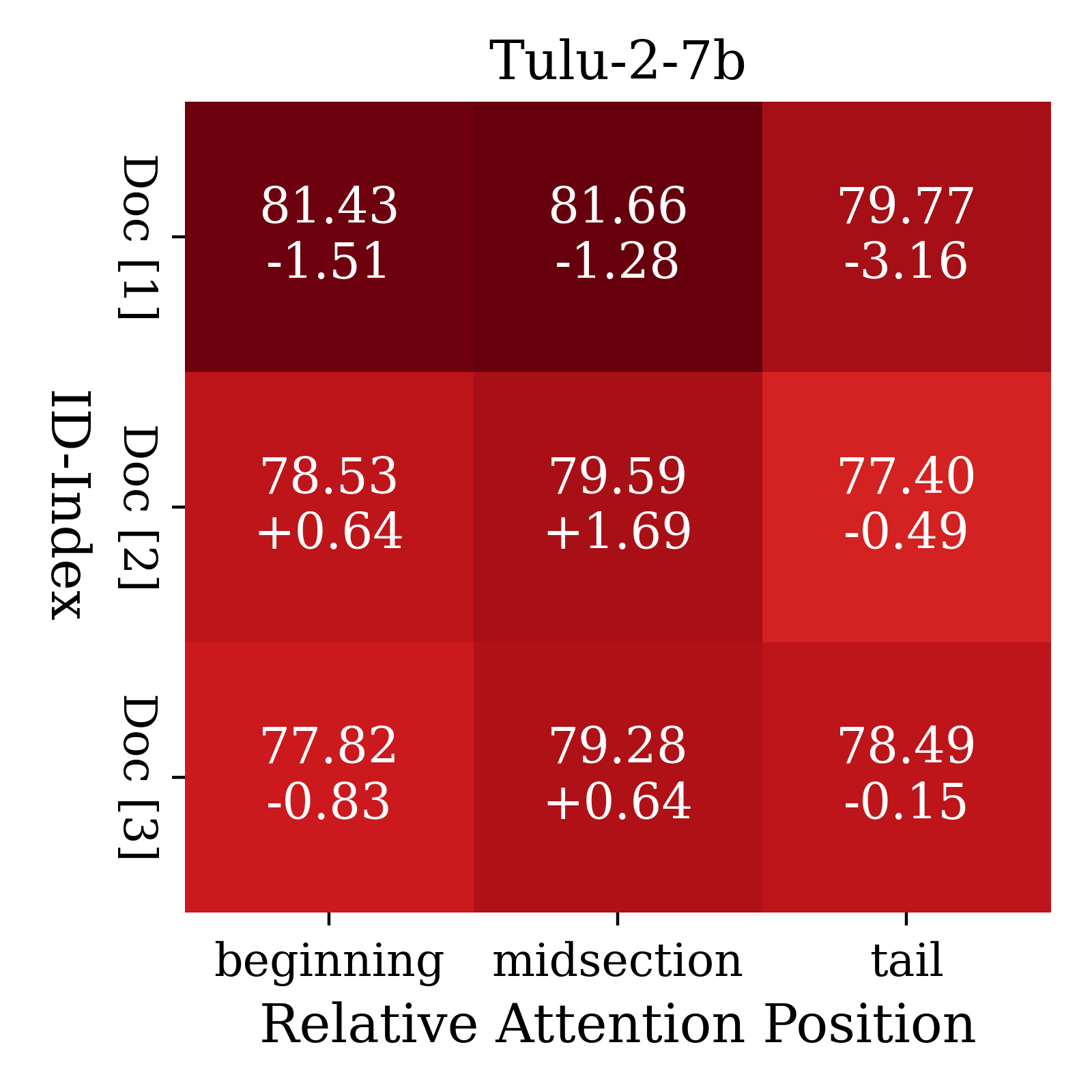}
    \end{minipage}
    \begin{minipage}[b]{0.47\columnwidth}
        \centering
        \includegraphics[width=\columnwidth]{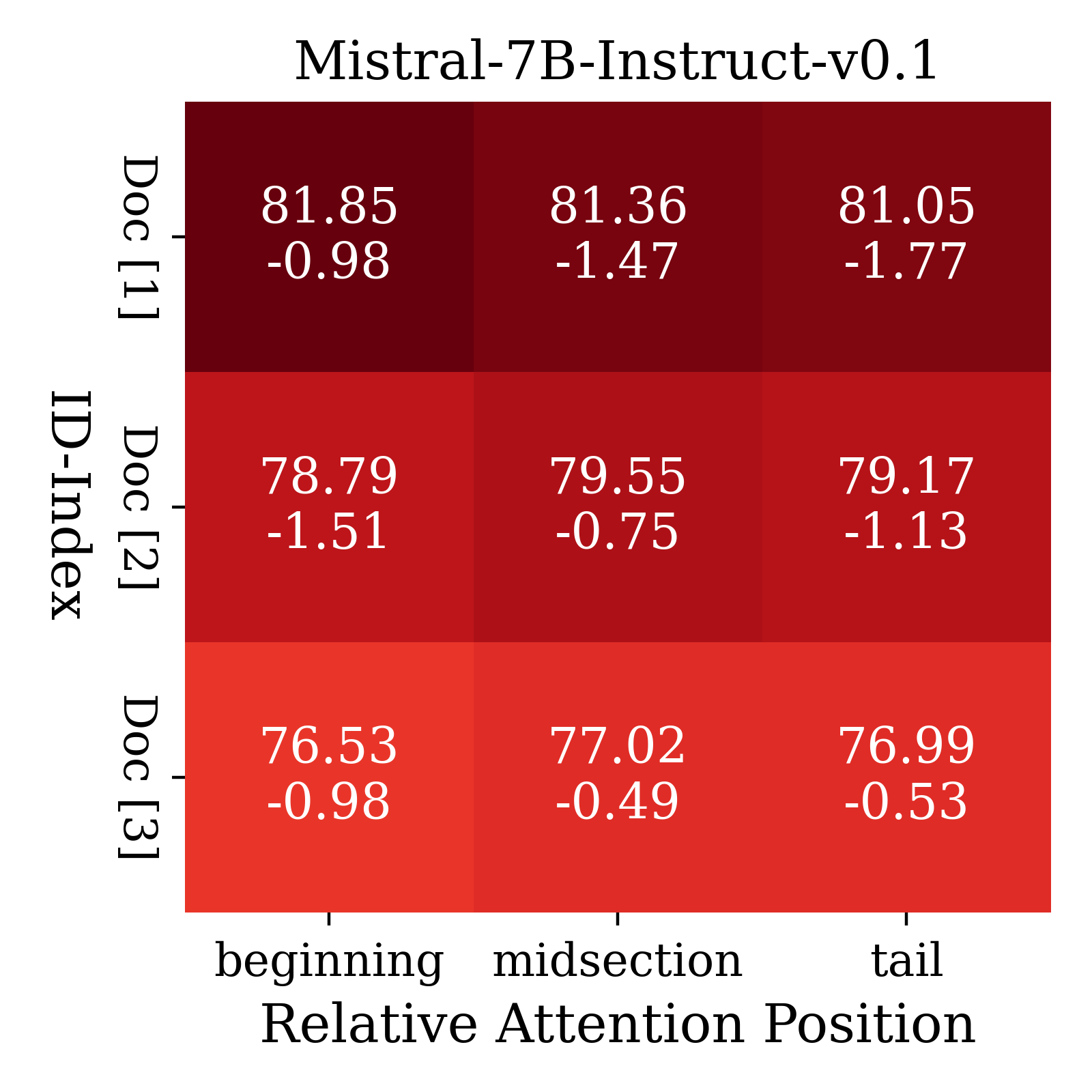}
    \end{minipage}
    \begin{minipage}[b]{0.47\columnwidth}
        \centering
        \includegraphics[width=\columnwidth]{images/Mistral-7B-Instruct-v0.2/3_documents_position_level_have_docid.png}
    \end{minipage}
    \caption{3-document: relative attention instruction under ID-index setting}
    \label{fig:all_position_have_docid_3docs}
\end{figure}

\begin{figure}[h]
    \centering
    \begin{minipage}[b]{0.47\columnwidth}
        \centering
        \includegraphics[width=\columnwidth]{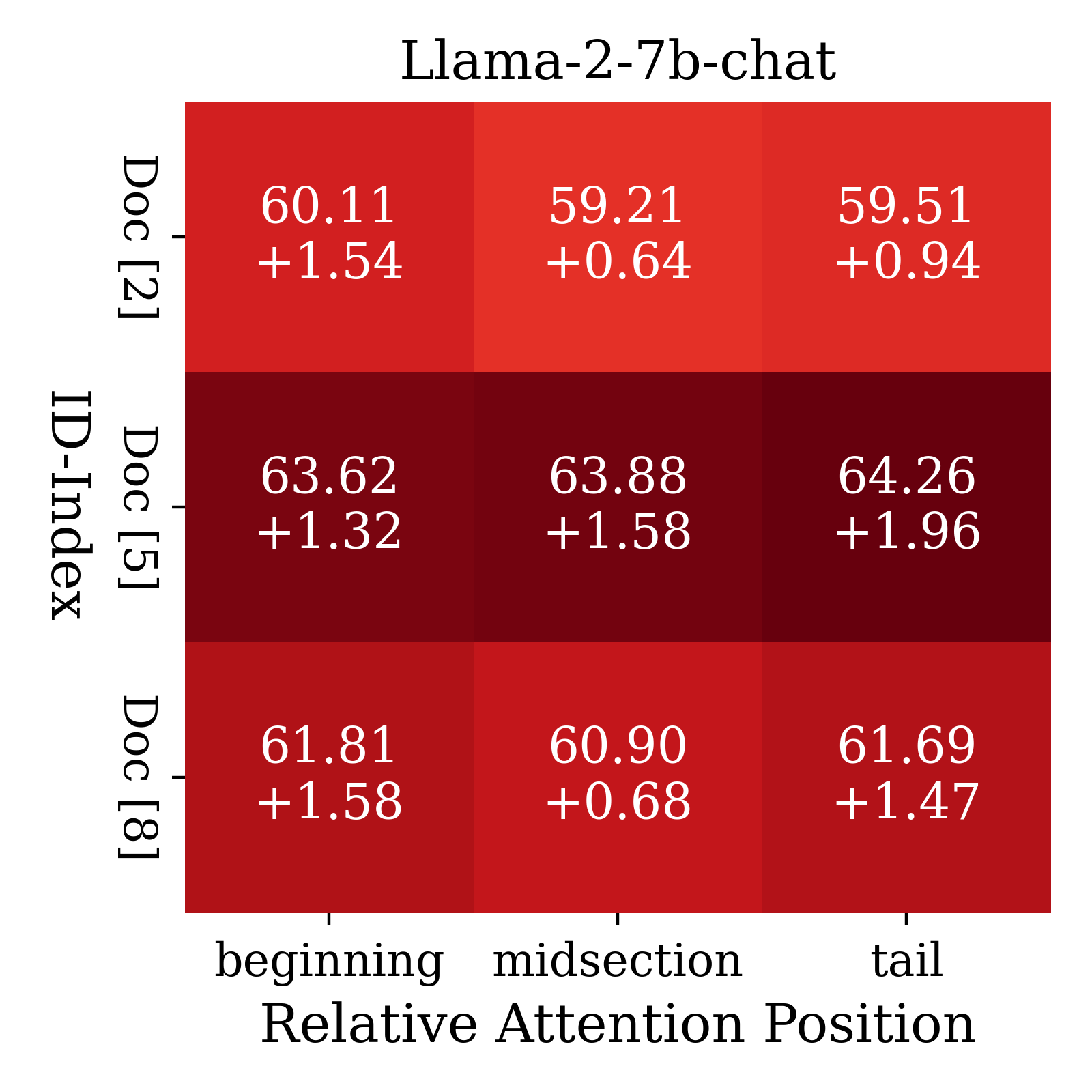}
    \end{minipage}
    \begin{minipage}[b]{0.47\columnwidth}
        \centering
        \includegraphics[width=\columnwidth]{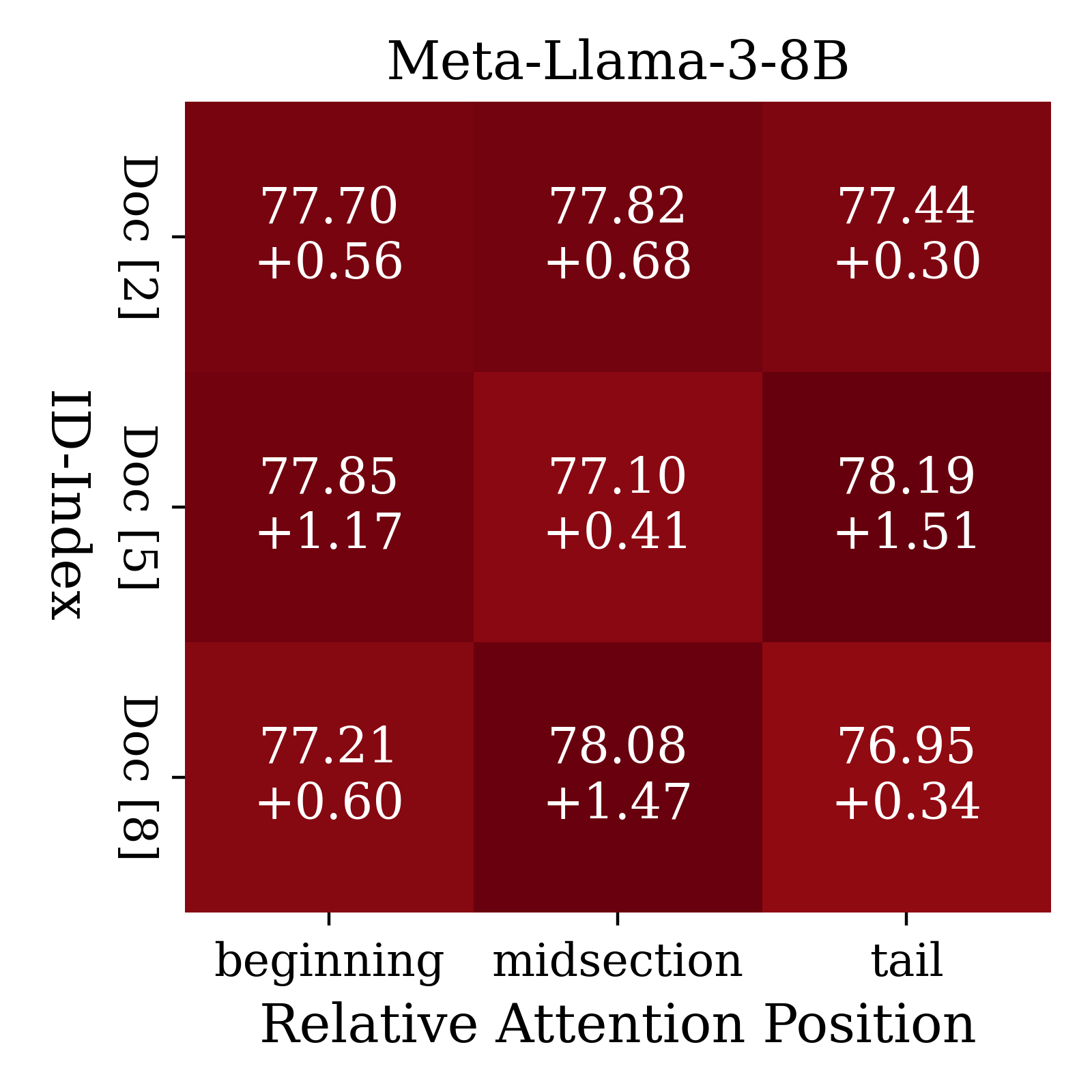}
    \end{minipage}
    \begin{minipage}[b]{0.47\columnwidth}
    \centering
    \includegraphics[width=\columnwidth]{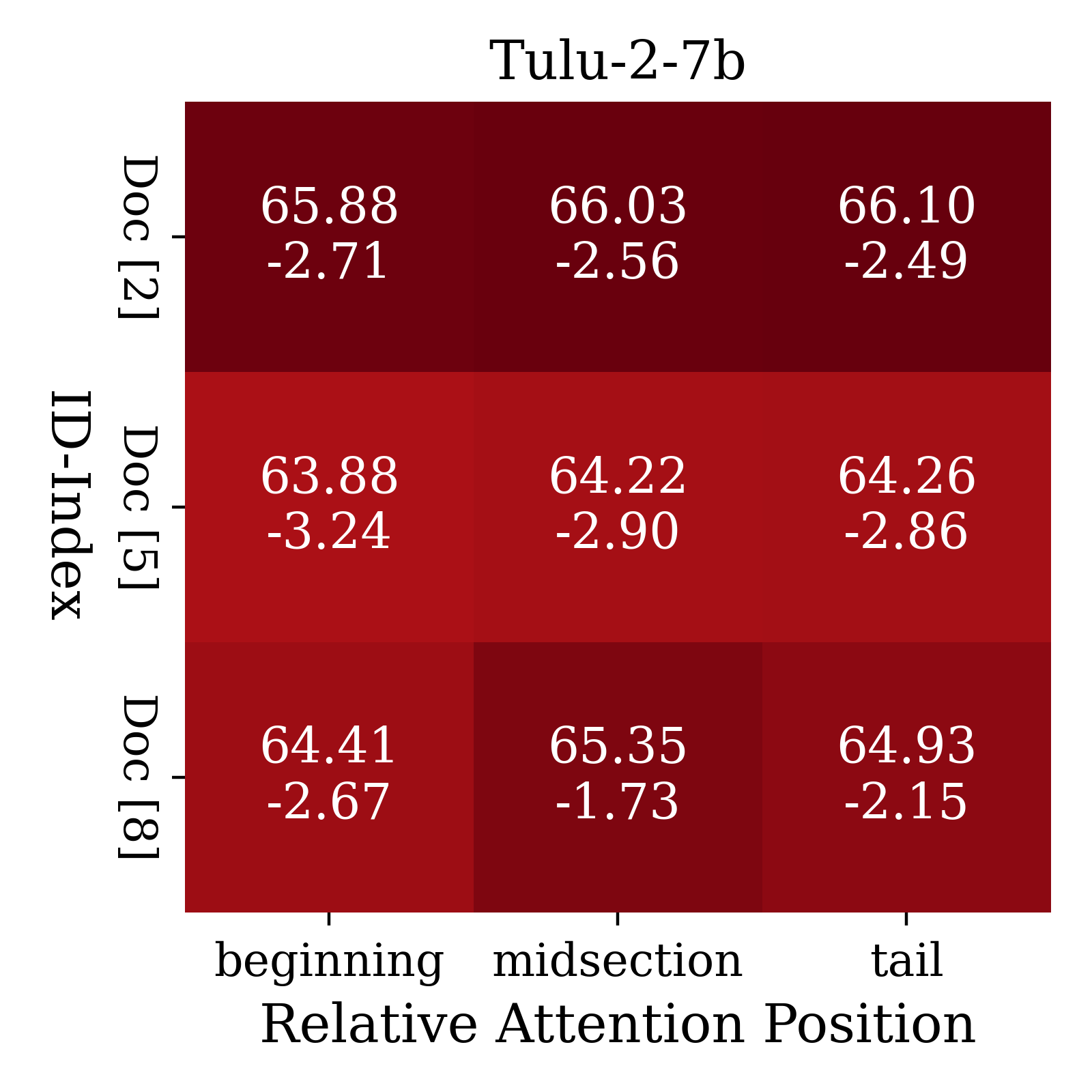}
    \end{minipage}
    \begin{minipage}[b]{0.47\columnwidth}
        \centering
        \includegraphics[width=\columnwidth]{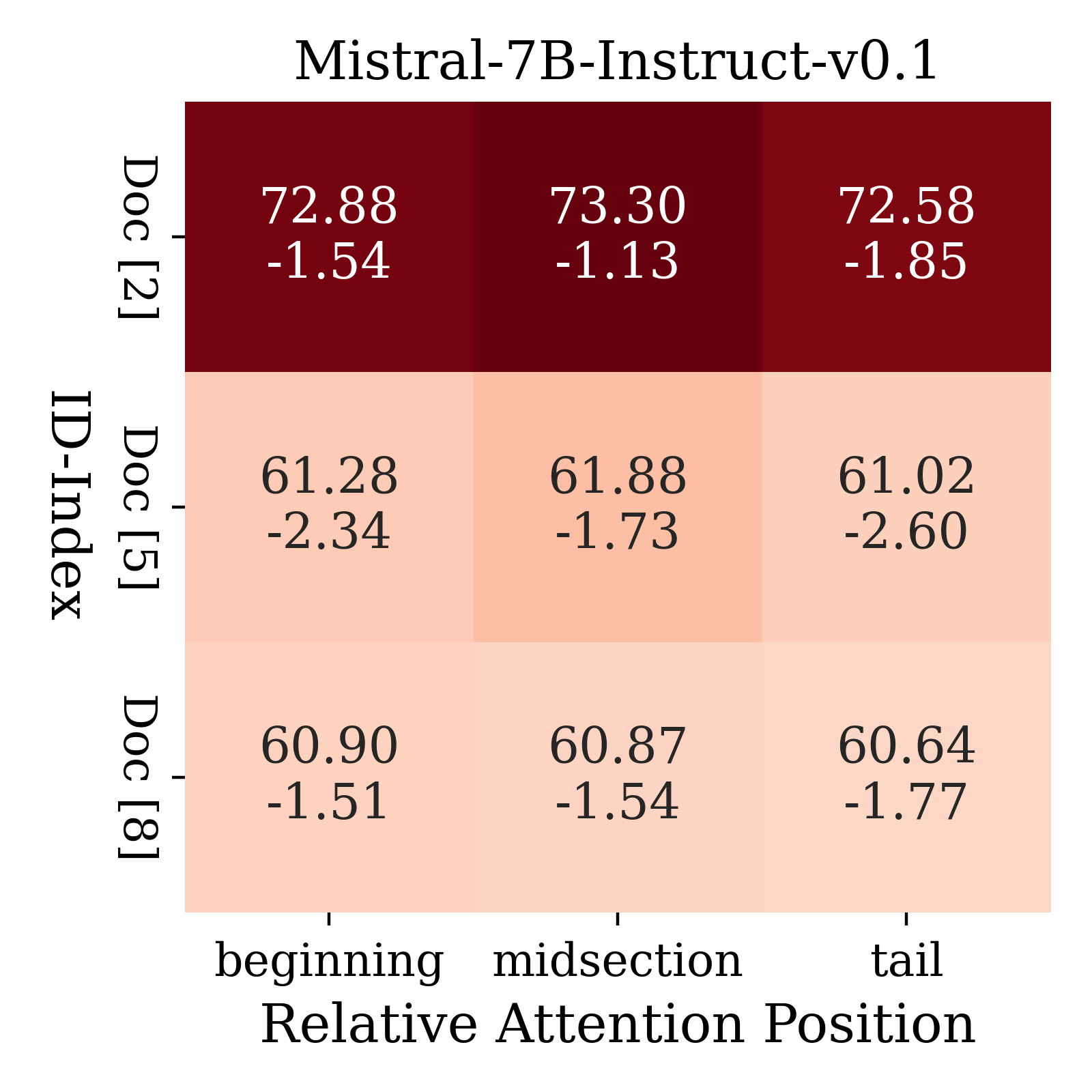}
    \end{minipage}
    \begin{minipage}[b]{0.47\columnwidth}
        \centering
        \includegraphics[width=\columnwidth]{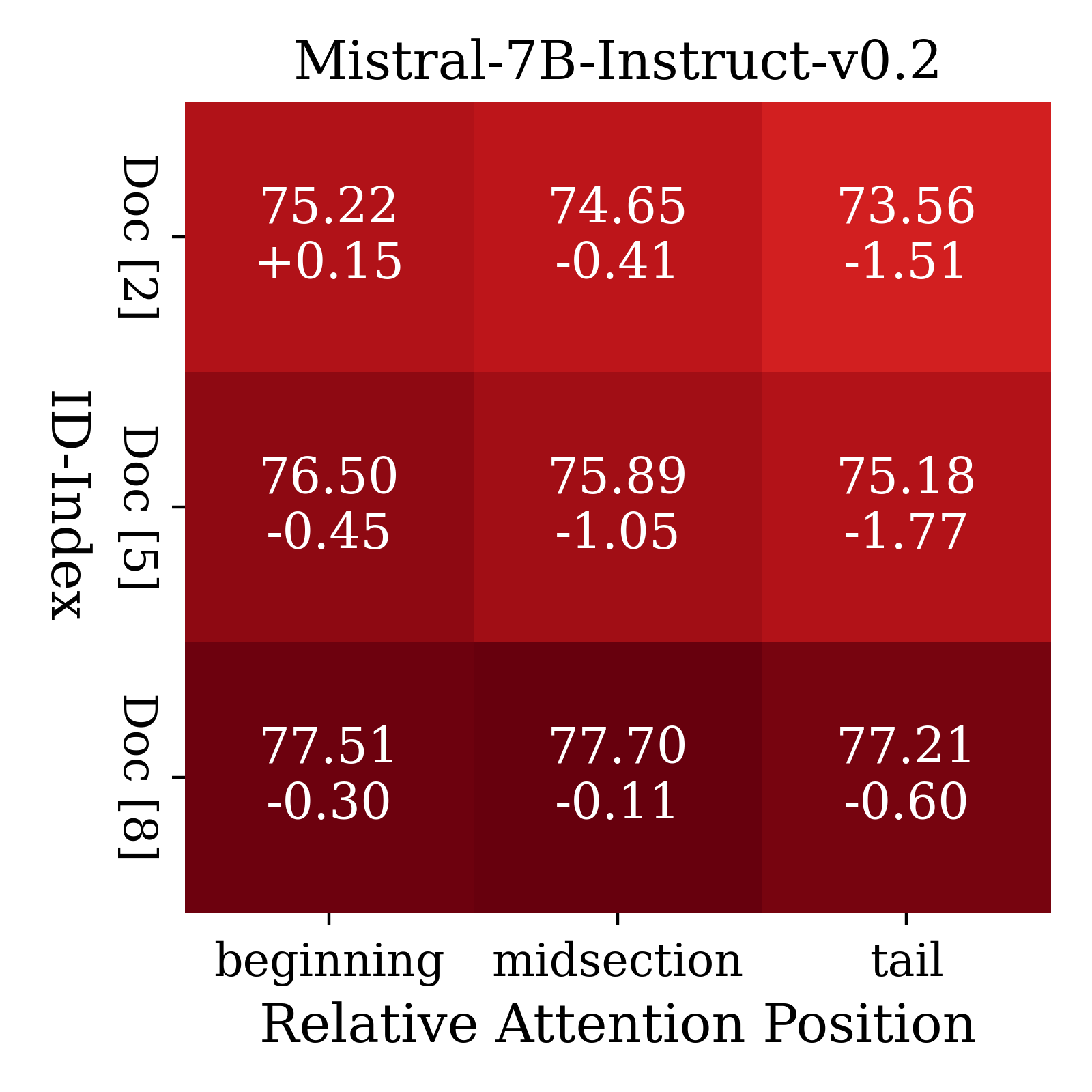}
    \end{minipage}
    \caption{9-document: relative attention instruction under ID-index setting}
    \label{fig:all_position_have_docid_9docs}
\end{figure}

3-document results in \Cref{fig:all_position_have_docid_3docs}, 9-document results in \Cref{fig:all_position_have_docid_9docs}

\clearpage

\subsection{Absolute Attention Instruction with Ascending Document ID Index}
\label{appx:absolute_docid}
\begin{figure}[h]
    \centering
    \begin{minipage}[b]{0.47\columnwidth}
        \centering
        \includegraphics[width=\columnwidth]{images/Llama-2-7b-chat-hf_notchat/3_documents_token_level_have_docid.png}
    \end{minipage}
    \begin{minipage}[b]{0.47\columnwidth}
        \centering
        \includegraphics[width=\columnwidth]{images/Meta-Llama-3-8B-Instruct/3_documents_token_level_have_docid.png}
    \end{minipage}
    \begin{minipage}[b]{0.47\columnwidth}
    \centering
    \includegraphics[width=\columnwidth]{images/tulu-2-7b_formated/3_documents_token_level_have_docid.png}
    \end{minipage}
    \begin{minipage}[b]{0.47\columnwidth}
        \centering
        \includegraphics[width=\columnwidth]{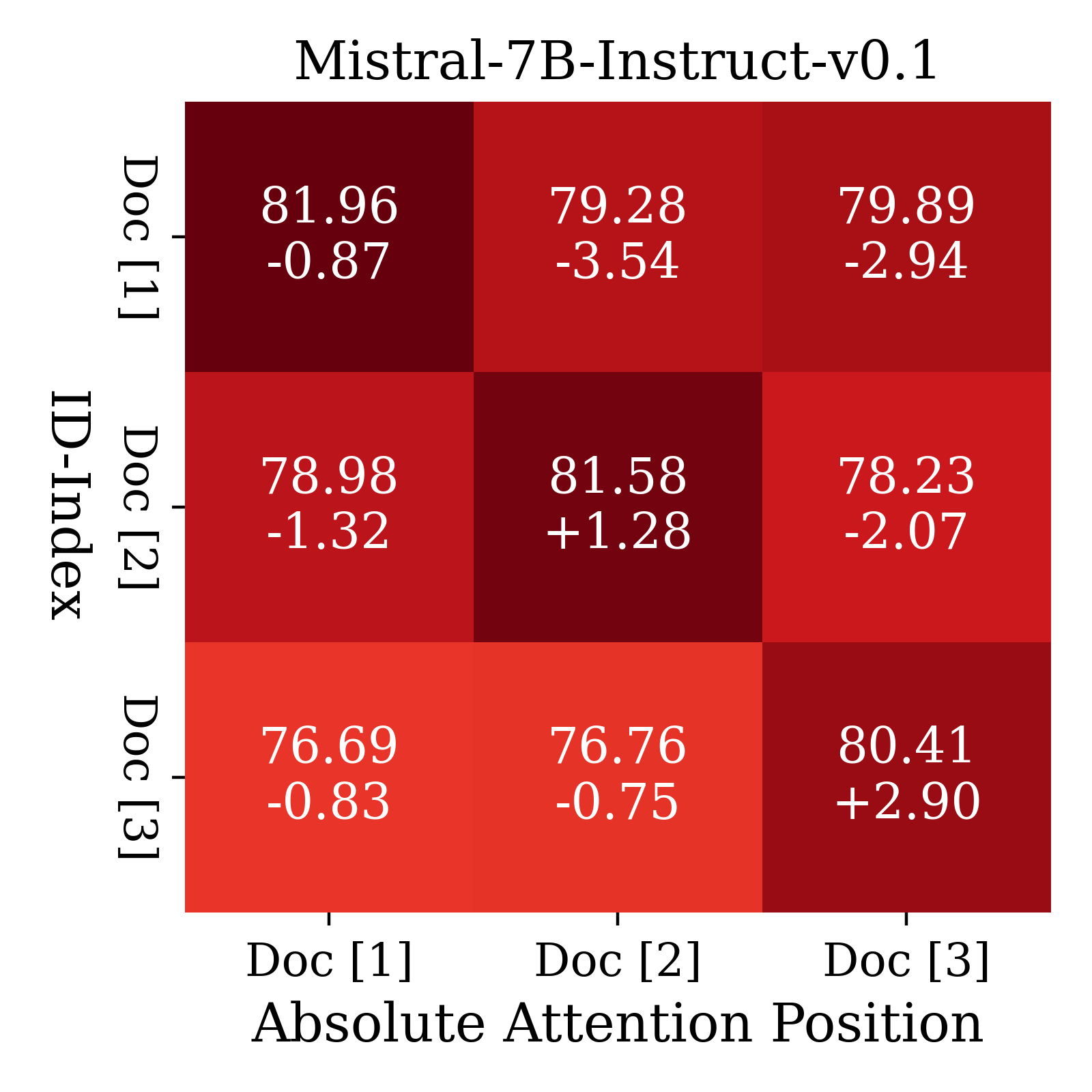}
    \end{minipage}
    \begin{minipage}[b]{0.47\columnwidth}
        \centering
        \includegraphics[width=\columnwidth]{images/Mistral-7B-Instruct-v0.2/3_documents_token_level_have_docid.png}
    \end{minipage}
    \caption{3-document: absolute attention instruction under ID-index setting}
    \label{fig:all_token_have_docid_3docs}
\end{figure}

\begin{figure}[h]
    \centering
    \begin{minipage}[b]{0.47\columnwidth}
        \centering
        \includegraphics[width=\columnwidth]{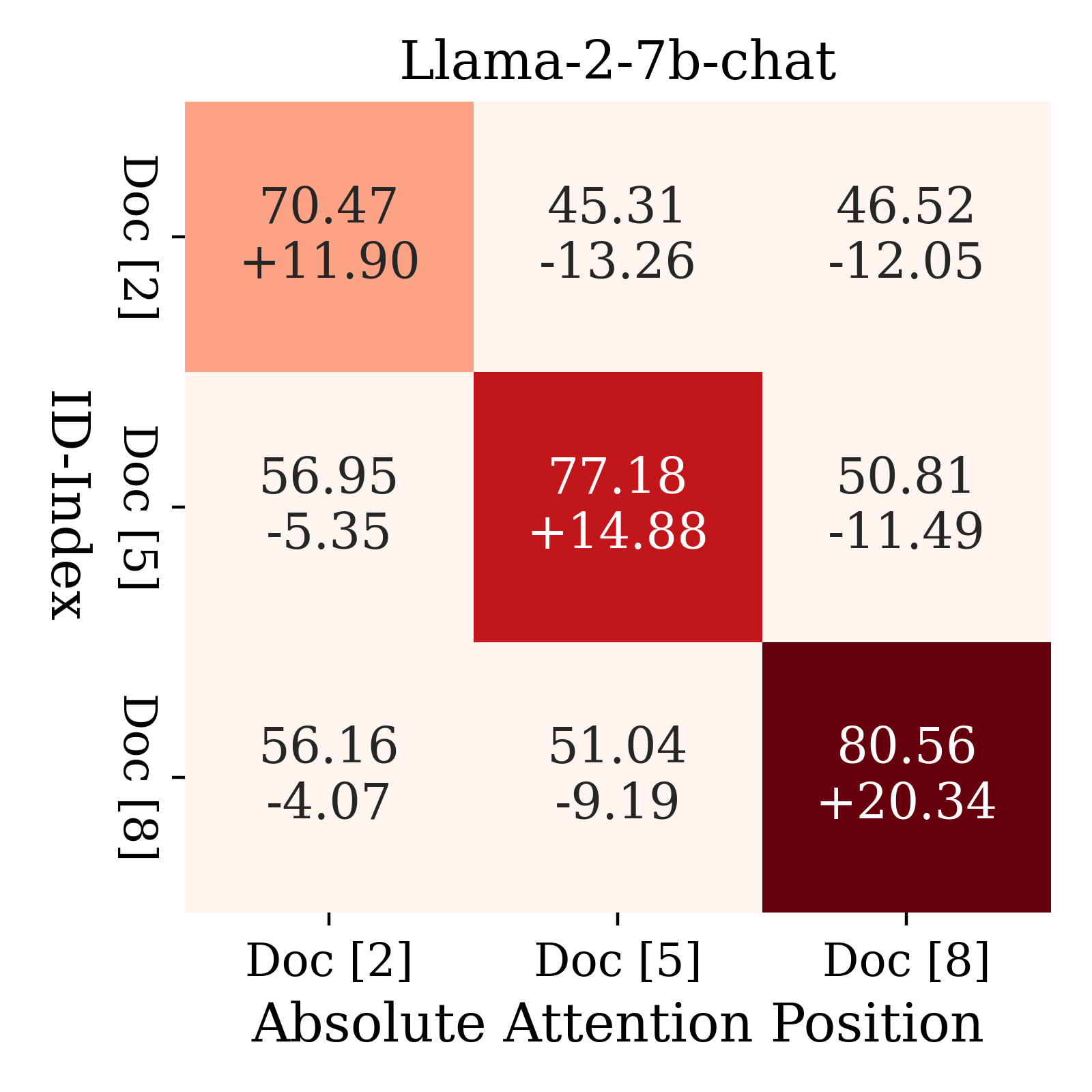}
    \end{minipage}
    \begin{minipage}[b]{0.47\columnwidth}
        \centering
        \includegraphics[width=\columnwidth]{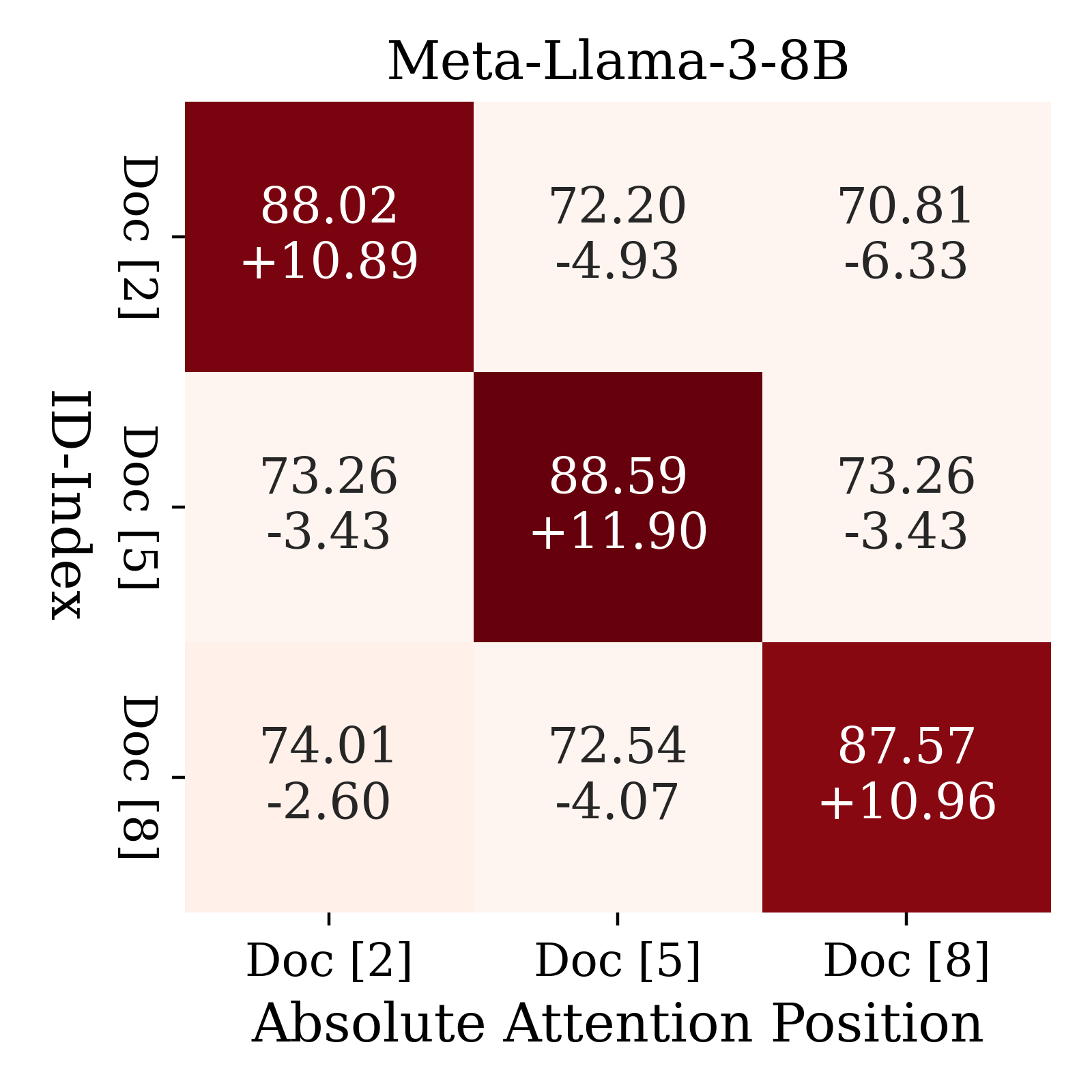}
    \end{minipage}
    \begin{minipage}[b]{0.47\columnwidth}
    \centering
    \includegraphics[width=\columnwidth]{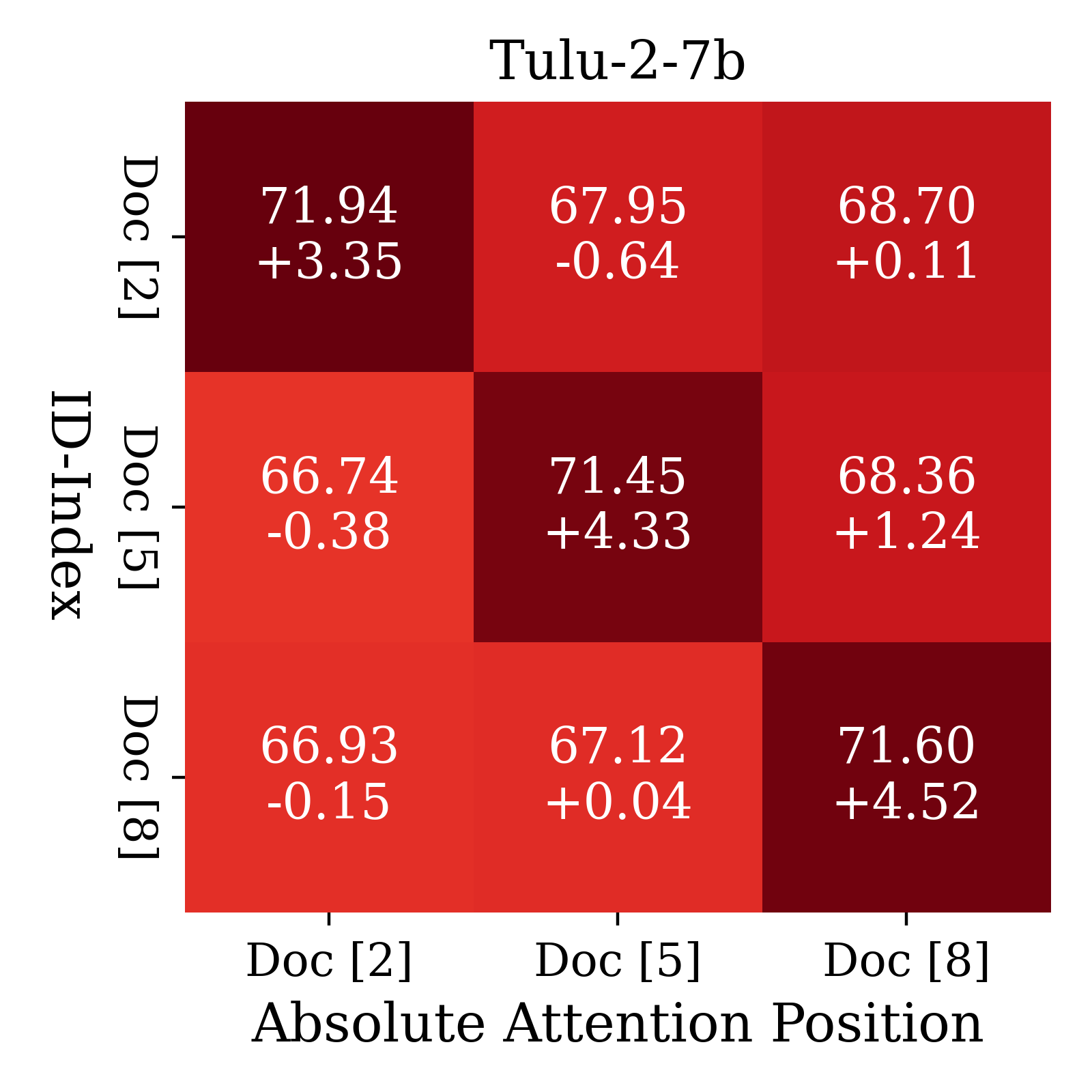}
    \end{minipage}
    \begin{minipage}[b]{0.47\columnwidth}
        \centering
        \includegraphics[width=\columnwidth]{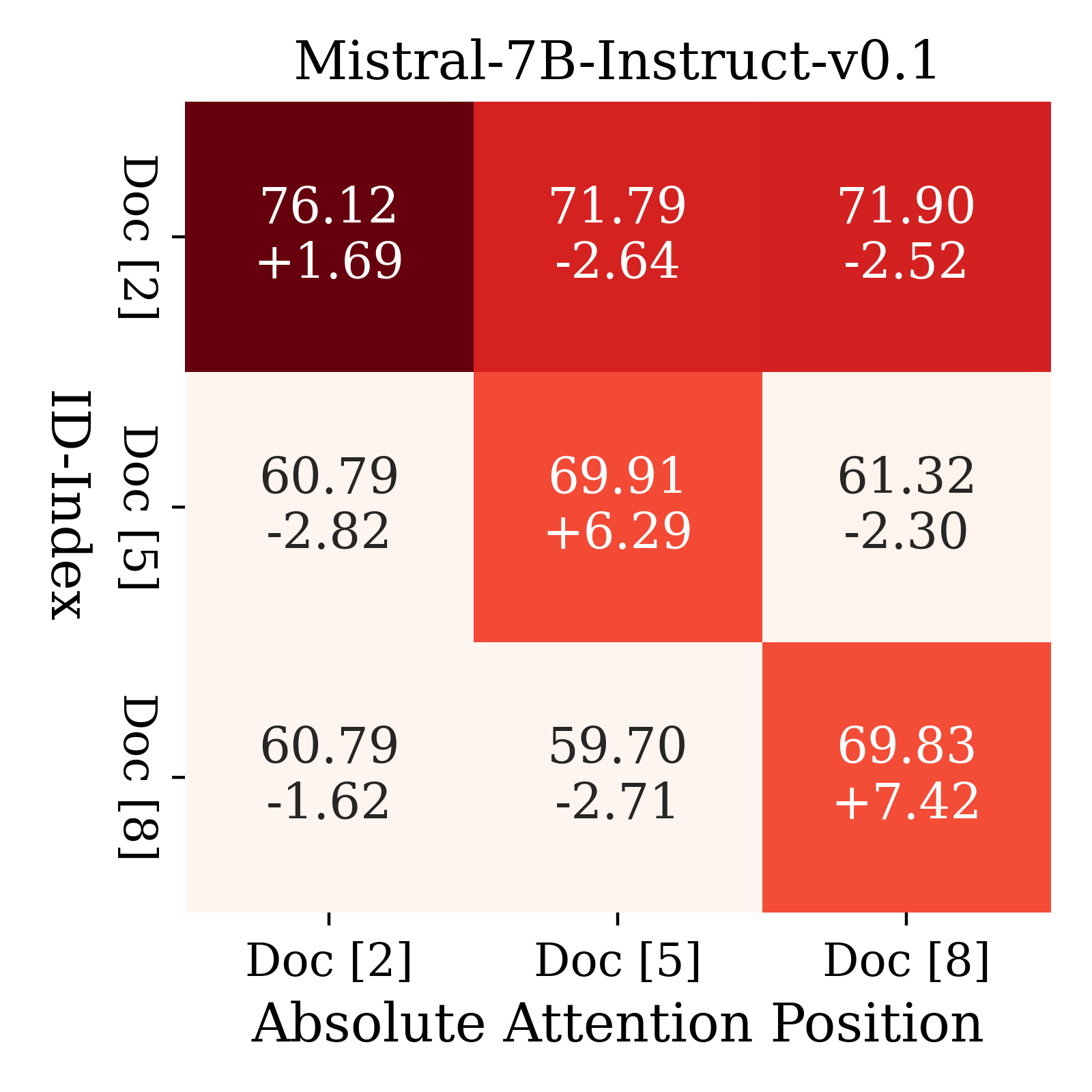}
    \end{minipage}
    \begin{minipage}[b]{0.47\columnwidth}
        \centering
        \includegraphics[width=\columnwidth]{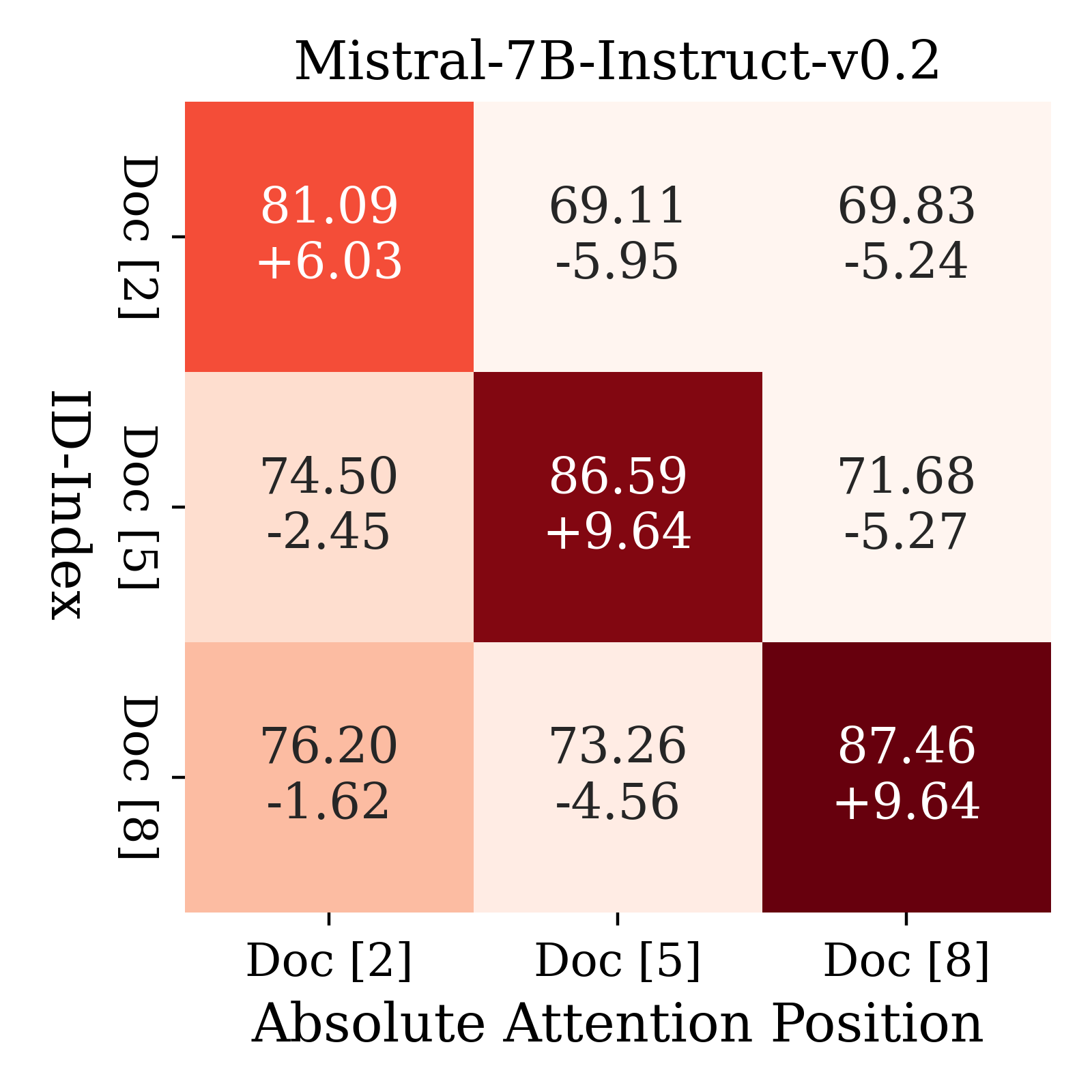}
    \end{minipage}
    \caption{9-document: absolute attention instruction under ID-index setting}
    \label{fig:all_token_have_docid_9docs}
\end{figure}

3-document results in \Cref{fig:all_token_have_docid_3docs}, 9-document results in \Cref{fig:all_token_have_docid_9docs}

\clearpage

\subsection{Absolute Attention Instruction with Reversed Document ID Index}
\label{appx:absolute_reverse_docid}
\begin{figure}[h]
    \centering
    \begin{minipage}[b]{0.47\columnwidth}
        \centering
        \includegraphics[width=\columnwidth]{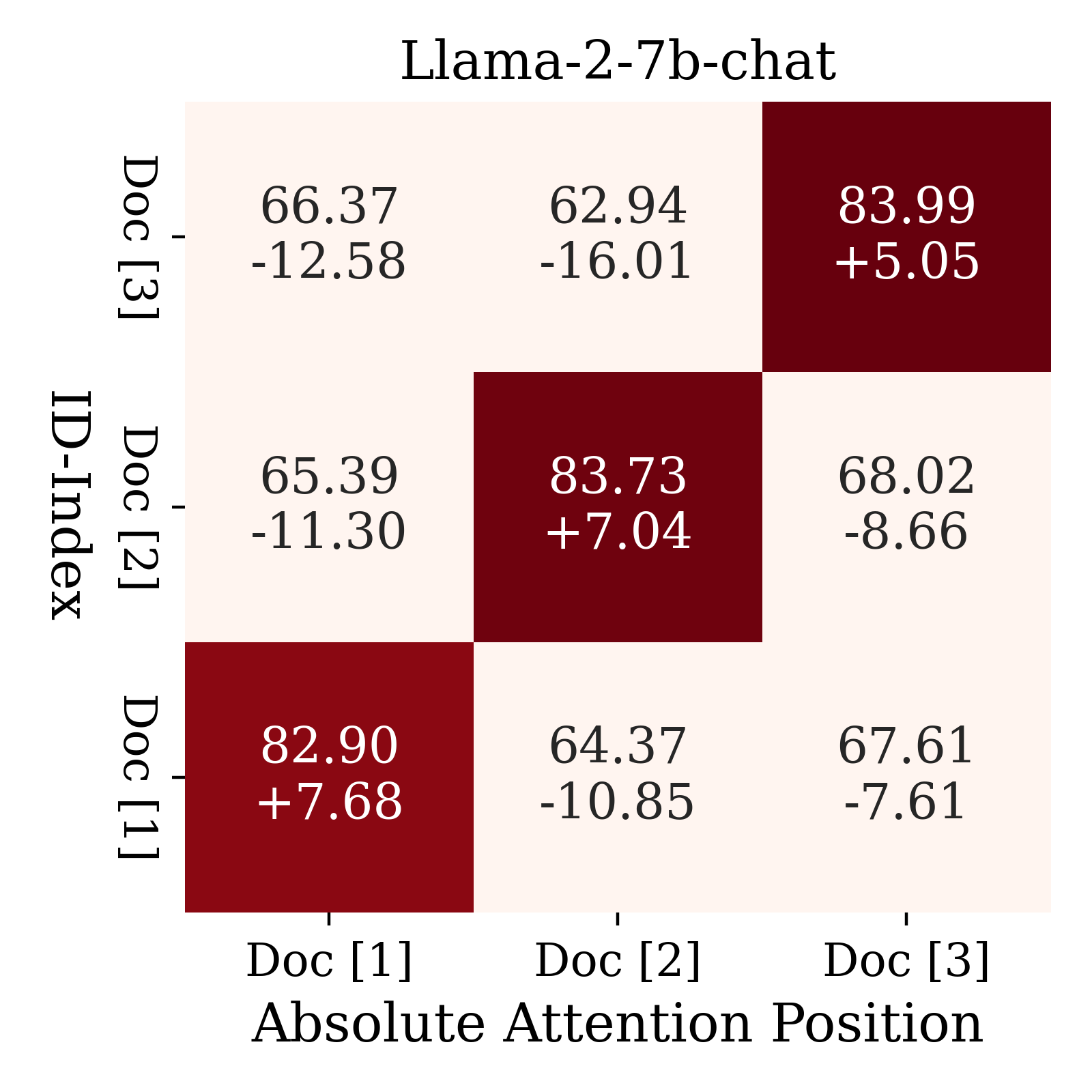}
    \end{minipage}
    \begin{minipage}[b]{0.47\columnwidth}
        \centering
        \includegraphics[width=\columnwidth]{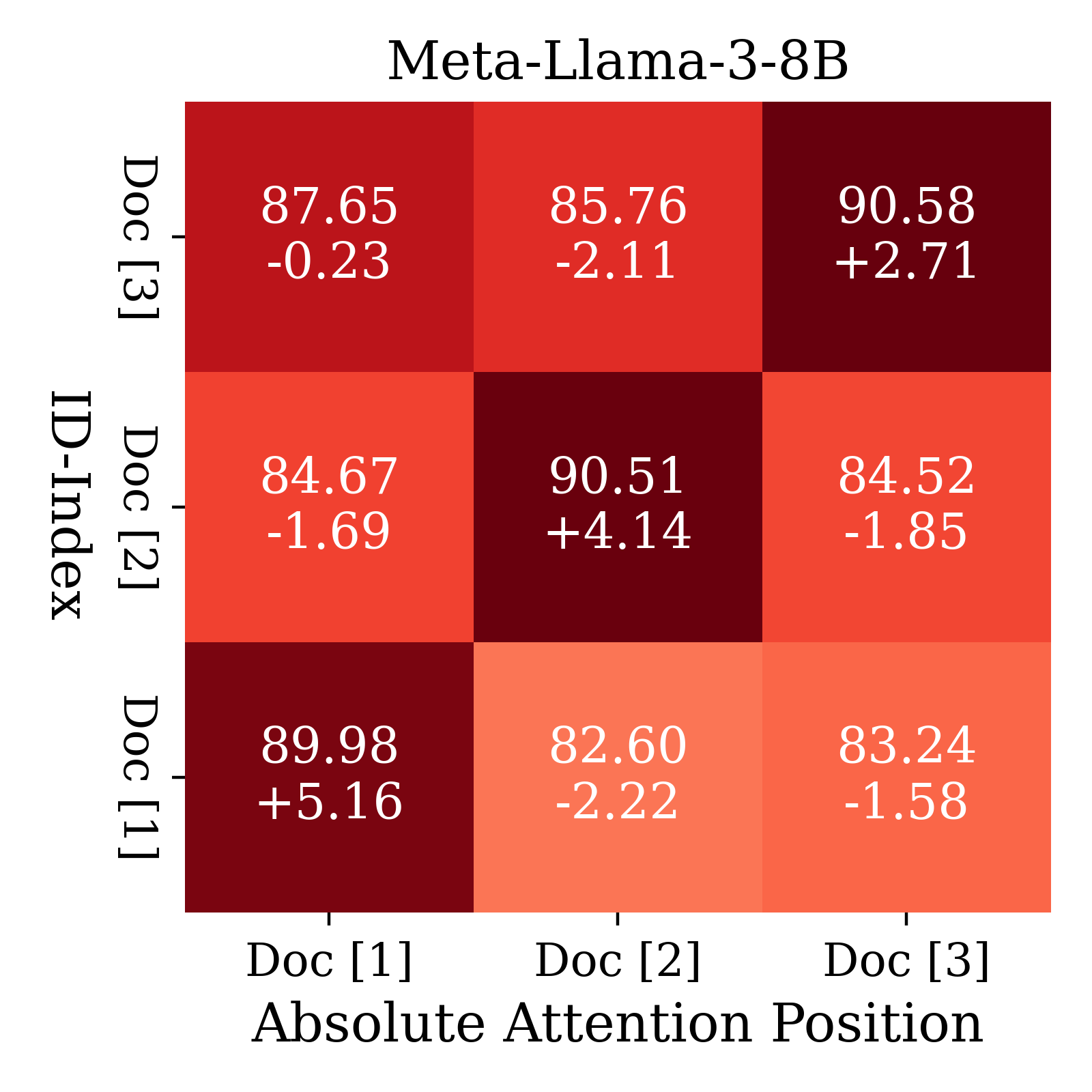}
    \end{minipage}
    \begin{minipage}[b]{0.47\columnwidth}
    \centering
    \includegraphics[width=\columnwidth]{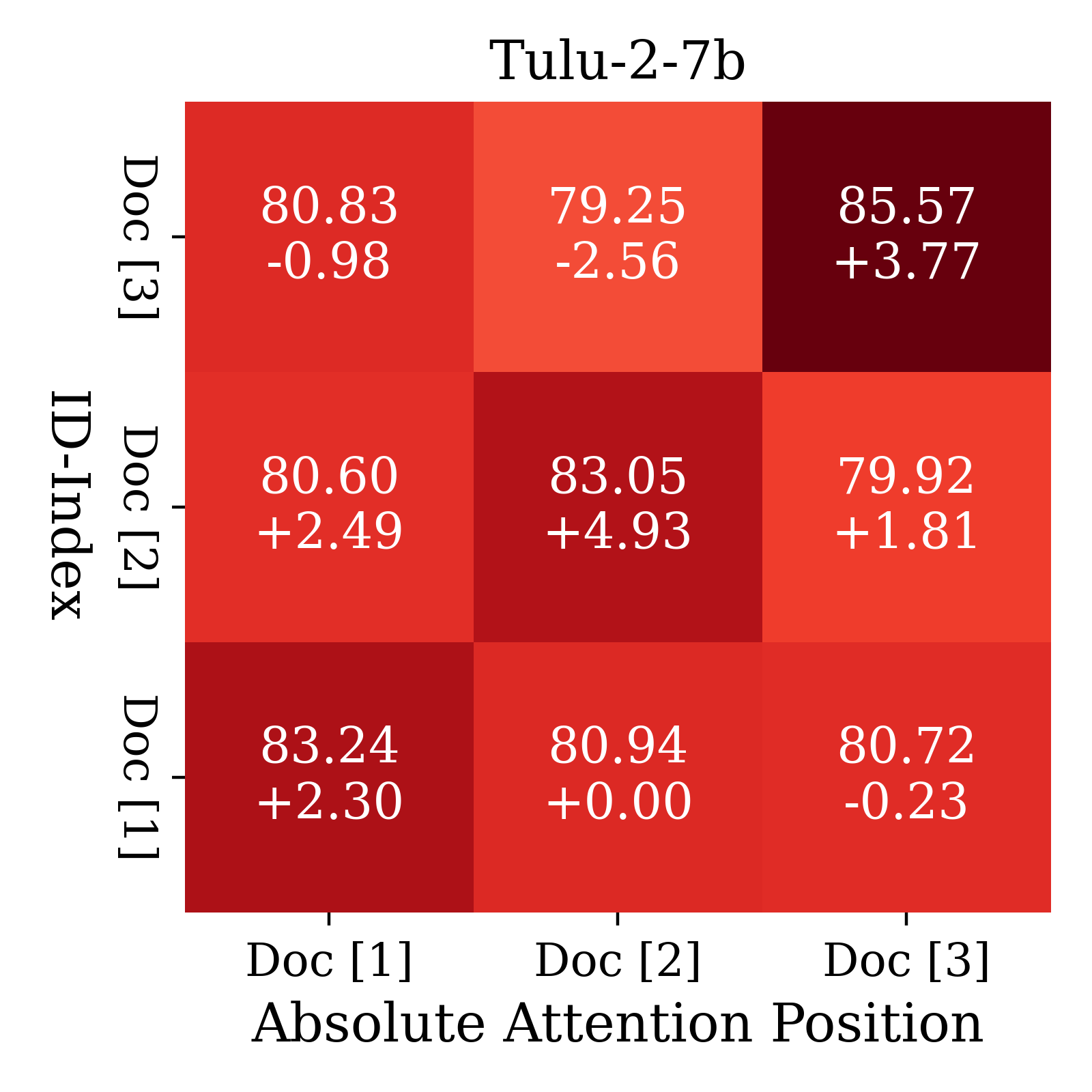}
    \end{minipage}
    \begin{minipage}[b]{0.47\columnwidth}
        \centering
        \includegraphics[width=\columnwidth]{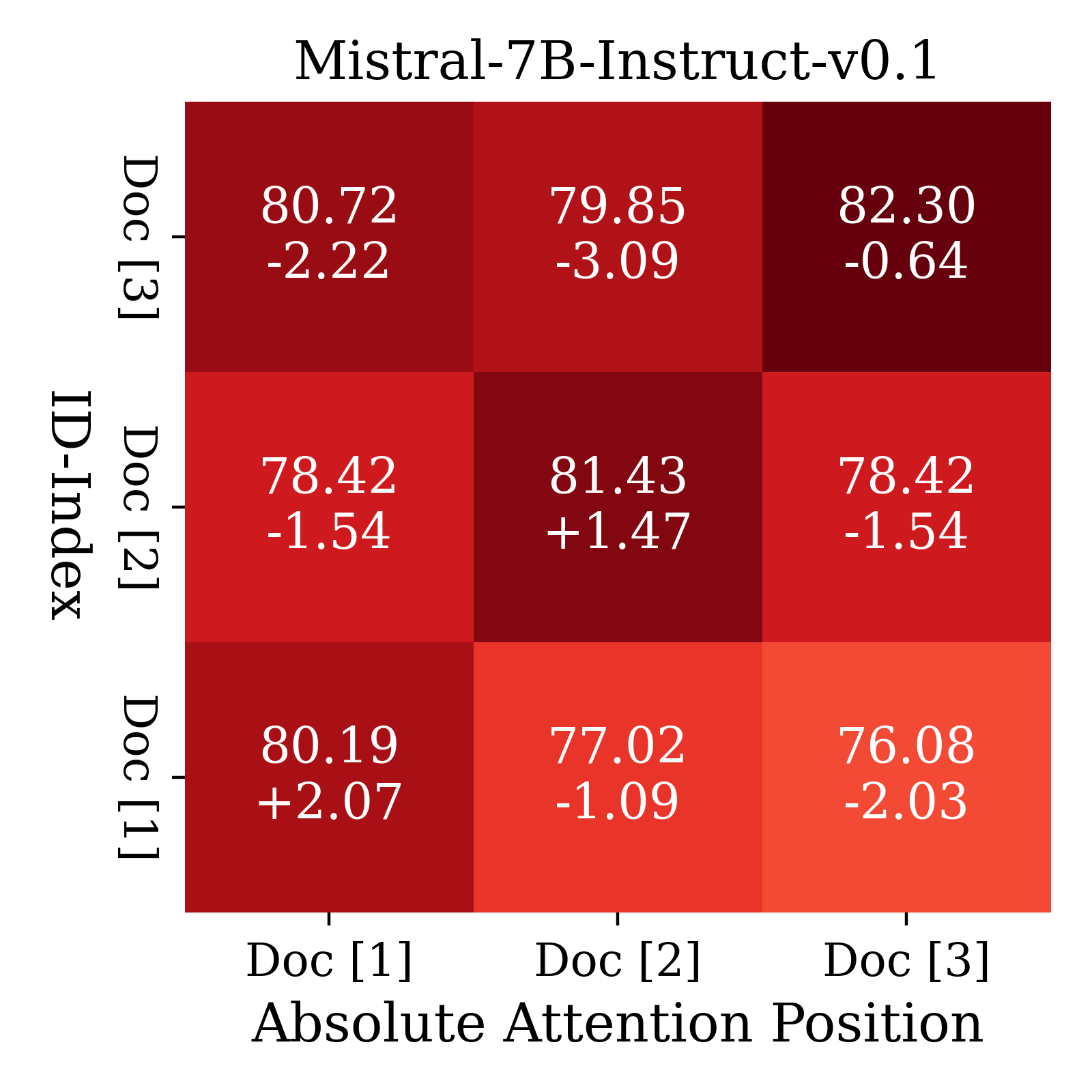}
    \end{minipage}
    \begin{minipage}[b]{0.47\columnwidth}
        \centering
        \includegraphics[width=\columnwidth]{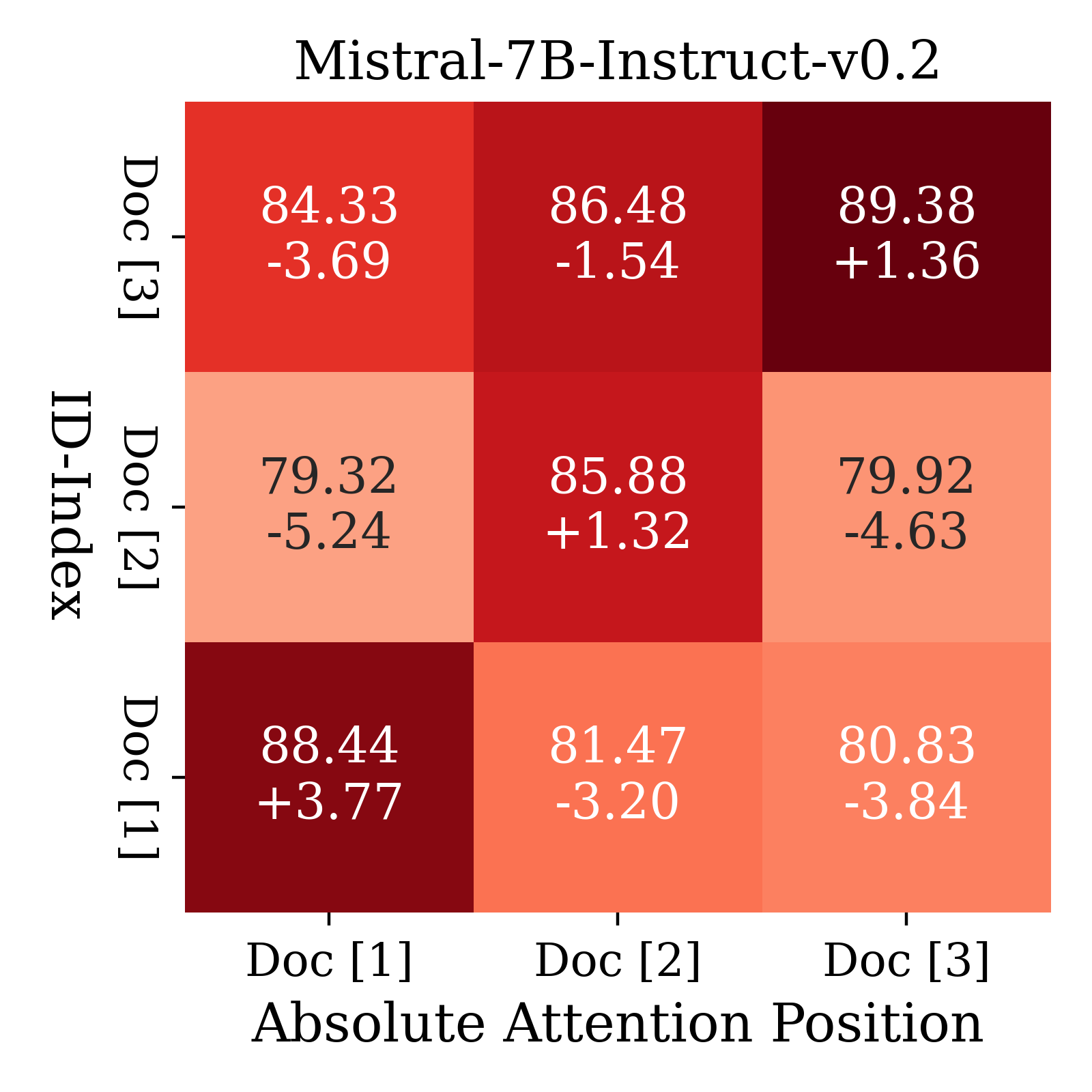}
    \end{minipage}
    \caption{3-document: absolute attention instruction under reversed ID-index setting}
    \label{fig:all_token_have_docid_reversed_3docs}
\end{figure}

\begin{figure}[h]
    \centering
    \begin{minipage}[b]{0.47\columnwidth}
        \centering
        \includegraphics[width=\columnwidth]{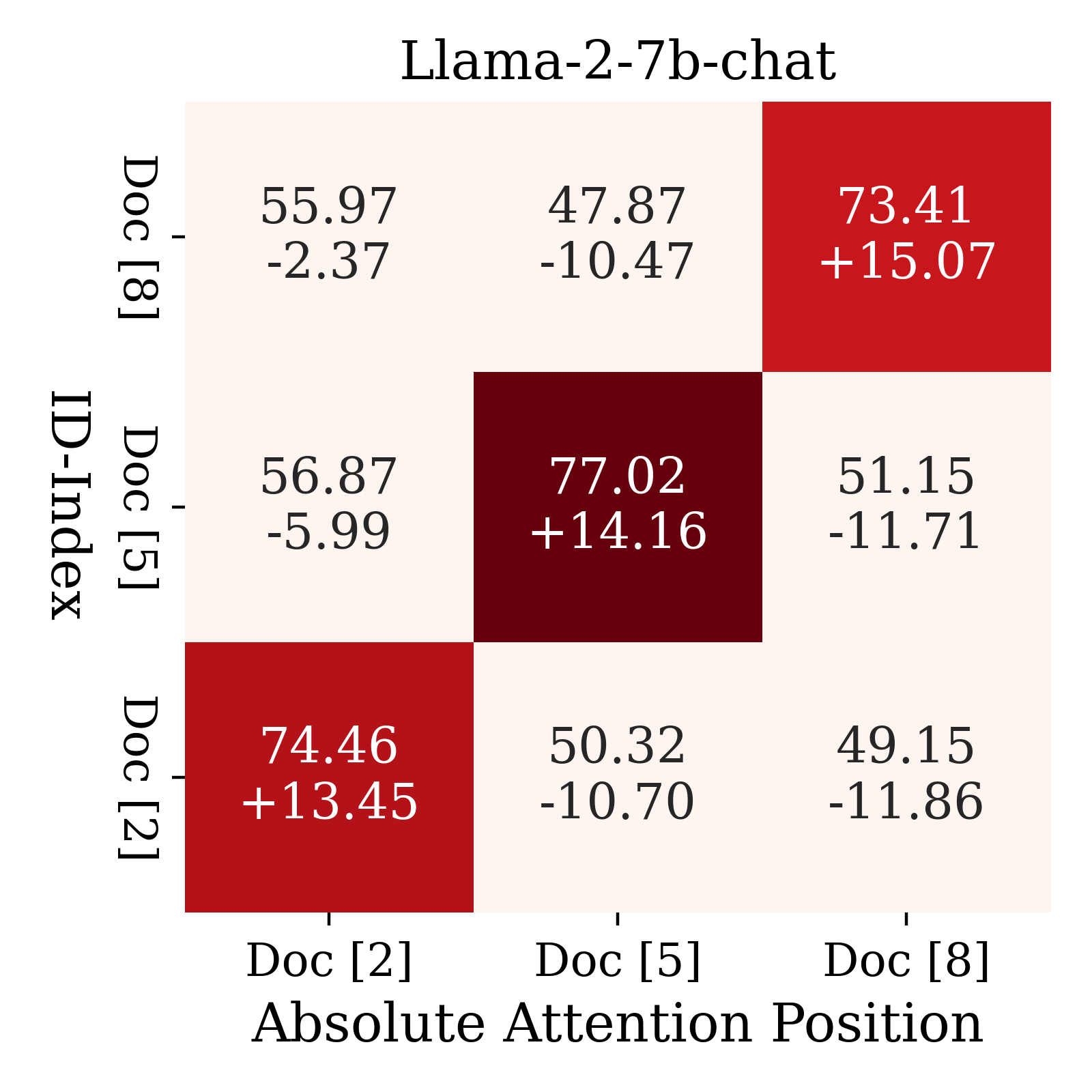}
    \end{minipage}
    \begin{minipage}[b]{0.47\columnwidth}
        \centering
        \includegraphics[width=\columnwidth]{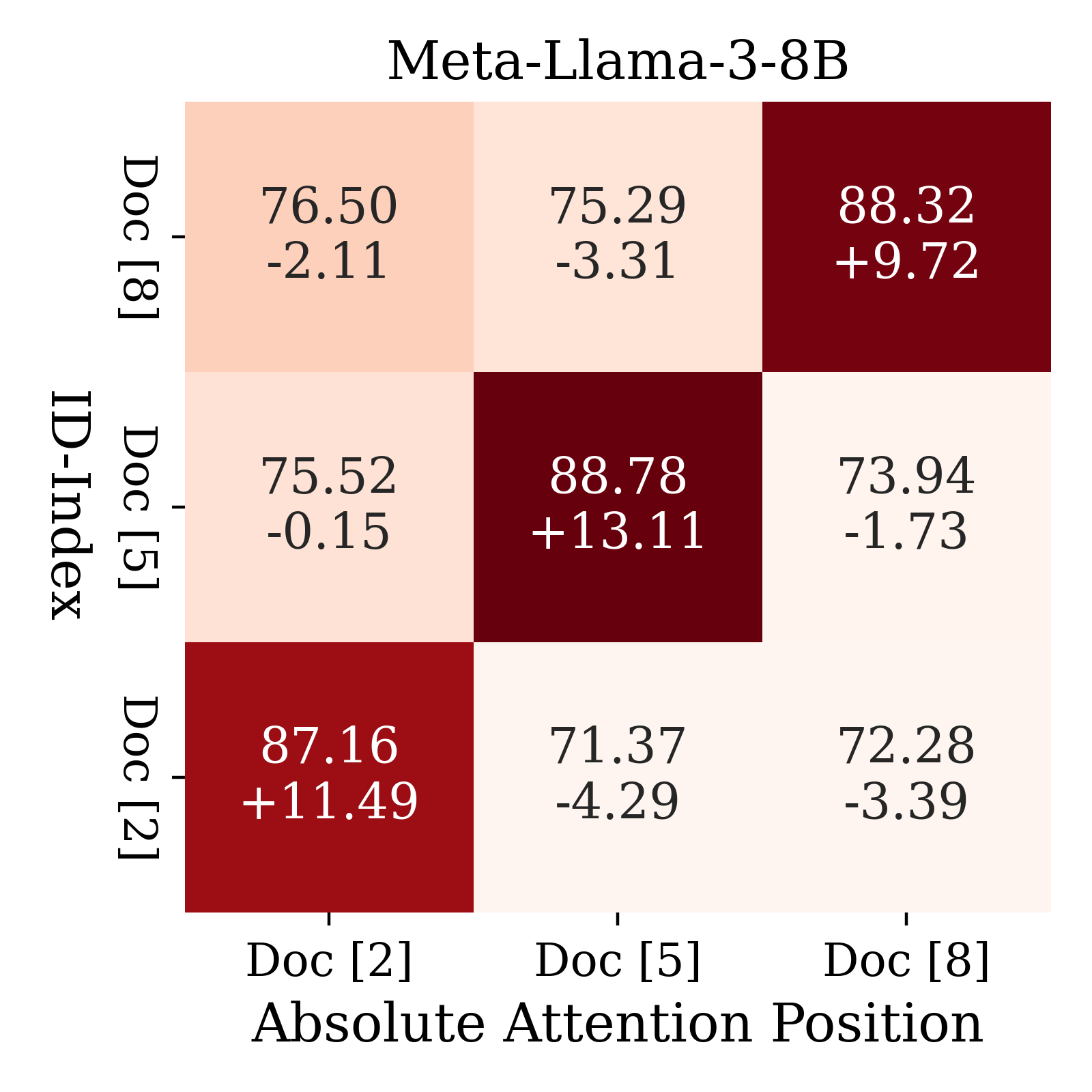}
    \end{minipage}
    \begin{minipage}[b]{0.47\columnwidth}
    \centering
    \includegraphics[width=\columnwidth]{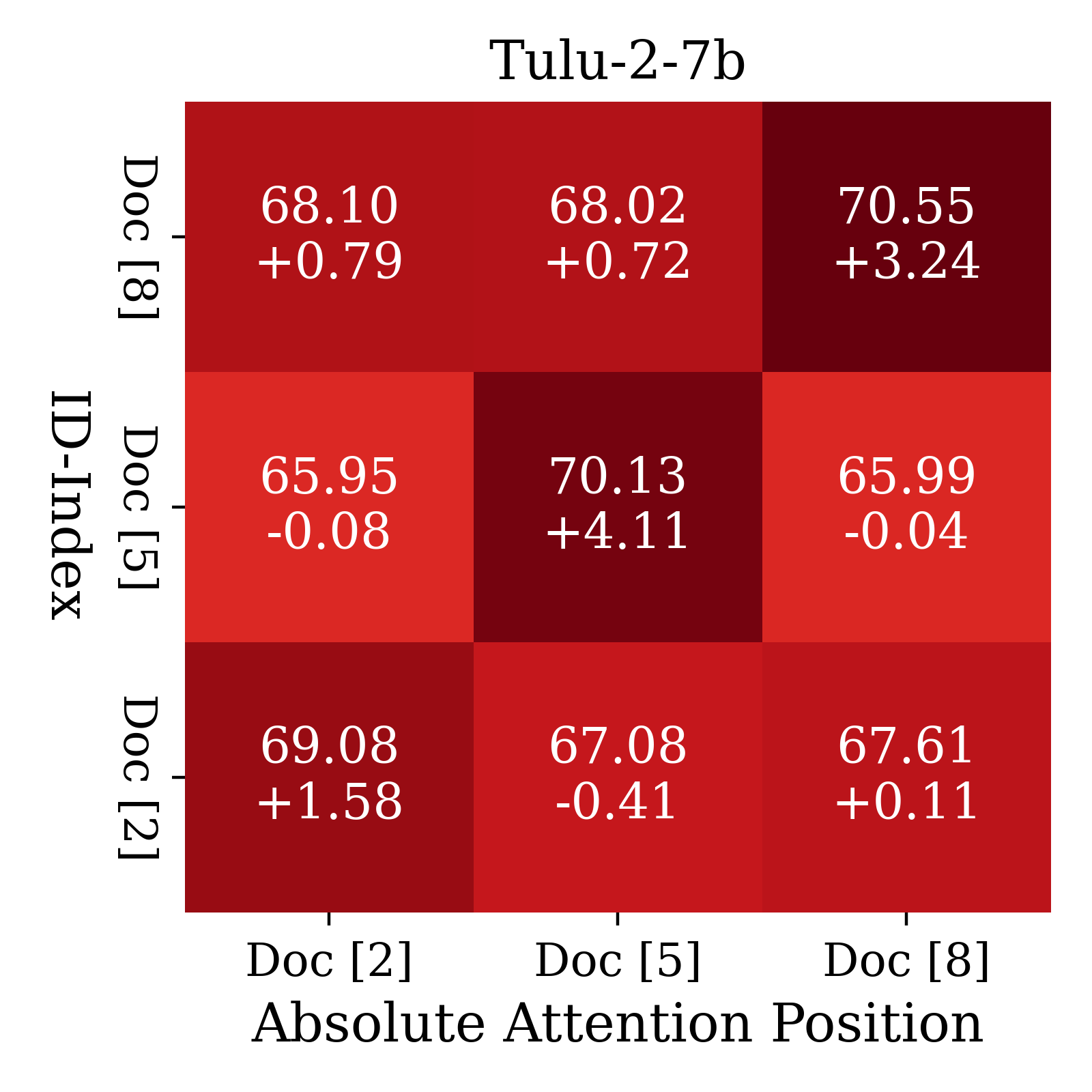}
    \end{minipage}
    \begin{minipage}[b]{0.47\columnwidth}
        \centering
        \includegraphics[width=\columnwidth]{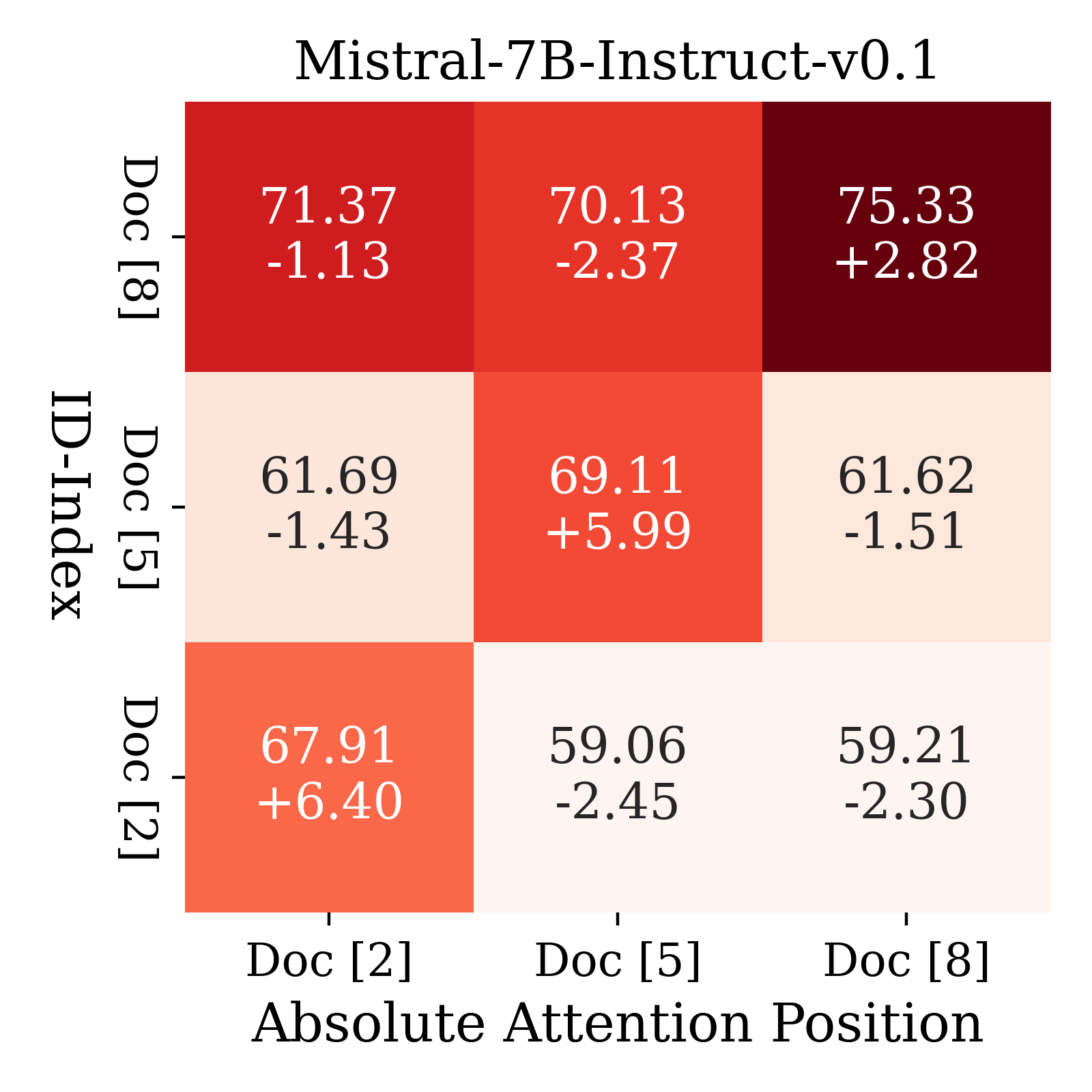}
    \end{minipage}
    \begin{minipage}[b]{0.47\columnwidth}
        \centering
        \includegraphics[width=\columnwidth]{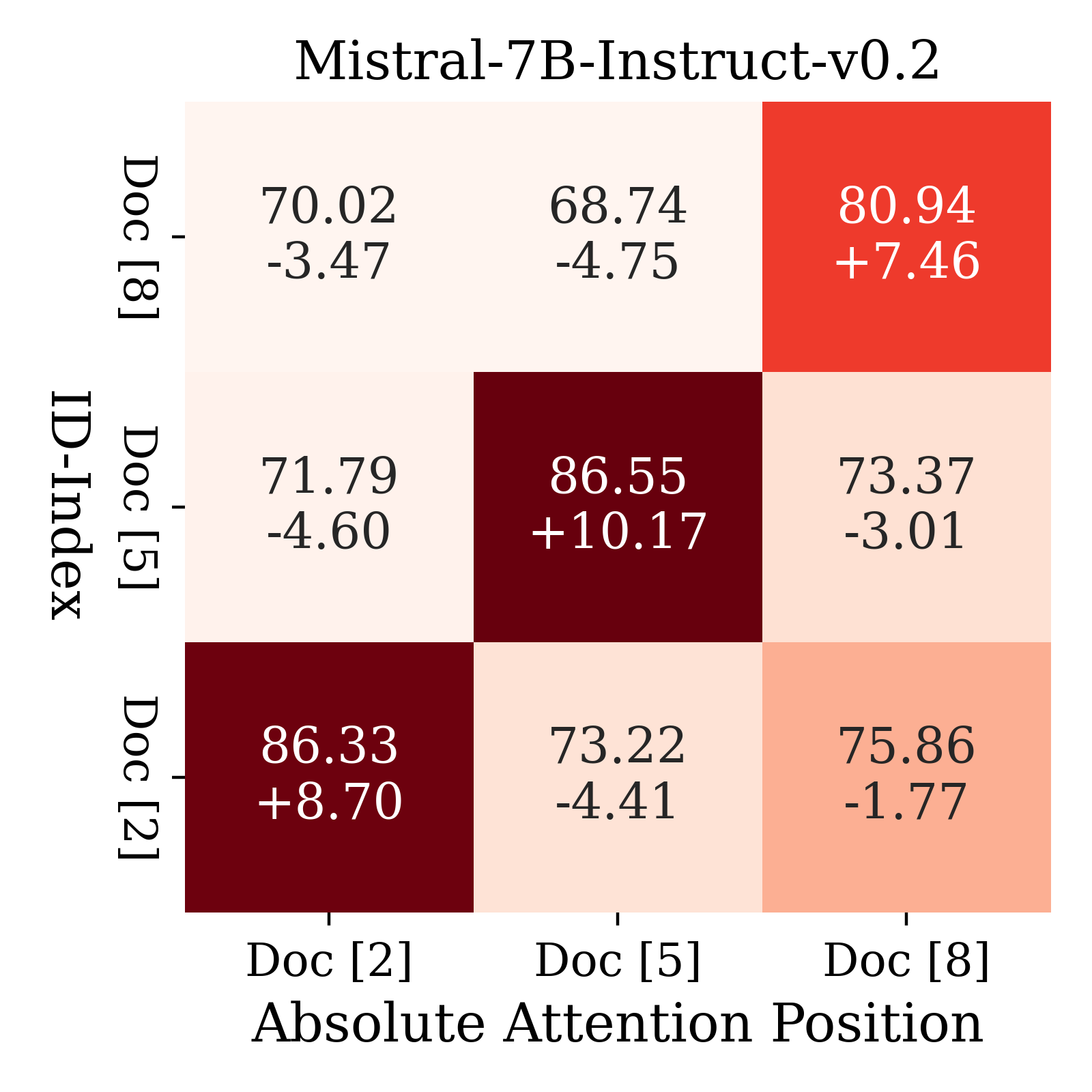}
    \end{minipage}
    \caption{9-document: absolute attention instruction under reversed ID-index setting}
    \label{fig:all_token_have_docid_reversed_9docs}
\end{figure}

3-document results in \Cref{fig:all_token_have_docid_reversed_3docs}, 9-document results in \Cref{fig:all_token_have_docid_reversed_9docs}

\clearpage

\subsection{Absolute Attention Instruction with Position Index}
\label{absolute_posword}
\begin{figure}[h]
    \centering
    \begin{minipage}[b]{0.47\columnwidth}
        \centering
        \includegraphics[width=\columnwidth]{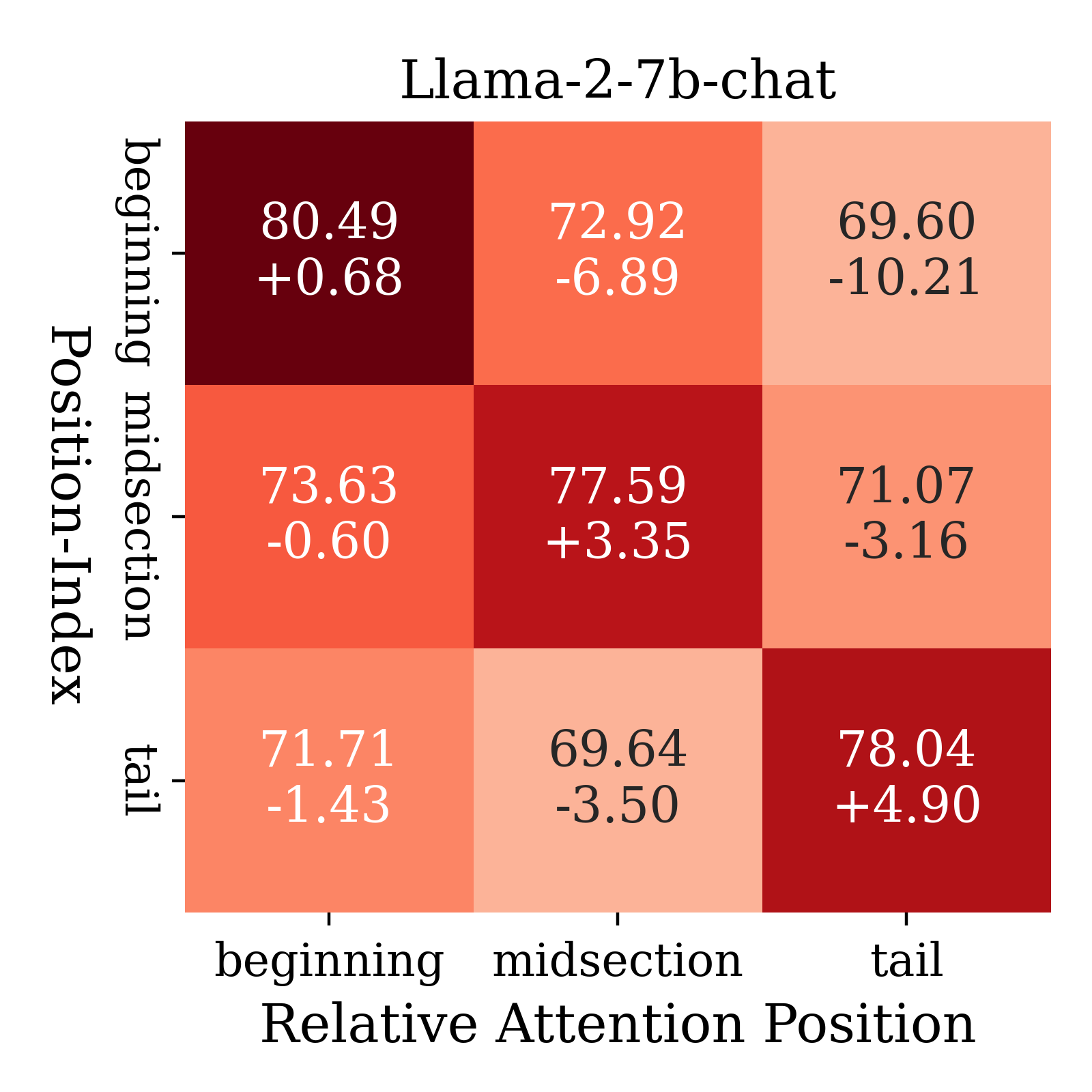}
    \end{minipage}
    \begin{minipage}[b]{0.47\columnwidth}
        \centering
        \includegraphics[width=\columnwidth]{images/Meta-Llama-3-8B-Instruct/3_documents_position_level_have_docid_with_ascending_posword.png}
    \end{minipage}
    \begin{minipage}[b]{0.47\columnwidth}
    \centering
    \includegraphics[width=\columnwidth]{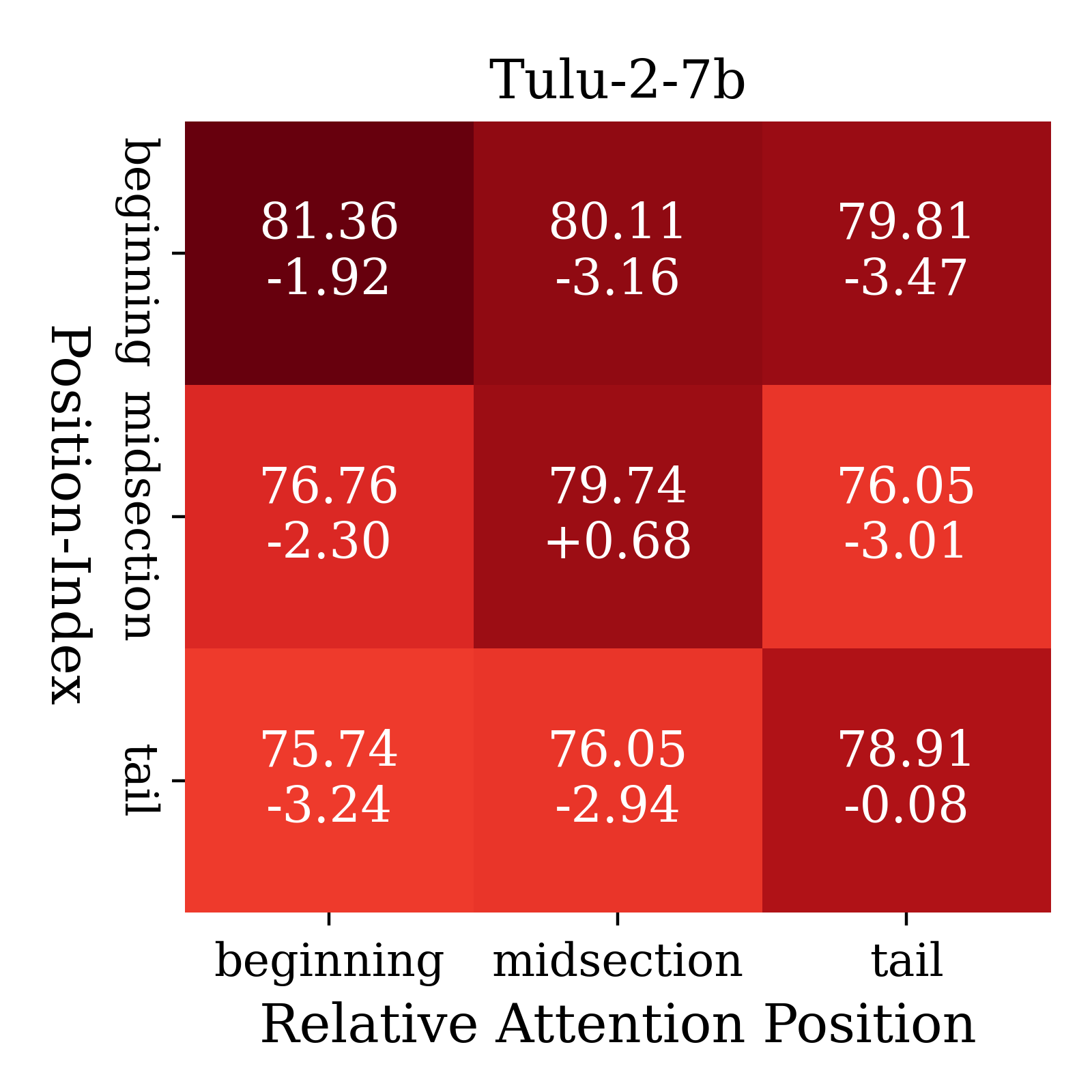}
    \end{minipage}
    \begin{minipage}[b]{0.47\columnwidth}
        \centering
        \includegraphics[width=\columnwidth]{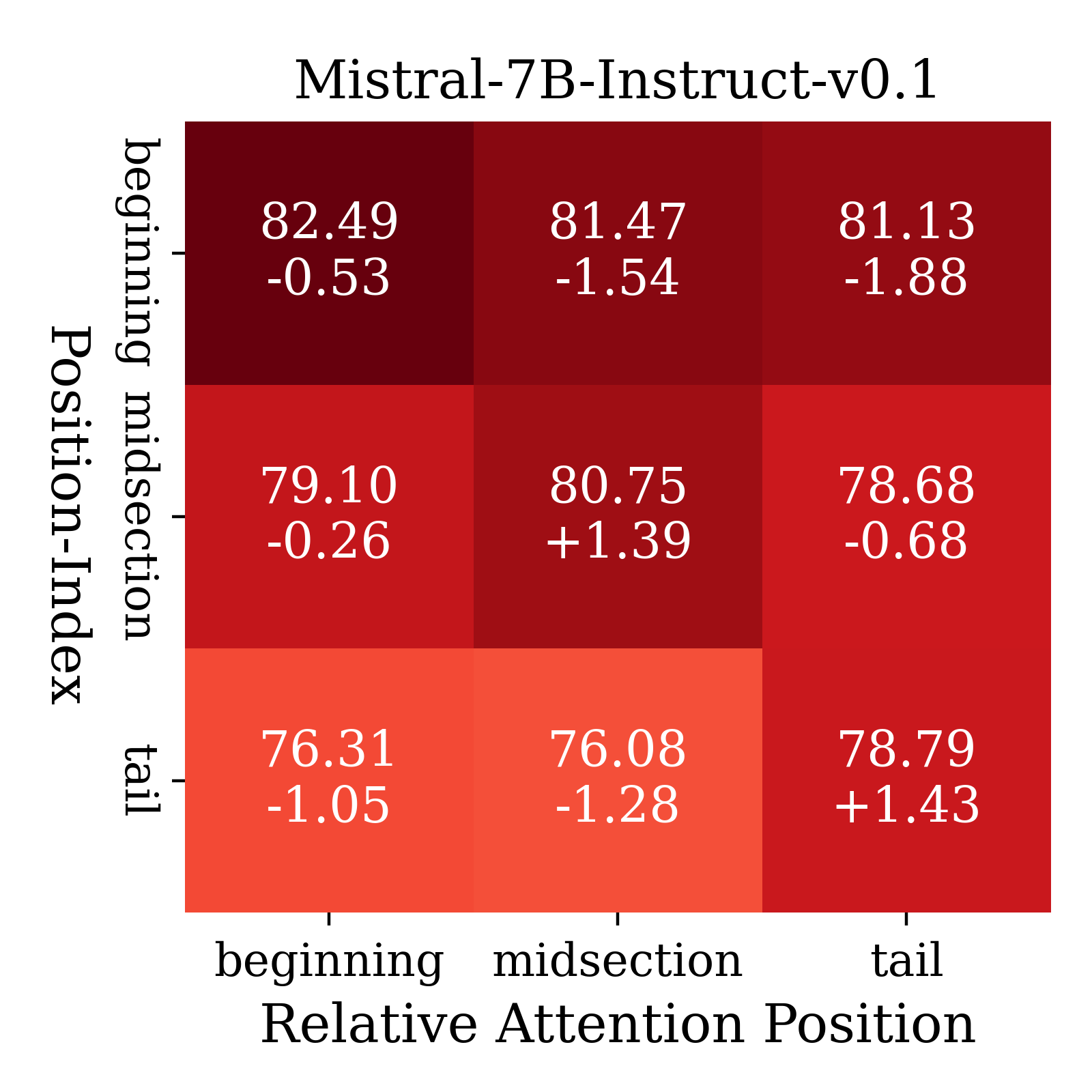}
    \end{minipage}
    \begin{minipage}[b]{0.47\columnwidth}
        \centering
        \includegraphics[width=\columnwidth]{images/Mistral-7B-Instruct-v0.2/3_documents_position_level_have_docid_with_ascending_posword.png}
    \end{minipage}
    \caption{3-document: absolute attention instruction under Position-index setting}
    \label{fig:all_token_have_docid_posword_3docs}
\end{figure}

\begin{figure}[h]
    \centering
    \begin{minipage}[b]{0.47\columnwidth}
        \centering
        \includegraphics[width=\columnwidth]{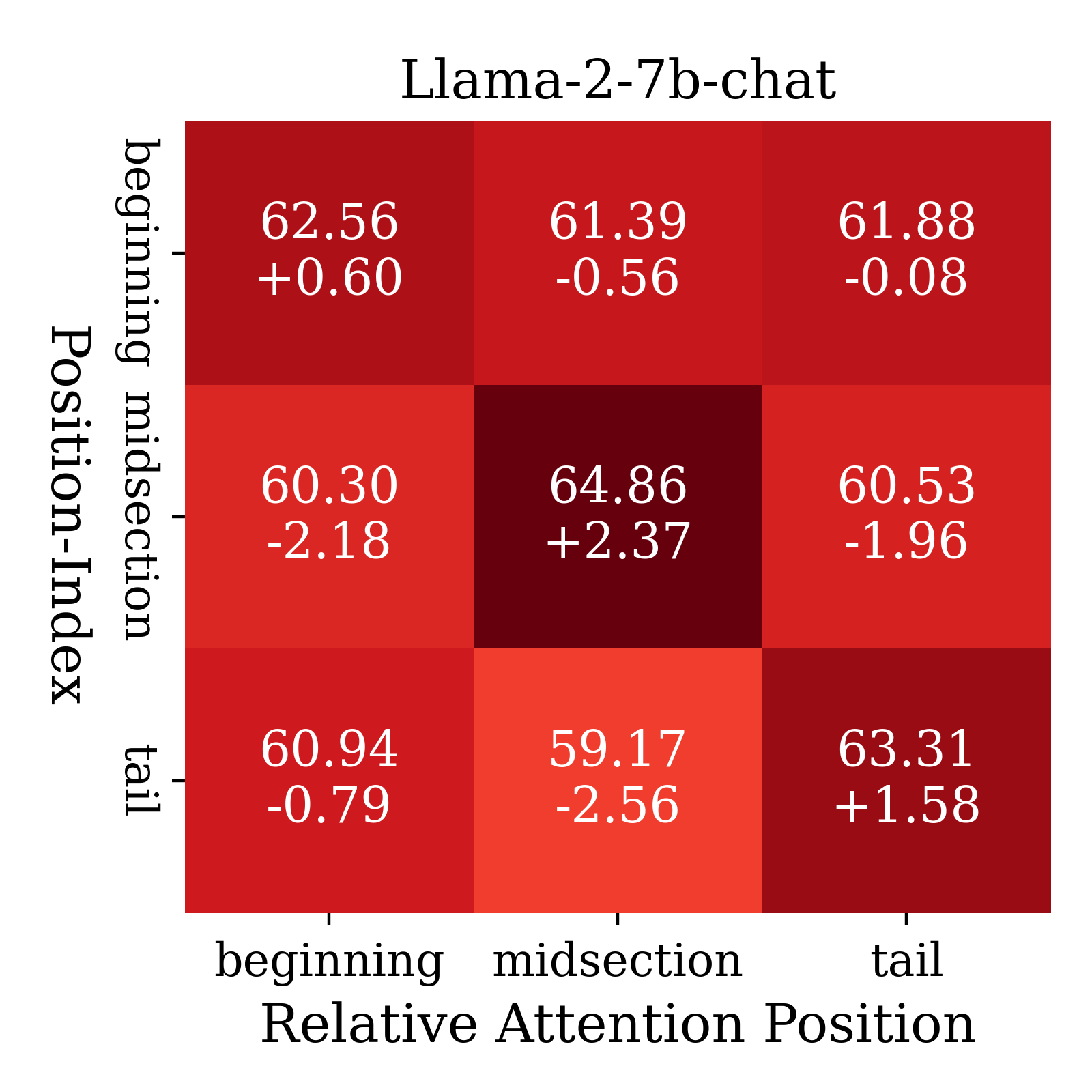}
    \end{minipage}
    \begin{minipage}[b]{0.47\columnwidth}
        \centering
        \includegraphics[width=\columnwidth]{images/Meta-Llama-3-8B-Instruct/9_documents_position_level_have_docid_with_ascending_posword.png}
    \end{minipage}
    \begin{minipage}[b]{0.47\columnwidth}
    \centering
    \includegraphics[width=\columnwidth]{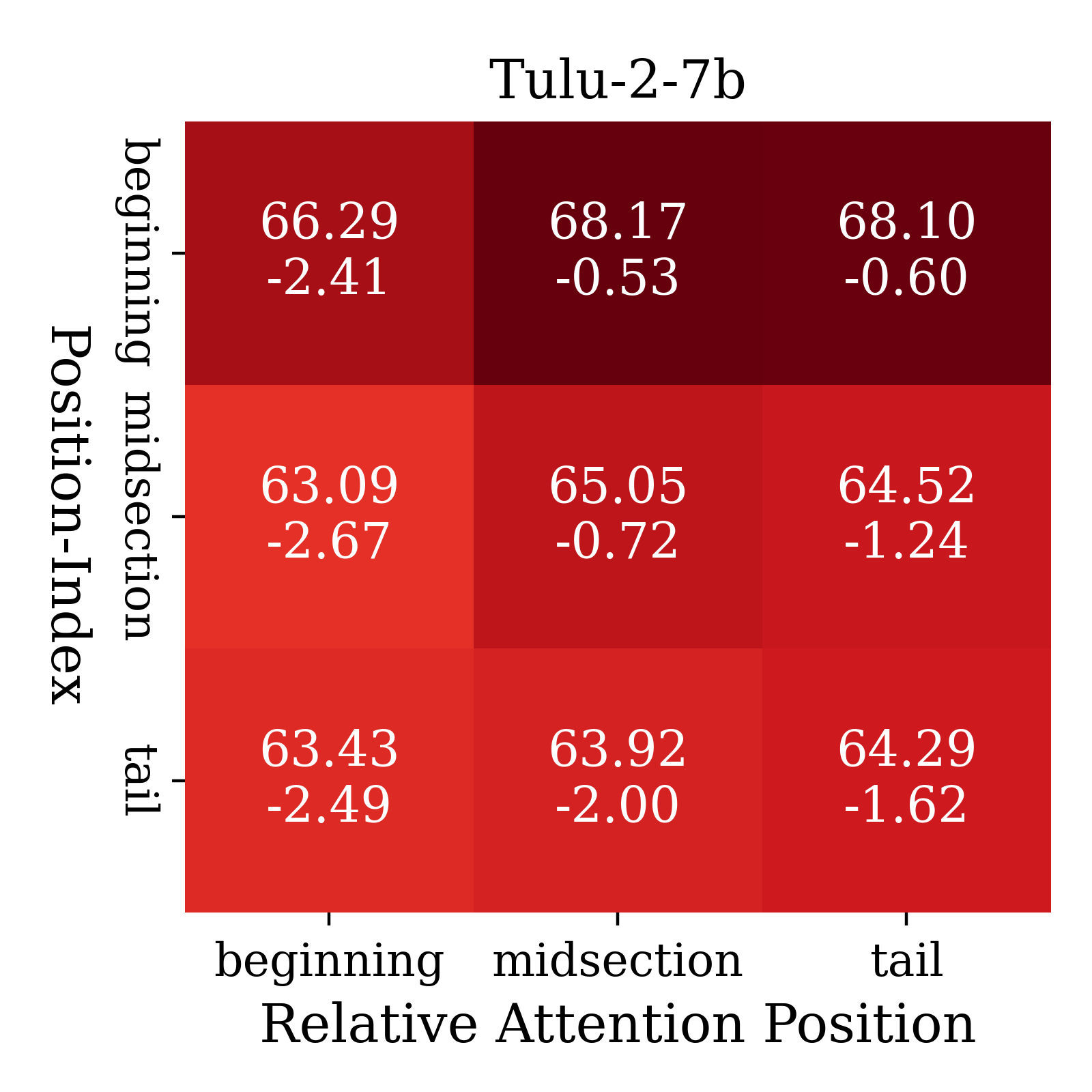}
    \end{minipage}
    \begin{minipage}[b]{0.47\columnwidth}
        \centering
        \includegraphics[width=\columnwidth]{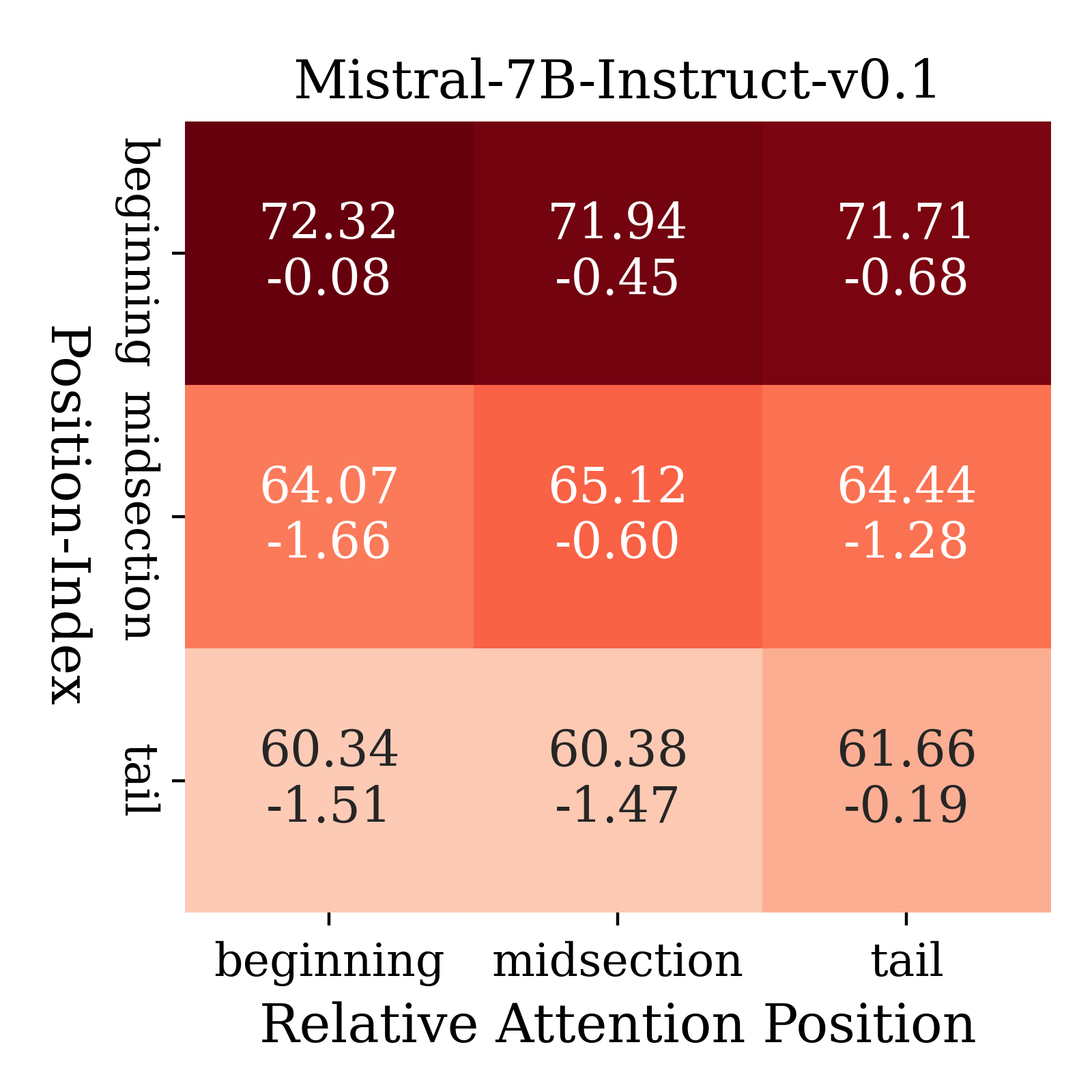}
    \end{minipage}
    \begin{minipage}[b]{0.47\columnwidth}
        \centering
        \includegraphics[width=\columnwidth]{images/Mistral-7B-Instruct-v0.2/9_documents_position_level_have_docid_with_ascending_posword.png}
    \end{minipage}
    \caption{9-document: absolute attention instruction under Position-index setting}
    \label{fig:all_token_have_docid_posword_9docs}
\end{figure}

3-document results in \Cref{fig:all_token_have_docid_posword_3docs}, 9-document results in \Cref{fig:all_token_have_docid_posword_9docs}

\clearpage

\end{document}